\documentclass[twoside]{article}

\usepackage[accepted]{aistats2022}
\usepackage[round]{natbib}

\usepackage[utf8]{inputenc} % allow utf-8 input
\usepackage[T1]{fontenc}    % use 8-bit T1 fonts
\usepackage{hyperref}       % hyperlinks
\usepackage{url}            % simple URL typesetting
\usepackage{booktabs}       % professional-quality tables
\usepackage{amsfonts}       % blackboard math symbols
\usepackage{nicefrac}       % compact symbols for 1/2, etc.
\usepackage{microtype}      % microtypography

\usepackage{microtype}
\usepackage{graphicx}
\usepackage{subfigure}
\usepackage{booktabs} % for professional tables

\usepackage{amsmath,amsthm,amsfonts,amssymb,amscd}

\usepackage{enumerate}
\usepackage{fancyhdr}
\usepackage{mathrsfs}
\usepackage{xcolor}
\usepackage{graphicx}
\usepackage{listings}
\usepackage{hyperref}
\usepackage{caption}
\usepackage{adjustbox}
\usepackage{multirow}
\usepackage{microtype}
\usepackage{tikz}
\usepackage{pgfplots}
\pgfplotsset{compat=newest}
\usepgfplotslibrary{groupplots}
\usepgfplotslibrary{dateplot}
\usepackage{caption}
\usepackage{todonotes}
\usepackage{enumitem}
\tikzset{
  causalvar/.style      = {draw, circle, node distance = 2cm}
}

\usepackage{amsmath}
\usetikzlibrary{matrix}
\usetikzlibrary{shapes,arrows}
\usepackage{mathtools}
\usepackage{bbm}
\DeclareMathOperator*{\argmin}{arg\,min}
\DeclareMathOperator*{\argmax}{arg\,max}

\DeclareMathOperator{\E}{\mathbb{E}}
\newcommand{\cH}{\mathcal{H}}
\newcommand{\cX}{\mathcal{X}}
\newcommand{\cA}{\mathcal{A}}
\newcommand{\cC}{\mathcal{C}}
\newcommand{\cZ}{\mathcal{Z}}
\newcommand{\cS}{\mathcal{S}}
\newcommand{\cO}{\mathcal{O}}
\newcommand{\cF}{\mathcal{F}}
\newcommand{\deff}{d_{\mathrm{eff}}}

\renewcommand{\epsilon}{\varepsilon}
\renewcommand{\Im}{\mathrm{Im}}
\newcommand{\R}{\mathbb{R}}
\newcommand{\defeq}{\vcentcolon=}
\newcommand{\eqdef}{=\vcentcolon}
\DeclareMathOperator{\Tr}{Tr}

\newtheorem{theorem}{Theorem}[section]
\newtheorem{proposition}{Proposition}[section]
\newtheorem{corollary}{Corollary}[section]
\newtheorem{lemma}{Lemma}[section]

\newtheorem{definition}{Definition}[section]
\usepackage{tabularx}
\newcommand\eg{\emph{e.g.}}

\usepackage[ruled,vlined]{algorithm2e}
\SetKwComment{Comment}{\#}{}

\usepackage{microtype}
\usepackage{graphicx}
\usepackage{subfigure}
\usepackage{booktabs} % for professional tables
\PassOptionsToPackage{numbers}{natbib}
\usepackage{natbib}

\usepackage{amsmath,amsthm,amsfonts,amssymb,amscd}

\usepackage{enumerate}
\usepackage{fancyhdr}
\usepackage{mathrsfs}
\usepackage{xcolor}
\usepackage{graphicx}
\usepackage{listings}
\usepackage{hyperref}
\usepackage{caption}
\usepackage{adjustbox}
\usepackage{multirow}
\usepackage{microtype}
\usepackage{tikz}
\usepackage{pgfplots}
\pgfplotsset{compat=newest}
\usepgfplotslibrary{groupplots}
\usepgfplotslibrary{dateplot}
\usepackage{caption}
\usepackage{todonotes}
\tikzset{
  causalvar/.style      = {draw, circle, node distance = 2cm}
}
\usetikzlibrary{matrix}
\usetikzlibrary{shapes,arrows}
\usepackage{mathtools}
\usepackage{dsfont}

\usepackage{amssymb}% http://ctan.org/pkg/amssymb
\usepackage{pifont}% http://ctan.org/pkg/pifont

\newtheorem{innercustomthm}{Theorem}
\newenvironment{customthm}[1]
  {\renewcommand\theinnercustomthm{#1}\innercustomthm}
  {\endinnercustomthm}

 \newtheorem{innercustomlemma}{Lemma}
\newenvironment{customlemma}[1]
  {\renewcommand\theinnercustomlemma{#1}\innercustomlemma}
  {\endinnercustomlemma}
  
 \newtheorem{innercustomprop}{Proposition}
\newenvironment{customprop}[1]
  {\renewcommand\theinnercustomprop{#1}\innercustomprop}
  {\endinnercustomprop}
  
   \newtheorem{innercustomcor}{Corollary}
\newenvironment{customcor}[1]
  {\renewcommand\theinnercustomcor{#1}\innercustomcor}
  {\endinnercustomcor}

\begin{document}


\runningauthor{Zenati, Bietti, Diemert, Mairal, Martin, Gaillard}
\newcommand{\codeurl}{\url{https://github.com/criteo-research/Efficient-Kernel-UCB}}
\twocolumn[

\aistatstitle{Efficient Kernel UCB for Contextual Bandits}

\title{Efficient Kernel UCB for Contextual Bandits}

% \aistatsauthor{Houssam Zenati \And Alberto Bietti \And Eustache Diemert \\ \And Julien~Mairal \And Matthieu Martin \And Pierre Gaillard}
\aistatsauthor{Houssam Zenati \And Alberto Bietti \And Eustache Diemert}

\aistatsaddress{Criteo AI Lab, Inria\footnotemark[1] \And  NYU Center for Data Science \And Criteo AI Lab}

\aistatsauthor{Julien Mairal \And Matthieu Martin \And Pierre Gaillard}

\aistatsaddress{Inria\footnotemark[1]  \And Criteo AI Lab \And Inria\footnotemark[1]}
%Univ. Grenoble Alpes, Inria, \\ CNRS Grenoble INP, LJK, 38000 \And  \And Criteo AI Lab} 

% \aistatsaddress{Criteo AI Lab \And  Center for Data Science, New York University. New York, USA \And Criteo AI Lab \And  Univ. Grenoble Alpes, Inria, CNRS, Grenoble INP, LJK, 38000 Grenoble, France \And Criteo AI Lab \And  Univ. Grenoble Alpes, Inria, CNRS, Grenoble INP, LJK, 38000 Grenoble, France}

%\aistatsauthor{Julien~Mairal \And Matthieu Martin \And Pierre Gaillard}

%\aistatsaddress{Univ. Grenoble Alpes, Inria, \\ CNRS Grenoble INP, LJK, 38000 \And  Criteo AI Lab \And Univ. Grenoble Alpes, Inria, \\ CNRS Grenoble INP, LJK, 38000}

]

\footnotetext[1]{Univ. Grenoble  Alpes,  Inria,  CNRS, Grenoble  INP,  LJK,  38000  Grenoble, France.}

\begin{abstract}
In this paper, we tackle the computational efficiency of kernelized UCB algorithms in contextual bandits. While standard methods require a $\cO(CT^3)$ complexity where~$T$ is the horizon and the constant $C$ is related to optimizing the UCB rule, we propose an efficient contextual algorithm for large-scale problems. Specifically, our method relies on incremental Nystr\"om approximations of the joint kernel embedding of contexts and actions. This allows us to achieve a complexity of $\cO(CTm^2)$ where $m$ is the number of Nystr\"om points. To recover the same regret as the standard kernelized UCB algorithm, $m$ needs to be of order of the effective dimension of the problem, which is at most $\cO(\sqrt{T})$ and nearly constant in some cases. 
%. We analyze the regret in different regimes where the size of dictionary varies and show how the projected algorithm recovers the standard regret while being computationally efficient. Finally, we validate our method with numerical experiments and illustrate the computational gains. 
\end{abstract}

\section{Introduction}

Contextual bandits for sequential decision making have become ubiquitous in many applications such as online recommendation systems \citep{li2010}. % and offer many promises  precision medicine \citep{bastani2019}.
% While such applications may be performed in an offline setting \citep{bottou2012}, the bandit setting \citep{lattimore_szepesvari_2020} assumes sequential decision making and information collection.
At each round, an agent observes a \emph{context} vector and chooses an \textit{action}; then, the \textit{environment} generates a \textit{reward} based on the chosen action. The goal of the agent is to maximize the cumulative reward over time, which requires a careful balancing between exploitation (maximizing reward using past observations) and  exploration (increasing the diversity of observations). 

In this paper, we consider a kernelized contextual bandit framework, where the rewards are modeled by a function in a reproducing kernel Hilbert space (RKHS). In other words, we assume the expected reward to be linear with respect to a joint context-action feature map of possibly infinite dimension.
This setup provides flexible modeling choices through the feature map for both discrete and continuous action sets, and exploration algorithms typically rely on constructing confidence sets for the parameter vector and exploring using upper confidence bound (UCB) rules~\citep{li2010}.
The extensions to infinite-dimensional feature maps we consider has been introduced by~\citet{krause2011contextual, valko2013} using kernelized variants of UCB, which allow effective exploration even for rich non-parametric reward functions lying in a RKHS, such as smooth functions over contexts and/or actions.

% Upper confidence bounds (UCB) algorithms in the bandit literature have been adapted to the contextual bandit setting by extending the stochastic linear setting (see, e.g LinUCB \citep{li2010}) to include contextual information to the decision process, and has been extended to a kernel version (see \citep{krause2011contextual, valko2013}). Such methods estimate at each round over an upper confidence bound of the reward over the action space, given a contextual information. Thus, such optimistic methods explore directions that sequentially improve the predictions by reducing the uncertainty over the confidence sets and then leading to more confident predictions over the joint context-action space to minimize the agent regret. 

Despite the rich modeling capabilities of such kernelized UCB algorithms,
% While such methods have the advantage to use prior knowledge on the reward structure such as smoothness priors,
they lack scalability since standard algorithms scale at best as $\cO(CT^3)$ where $T$ is the horizon (total number of rounds) and the constant $C$ is the cost of selecting an action according to the UCB optimization rule.
This large cost is due to the need to solve linear systems involving a $t \times t$ kernel matrix at each round~$t$,
and motivates developing efficient versions of these algorithms for large problems.
In supervised learning, a common technique for reducing computation cost is to leverage the fact that the kernel matrix is often approximately low-rank, and to use Nystr\"om approximations~\citep{williams2001using,alessandro2015}.
We extend such approximations to the contextual bandit setting, by relying on incremental updates of a dictionary of Nystr\"om anchor points, which allows us to reduce the complexity to~$\cO(CT m^2)$, where~$m$ is the final number of dictionary elements. In order to preserve a small regret comparable to the vanilla kernel UCB method, $m$ is of the order of an \emph{effective dimension} quantity, which is typically much smaller than~$T$, and at most~$\sqrt{T}$.

Closely related to our work, \citet{calandriello19a,calandriello20a} recently considered Nystr\"om approximations in the non-contextual setting with finite actions, corresponding to a Bayesian optimization problem. Whereas their algorithm is effective when there are no contexts, a direct extension to the contextual setting yields a complexity of $\cO(T m^3)$, which may be $\cO(T^{2.5})$ in the worst case, despite a batching strategy allowing to recompute a new dictionary only about $m$ times.
In contrast, our incremental strategy reduces the previous complexity to $\cO(T m^2)$, and thus at most $\cO(T^2)$.

Even though adopting an incremental strategy for updating the Nystr\"om dictionary may seem to be a simple idea, achieving the previously-mentioned complexity while preserving a regret that is comparable to the original kernel UCB approach is non-trivial. Nyström approximations cause dependencies in the projected kernel matrix that makes it difficult to use martingale arguments, which led \citet{calandriello20a} to use other mathematical tools that are compatible with updates resampling a new Nystr\"om dictionary.
In contrast, we manage to use martingale arguments for an incremental strategy that is less computationally expensive. For that, we extend the standard analysis of the OFUL algorithm for linear bandits~\citep{oful2011,chowdhury2017kernelized} to the kernel setting with Nyström approximations. In particular, this requires non-trivial extensions of concentration bounds to infinite-dimensional objects. Our analysis also uses the incremental structure of the projections that \citet{calandriello19a} do not have.  This allows us to prove the complexity of our algorithm. Moreover, unlike previous works, we explicit the regret-complexity trade-off under the capacity condition assumption. Finally, we also provide numerical experiments showing that our theoretical gains are also observed in practice.

\begin{table*}[h]
\centering
\begin{tabular}{c|c|c|c}
Algorithm &  Regret & Space & Time Complexity \\ \hline
CGP-UCB \citep{krause2011contextual} &  $\cO(\sqrt{T} \deff(\lambda, T))$  & $\cO(T^2)$ & $\cO(CT^3)$ \\ 
SupKernelUCB \citep{valko2013}&   $\cO(\sqrt{T \deff(\lambda, T) \log(C)})$ & $\cO(T^2)$ & $\cO(CT^3)$ \\ \hline
C-BKB \citep{calandriello19a} & $\cO(\sqrt{T} (\sqrt{\lambda \deff(\lambda, T)} + \deff(\lambda, T))$  & $\cO\left(T \deff \right)$ & $\cO\left(T^2 \deff^2 + CT\deff^2\right)$ \\ 
C-BBKB \citep{calandriello20a}  & $\cO(\sqrt{T} (\sqrt{\lambda \deff(\lambda, T)} + \deff(\lambda, T))$  & $\cO\left(T \deff\right)$  & $\cO\left(T \deff^{3} + CT \deff^{2}\right)$ \\  \hline
K-UCB (ours)  & $\cO(\sqrt{T} (\sqrt{\lambda \deff(\lambda, T)} + \deff(\lambda, T))$  & $\cO(T^2)$ &  $\cO(CT^3)$ \\ 
EK-UCB (ours)  & $\cO(\sqrt{T} (\sqrt{\lambda \deff(\lambda, T)} + \deff(\lambda, T))$ & $\cO\left(T \deff \right)$  & $\cO(CT \deff^{2})$  \\

\end{tabular}

%\caption{Comparison of regret bounds and total time complexity. $\tilde{\cO}$ denotes a $\cO$ up to logarithmic factors. If the action space is finite as in SupKernelUCB, we write $C=|\mathcal{A}|$ its cardinality, and note that the argmax is obtained in $C$ computations of the UCB rule. Note that the reported regret of CGP-UCB, SupKernel UCB and CBBKB use here the definition of the effective dimension $\deff$ in Eq.~\eqref{eq:deff}. This quantity is equivalent, up to logarithmic factors, to the information gain used by \cite{krause2009, calandriello20a} and the effective dimension used by \cite{valko2013} (see Appendix \ref{app:comparison-regret}). Note that $\deff$ is an abbreviation for $\deff(\lambda, T)$, it depends on the horizon $T$ and $\lambda$. The parameter $\lambda$ in the bayesian algorithms CGP-UCB, BKB and BBKB corresponds to the inverse of the noise of the Gaussian process. Moreover, we report the complexities of the contextualized versions of BKB and BBKBs, noting that the non-contextual versions may benefit from certain optimizations when the action space is discrete~\citep{calandriello19a,calandriello20a}. Note that under a standard capacity condition on the kernel eigenvalues (see Corollary~\ref{cor:bound_capacity_condition}), the optimal parameters for all methods except SupKernelUCB satisfy $\deff \approx m \approx \lambda \approx T^{\alpha/( 1 + \alpha)}$ where $m$ is the dictionary size and $\alpha$ is in $[0,1]$.}

\caption{Comparison of regret bounds  (up to logarithmic factors in $T$) and total time complexity. When the action space is finite, for e.g in SupKernelUCB, we write $C=|\mathcal{A}|$ its cardinality and note that the argmax is obtained in $C$ computations of the UCB rule. Note that the reported regret of CGP-UCB, SupKernel UCB and CBBKB use here the definition of the effective dimension $\deff(\lambda, T)$ in Eq.~\eqref{eq:deff} which depends on the horizon $T$ and the parameter $\lambda$ (i.e the inverse of the GP noise in CGP-UCB, BKB and BBKB). This effective dimension $\deff$ is equivalent, up to logarithmic factors, to the information gain used by \cite{krause2009, calandriello20a} and the definition used by \cite{valko2013} (see Appendix \ref{app:comparison-regret}). Moreover, we report the complexities of the contextualized versions of BKB and BBKBs, noting that the non-contextual versions may benefit from certain optimizations when the action space is discrete~\citep{calandriello19a,calandriello20a}. 
% Note that under a standard capacity condition on the kernel eigenvalues (see Corollary~\ref{cor:bound_capacity_condition}), the optimal parameters for all methods except SupKernelUCB satisfy $\deff \approx m \approx \lambda \approx T^{\alpha/( 1 + \alpha)}$ where $m$ is the dictionary size and $\alpha$ is in $[0,1]$.
}
\label{table:comparison_algos}
\end{table*}

\section{Related Work}

UCB algorithms are commonly used in the bandit literature to carefully balance exploration and exploitation by defining confidence sets on unknown reward functions~\citep{lattimore_szepesvari_2020}.
For stochastic linear contextual bandits, the OFUL algorithm~\citep{oful2011} obtains improved guarantees compared to previous analyses~\citep[\eg,][]{li2010} by providing tighter confidence bounds based on self-normalized tail inequalities.

Extensions of linear contextual bandits and UCB algorithms to infinite-dimensional representations of contexts or actions have been studied by~\citet{krause2011contextual} and \citet{valko2013} by using kernels and Gaussian processes.
While their analyses involve different concepts of effective dimension, it can be shown that these are closely related (see Section~\ref{sub:kucb_regret}).
\citet{valko2013} notably achieves a better scaling in the horizon in the regret, but requires a finite action space.
\citet{chowdhury2017kernelized} improves the analysis of GP-UCB using tools inspired by~\citet{oful2011} and similar to our analysis of kernel-UCB, though it considers the non-contextual setting. \citet{tirinzoni20} in the contextual linear bandit problem use a primal-dual algorithm to achieve an optimal asymptotical regret bound but does not address the issue of computational complexity nor the kernelized setting. Likewise, \citet{camilleri2021} propose a new estimator in the non-contextual kernelized bandit problem to achieve a tighter regret bound using an elimination algorithm but does no focus on computational efficiency neither. 

In the Bayesian experimental design literature \citet{derezinski20a} propose an efficient sampling scheme using determinant point processes in the non-kernel case and a non-contextual framework. For improving the computational complexity of kernelized UCB procedures in a non-contextual setting as well, \citet{calandriello19a} use a Nystr\"om approximation of the kernel matrix which is recomputed at each step. Because the corresponding algorithm is not practical when a large number of steps are needed, \citet{calandriello20a} consider a batched version, which significantly improves its computation and complexity.

In contrast, we use an incremental construction based on the KORS method~\citep{calandriello2017}, which has been used previously with full information feedback~\citep[see also][]{JezequelEtAl2019}, allowing us to significantly improve the computational complexity of the contextual GP-UCB algorithm, for the same regret guarantee. Such an incremental approach appears to be a key to achieve better complexity than a natural contextual variant of the algorithm of \citet{calandriello20a}, see Table~\ref{table:comparison_algos}, both in theory and in practice (see Section~\ref{sec:exps}).
Such an extension is unfortunately non-trivial and requires a different regret analysis, as discussed earlier.

\citet{mutny2019efficient} also study kernel approximations for efficient variants of GP-UCB, focusing on random feature expansions. Nevertheless, the number of random features may need to be very large--often exponential in the dimension--in order to achieve good regret, due to a misspecification error which requires stronger, uniform approximation guarantees.
Finally, \citet{kuzborskij19a} also considers leverage score sampling for computational efficiency, but focuses on linear bandits in finite dimension.

\section{Warm-up: Kernel-UCB for Contextual Bandits}
\label{sec:kernelUCB}

In this section, we introduce stochastic contextual bandits with reward functions lying in a RKHS, and provide an analysis of the Kernel-UCB algorithm (similar to GP-UCB) which will be a starting point for studying the computationally efficient version in Section~\ref{sec:EKUCB}.
% In this section we provide the background for stochastic contextual bandits and extend the linear bandits framework to contextual bandits. We show how previous algorithms relate to this extension and provide a general upper bound of this algorithm.  

\paragraph{Notations.} We define here basic notations. Given a vector $v \in \mathbb{R}^d$ we write its entries $[v_i]_{1 \leq i \leq d}$ and we will write $v^\top w$ or $\langle v, w \rangle$ the dot product for elements in $\mathbb{R}^d$ and in the Hilbert space $\mathcal{H}$. We denote by $\Vert \cdot \Vert$ the Euclidean norm and the norm in $\mathcal{H}$. The conjugate transpose for a linear operator $L$ on $\mathcal{H}$ is denoted by $L^*$. For two operators $L, L'$ on $\mathcal{H}$, we write $L \preccurlyeq L'$ when  $L-L'$ is positive semi-definite and we use $\lesssim$ for approximate inequalities up to logarithmic multiplicative or additive terms. A summary of the notations is provided in Appendix~\ref{app:notations}.

\subsection{Setup}

In the contextual bandit problem, at each time $t$ in $1, \dots, T$, where $T$ is the horizon, for each context $x_t$ in~$\mathcal{X}$, an action $a_t$ in $\cA$ is chosen by an agent and induces a reward $r_t$ in $\R$. The input and action spaces $\cX$ and $\cA$ can be arbitrary (e.g., finite or included in $\R^d$ for some $d\geq 1$). Note that $\cA$ may change over time, but we keep it fixed here for simplicity.

In this paper, we focus on stochastic kernel contextual bandits and assume that there exists a reproducing kernel Hilbert space (RKHS) $\cH$ such that 
\[
    r_t = \langle \theta^*, \phi(x_t,a_t)\rangle + \epsilon_t \,,
\]
where $\epsilon_t$ are i.i.d. centered subGaussian noise, $\theta^* \in \cH$ is an unknown parameter, and $\phi: \cX \times \cA \to \cH$ is a known feature map associated to $\cH$. It satisfies
\[
    \langle \phi(x,a) , \phi(x',a')\rangle = k\big( (x,a), (x', a') \big)\, ,
\]
where $k$ is a positive definite kernel associated to $\cH$. We assume $k$ to be bounded, i.e., there exists $\kappa >0$ such that $k(s,s) \leq \kappa^2$ for any $s \in \cX \times \cA$.

Thus, the goal of the agent is, given the previously observed contexts, actions and rewards $(x_s, a_s, r_s)_{s=1 \dots t-1}$ and the current context $x_t$, to choose an action $a_t$ in order to minimize the following regret after $T$ rounds
\begin{equation}
    R_T := \E \left[ \sum_{t=1}^T \max_{a \in \cA} \langle \theta^*, \phi(x_t,a) \rangle - \sum_{t=1}^T r_t \right].
    \label{eq:regret}
\end{equation}

\subsection{Algorithm: Kernel-UCB}
Upper confidence algorithm (UCB) algorithms maintain for each possible action an estimate of the mean reward as well as a confidence interval around that mean, and then chooses at each time the highest upper confidence bound. Formally, if we have a confidence set
% $\mathcal{C}_t \subset \mathbb R^d$ 
$\mathcal{C}_t \subset \cH$ 
based on samples $(x_{t'}, a_{t'}, y_{t'})$, for $t' \in \{ 1, \ldots, t-1 \}$ that contains the unknown parameter vector $\theta^*$ with high probability, we may define
\begin{equation}
    \text{K-UCB}_t(a) = \max_{\theta \in \mathcal{C}_t} \langle \theta, \phi(x_t, a) \rangle
    \label{eq:ucb}
\end{equation}
as an upper bound on the mean pay-off $\langle \theta^*, \phi(x_t, a) \rangle$ of $a$. To choose the highest upper confidence bound from the confidence set at time $t$, the algorithm then selects:
\begin{equation}
    a_t \in \argmax_{a \in \mathcal{A}} \text{K-UCB}_t(a).
    \label{eq:argmax}
\end{equation}
We then build an empirical estimate of the unknown quantity $\theta^*$ using regression. More precisely in the kernelized setting, we use the regularized least square estimator with
\begin{equation}
    \hat{\theta}_t \in \argmin_{\theta \in \mathcal{H}} \Big\{ \sum_{s=1}^{t} \left( \langle \theta, \phi(x_s, a_s) \rangle - r_s \right)^2 + \lambda \Vert \theta \Vert^2 \Big\} \,.
    \label{eq:theta_hat}
\end{equation}
Rearranging the terms $\varphi_s = \phi(x_s, a_s)$ and writing $V_t = \sum_{s=1}^T \varphi_s \otimes \varphi_s + \lambda I$, we obtain that the analytical solution for Eq. \eqref{eq:theta_hat} is $\hat{\theta}_t = V_{t}^{-1} \sum_{s=1}^{t} \varphi_s r_s$. The previous solution from time $t-1$ then defines the center of the ellipsoidal confidence set
\begin{equation}
\mathcal{C}_t = \{ \theta \in \cH: \Vert \theta - \hat{\theta}_{t-1} \Vert_{V_{t-1}} \leq \beta_{t}(\delta) \}.
    \label{eq:ellispoid}
\end{equation}
where $\Vert \theta \Vert_{V}^2 = \theta^\top V \theta$,  and $\beta_t(\delta)$ is its radius (see Lemma~\ref{lemma:beta_delta}). With $\mathcal{C}_t$ in that form, we can write the solution of Eq. \eqref{eq:ucb} as
\begin{equation}
    \text{K-UCB}_t(a) = \langle \hat{\theta}_{t-1}, \phi(x_t, a)  \rangle + \beta_t(\delta)^{1/2} \Vert \phi(x_t, a) \Vert_{V_{t-1}^{-1}}.
\label{eq:solution_kucb}
\end{equation}
Indeed, by defining $B_2 = \{ x \in \mathbb R^d: \Vert x \Vert_2 \leq 1  \}$ the unit ball with the Euclidean norm, it is easy to see that $\mathcal{C}_t = \hat{\theta}_t + \beta_t(\delta)^{1/2} V_{t-1}^{-1/2}B_2$. Then, for $\theta \in B_2$ maximising the quantity $ \langle \theta, \phi(x_t, a) \rangle = \phi(x_t, a)^\top \hat{\theta}_{t-1} + \beta_t(\delta)^{1/2} \phi(x_t, a)^\top V_{t-1}^{-1/2} \theta$ gives Eq.~\eqref{eq:solution_kucb}.

\subsection{Regret analysis}
\label{sub:kucb_regret}

We provide an analysis of the regret of the kernelized UCB rule in Eq.~\eqref{eq:solution_kucb} using standard statistical analysis definitions of the effective dimension. 

% \subsubsection{Upper bound of the KernelUCB algorithm}

Let us write the operator $\Phi_t : \mathcal{H} \rightarrow \mathbb{R}^t$ such that $\Phi_t^* = \left[ \varphi_1, \dots \varphi_t \right]$, where $\varphi_i = \phi(x_i, a_i)$ for $i \in [1, t]$. Let us define $K_{t}$ the kernel matrix associated to kernel $k$ and the set of pairs $(x_1, a_1), \dots, (x_t, a_t)$, $K_{t} = \Phi_t \Phi_t^*$ is a $t \times t$ matrix.
% We note that $V_t = \Phi^\top \Phi + \lambda I_d$ where $d$ is the dimension of the feature map $\phi$ being a $d \times d$ matrix.
We define the effective dimension of a kernel matrix as in \cite{hastie01statisticallearning} and will use the following in our work. 

\begin{definition}
The effective dimension of the matrix $K_T$ is defined as, 
\begin{equation}
    \deff(\lambda,T) \defeq \Tr (K_T (K_T + \lambda I_T)^{-1}) \,.
    \label{eq:deff}
\end{equation}
\end{definition}
In what follows, for simplicity of notation, we abbreviate $\deff(\lambda,T)$ to $\deff$ unless we use different parameters on $\deff$. To extend the analysis of OFUL \citep{oful2011} to the contextual kernel UCB algorithm, we will use the following proposition that has been proved and used by~\citet{JezequelEtAl2019}. 

\begin{proposition}
 For any horizon $T \geq 1,  \lambda> 0$ and all input sequences $(x_1, a_1), \dots, (x_T, a_T)$ 
 \begin{equation*}
     \sum_{k=1}^{T} \log \left( 1+\dfrac{\lambda_k (K_T)}{\lambda} \right) \leq \log \left( e + \dfrac{e T \kappa^2}{\lambda}  \right) \deff \,,
 \end{equation*}
 where $\lambda_k (K_T)$ denotes the $k$-th largest eigenvalue of~$K_T$.
 \label{prop:d_eff}
\end{proposition}
We now provide a regret bound extending the analysis of \cite{oful2011}  to the kernel setting. In particular, we start by providing an upper bound on the ellipsoid greater axis.
\begin{lemma}
Let $\delta \in (0,1)$ and define $\beta_{t+1}(\delta)$ by
\begin{equation*}
     \sqrt{\lambda} \Vert \theta^* \Vert + \sqrt{2 \log\frac{1}{\delta} + \log \left( e + \dfrac{e t \kappa^2}{\lambda}  \right) \deff }.
\end{equation*}
Then, with probability at least $1-T\delta$, for all $t \in [T]$
\begin{equation}
    \big\Vert \hat{\theta}_t - \theta^* \big\Vert_{V_t} \leq \beta_{t+1}(\delta).
\end{equation}
\label{lemma:beta_delta}
\end{lemma}
We use this lemma (which relies on Proposition \ref{prop:d_eff} whose proof is in Appendix \ref{appx:regret_bound_proof}) to bound the distance between the estimated parameter $\hat{\theta}_t$ at each round~$t$ and the true parameter $\theta^*$. By combining this result with Proposition \ref{prop:d_eff}, we then prove the following theorem that extends the LinUCB upper bound result from \cite{lattimore_szepesvari_2020}. 

\begin{theorem}
Let $T\geq 2$ and $\theta^* \in \mathcal{H}$. Assume that $|\langle \phi(x,a), \theta^* \rangle| \leq 1$ for all $a \in \bigcup_{t=1}^{T} \mathcal{A}_t \subset \cA$ and $x \in \mathcal{X}$. Then, the K-UCB rule defined in Eq. \eqref{eq:argmax} for the choice $\mathcal{C}_t$ as in \eqref{eq:ellispoid} with parameter $\lambda >0$, and $\delta=1/T^2$, satisfies the pseudo-regret bound
\[
    R_T  \lesssim \sqrt{T} \Big(  \Vert \theta^* \Vert \sqrt{ \lambda  \deff}  + \deff \Big)   \,,
\]
where~$\lesssim$ hides logarithmic factors in~$T$.
\label{thm:regret_bound}

\end{theorem}

The proof of Theorem \ref{thm:regret_bound} and the precise statement of the regret bound are given in Appendix \ref{appx:regret_bound_proof}. 

In particular, assuming the norm of the true parameter~$\theta^\star$ to be bounded, we obtain the following corollary with a capacity condition on the effective dimension.
\begin{corollary}
Assuming the capacity condition $\deff \leq (T/\lambda)^{\alpha}$ for $0 \leq \alpha \leq 1$ , the regret of K-UCB is bounded as $R_{T} \lesssim T^{\frac{1+ 3\alpha}{2 + 2\alpha}}$ with an optimal $\lambda \approx T^{\frac{\alpha}{1+\alpha}}$.
\label{cor:bound_capacity_condition}
\end{corollary}

As an example, if we consider a kernel that is a tensor product between a linear kernel on contexts and a Sobolev-type kernel (\eg, a Matern kernel) of order~$s$ on actions, with~$s > d/2$ (where~$d$ is the dimension of the continuous action space), then we may consider that the kernel eigenvalues decay as~$i^{-2s/d}$, leading to an effective dimension as above with~$\alpha = d/2s$, and a regret of~$T^{\frac{1}{2} \frac{2s + 3d}{2s + d}}$.

\paragraph{Discussion.}
We note that this regret is not optimal for such problems, but matches the regret of most other kernel or Gaussian process optimization algorithms~\citep[see, \eg,][]{scarlett2017lower}. More precisely, our analysis recovers classical rates of the GP-UCB algorithm~\citep{krause2009,chowdhury2017kernelized}, and extends them to the contextual bandit setting. We note that the analysis of~\citet{chowdhury2017kernelized} further removes some logarithmic factors, and similar improvements may be obtained in our setting since it is based on similar tools.
The SupKernelUCB algorithm by~\citet{valko2013} obtains improved dependencies on~$T$ in the regret bounds, but requires a finite set of actions, and therefore is not directly comparable to ours.
The CGP-UCB algorithm by~\citet{krause2011contextual} obtains similar results to ours in the contextual setting, but uses a different analysis.
Our result is therefore not new, and our analysis is meant as a starting point for the efficient variant based on incremental Nystr\"om approximations, which will be introduced in the sequel.

We note that these works use different notions than our effective dimension~$\deff$ to characterize complexity, namely the information gain
\begin{equation*}
    \gamma(\lambda, t) = \dfrac{1}{2} \log \left(\det \left( I + \dfrac{1}{\lambda} K_t \right)\right)
\end{equation*}
used by~\citet{krause2011contextual} as well as the different effective dimension definition in~\citep{valko2013}
\begin{equation*}
    \tilde{d}(\lambda, t)~=~\min \{j:  j\lambda \log T \geq \sum_{k>j} \lambda_k(K_t) \}.
\end{equation*}
It can be shown that these are equivalent up to logarithmic factors to our definition of the effective dimension $\deff$ (see Appendix~\ref{app:comparison-regret}). This allows us to compare up to logarithmic factors the algorithm regrets, as shown in Table \ref{table:comparison_algos}.

\section{Efficient Kernel-UCB}
\label{sec:EKUCB}

In this section, we introduce our efficient kernelized UCB (EK-UCB) algorithm based on incremental Nystr\"om projections.
We begin by extending the ellipsoidal confidence bounds from the previous section to the case with projections on finite-dimensional linear subspaces of the RKHS. Then, we present our main algorithm and analyze its complexity and regret.

% In this paper, we consider a variation of the Kernelized UCB algorithm that we call Efficient Kernelized UCB (EK-UCB). 

\subsection{Upper confidence bounds with projections}
\label{sec:conf_proj}

In this section, we study the UCB updates and corresponding high-probability confidence bounds for our EK-UCB algorithm.
Because these steps do not depend on a specific choice of projections, we consider generic projection operators onto subspaces of the RKHS, noting that the next sections will consider specific choices based on Nystr\"om approximations.

At round~$t \geq 1$, we consider a generic subspace~$\tilde{\mathcal H}_t$ of~$\mathcal H$, and let~$P_t: \mathcal H \rightarrow \tilde{\mathcal H}_t$ be the orthogonal projection operator on~$\tilde{\mathcal H}_t$, so that $P_t \mathcal H = \tilde{\mathcal H}_t$.
% Consider $\tilde{\mathcal H}_t$ an explicit subspace of the projection $P_t: \mathcal H \rightarrow \tilde{\mathcal H}_t$ the euclidean projection on $\tilde{\mathcal H}_t$. We have that $\tilde{\mathcal H}_t = \{ P_t \theta, \theta \in \mathcal H \}$.
For a fixed regularization parameter $\lambda > 0$, we consider the following regularized estimator restricted to~$\tilde{\mathcal H}_t$:
\begin{equation}
    \tilde{\theta}_t \in \argmin_{\theta \in \tilde{\mathcal H}_t} \bigg \{ \sum_{s=1}^{t} \left( \langle \theta, \phi(x_s, a_s) \rangle - r_s \right)^2 + \lambda \Vert \theta \Vert^2 \bigg \}.
    \label{eq:theta_tilde}
\end{equation}
Define $\tilde{V}_t = \sum_{s=1}^{t} P_t \varphi_s \otimes P_t \varphi_s  + \lambda I$, which may be written $\tilde{V}_t = P_t F_t P_t + \lambda I$ where $F_t = \Phi_t^* \Phi_t:\mathcal H \to \mathcal H$ is the covariance operator. Recalling the notation $Y_t = (r_1, \dots, r_t)^\top$, we obtain that $\tilde{\theta}_t = \tilde{V}_t^{-1} P_t \Phi_t^* Y_t$.
We may then define the following ellipsoidal confidence set:
\begin{equation}
\tilde{\mathcal{C}}_t \defeq \big\{ \theta \in  \cH: \Vert \theta - \tilde{\theta}_{t-1} \Vert_{\tilde{V}_{t-1}} \leq \tilde{\beta}_t(\delta) \big\} \,,
    \label{eq:ellispoid_tilde}
\end{equation}
for some radius~$\tilde \beta_t(\delta)$ to be specified later.
% Here, the norm uses the geometry induced by the operator $\tilde{V}_{t-1}$, which accounts for exploring the joint context-action feature space with the projected information gathered from previous rounds.
Note that the ellipsoid is not necessarily contained inside the projected space $\tilde {\mathcal{H}}_t$, and may in fact include~$\theta^*$ even if~$\theta^* \notin \tilde{\mathcal H}_t$.
This is a crucial difference with random feature kernel approximations~\citep{mutny2019efficient}, for which a standard confidence set would be finite dimensional, and thus generally does not include~$\theta^*$; this leads to larger regret due to misspecification, unless the number of random features is very large in order to ensure good \emph{uniform} approximation.
We may then define the following upper confidence bounds, which still rely on the original feature map~$\phi$:
\begin{equation}
    \text{EK-UCB}_t(a)  \defeq \max_{\theta \in \tilde{\mathcal{C}}_t} \langle \theta, \phi(x_t, a) \rangle.
    \label{eq:ek_ucb}
\end{equation}
This may again be written in closed form as
\begin{equation*}
    \text{EK-UCB}_t(a) = \langle \hat{\theta}_{t-1}, \phi(x_t, a)  \rangle + \tilde \beta_t(\delta)^{1/2} \Vert \phi(x_t, a) \Vert_{\tilde{V}_{t-1}^{-1}}.
    \label{eq:solution_ekucb}
\end{equation*}

We note that for appropriate choices of~$\tilde{\mathcal H}_t$, such a quantity can be explicitly computed using the kernel trick, as we discuss in Section~\ref{sub:ekucb_complexity}. 
% Here we did not use the projected feature map in the EK-UCB rule. First, the use of the original feature map allows to smaller great axis of the ellipsoids built by the UCB algorithm, and to alleviate linear regret term in the analysis of the mispecified term induced by the residual $(I-P_t)\theta^*$ (see Appendix \ref{app:proofEKUCB} for discussions). Second, its use does not hurt the complexity of the algorithm in its implementation, using the kernel trick. 
The following lemma shows that~$\tilde {\mathcal C}_t$ is a valid confidence set, which contains~$\theta^*$ with high probability, provided that the projection captures well the dominating directions in the covariance operator.
% The following lemma serves to compute the distance of the center $\tilde{\theta}_t$ to any point in the ellipsoid in the projected space $\tilde{\mathcal H}_t$. Note that the norm uses the geometry induced by the direction matrix $\tilde{V}_t$, which accounts for exploring the joint context-action feature space with the projected information gathered from previous rounds.
% We write $\mu_t \defeq \Vert (I-P_t) F_t^{1/2} \Vert^2$ and the projected gram matrix $\tilde{K}_t \defeq \Phi_t P_t \Phi_t^\top P_t$. 
\begin{lemma}
Let $\delta \in (0,1)$. Define $\tilde{\beta}_{t+1}(\delta)$ as
% , and $P_t$ incremental projections, and let us write
\begin{equation*}
     \left(\sqrt{\lambda} +\sqrt{\mu_t}  \right) \Vert \theta^* \Vert + \sqrt{4 \log\frac{1}{\delta} + 2 \log \left( e + \dfrac{e t \kappa^2}{\lambda}  \right) \deff}, 
\end{equation*}
where~$\mu_t \defeq \Vert (I-P_t) F_t^{1/2} \Vert^2$.
Then, with probability at least $1-T\delta$, for all $t  \in [T]$
\begin{equation}
    \big\Vert \tilde{\theta}_t - \theta^* \big\Vert_{\tilde V_t} \leq \tilde {\beta}_{t+1}(\delta).
\end{equation}
\label{lemma:beta_delta_tilde}
\end{lemma}
The quantity~$\mu_t$ controls how well the projection operator~$P_t$ captures the dominating eigen-directions of the covariance operator, and should be at most of order~$\lambda$ in order for the confidence bounds to be nearly as tight as for the vanilla K-UCB algorithm. The next section further discusses how this quantity is controlled with incremental Nystr\"om projections.

% Note that this lemma requires the orthogonal projections $P_t$ to be incremental, that is $P_t - P_{t-1} \succeq 0$. As for the projections, we propose in this work to leverage Nystr\"om incremental approximations to build an efficient contextualized kernel UCB algorithm. 

\subsection{Learning with incremental Nystr\"om projections}

We now consider specific choices of the projections~$P_t$ and subspaces~$\tilde{\mathcal H}_t$ obtained by Nystr\"om approximation~\citep{williams2001using,alessandro2015}.
In particular, the spaces~$\tilde{\mathcal H}_t$ now take the form
% The Nystr\"om approximations of kernel feature maps are useful to derive finite dimensional embeddings \citep{williams2001using}, are suited for any RKHS and are data dependant. Nystr\"om projections \citep{alessandro2015} consists in sequentially updating a dictionary  and using the projected space
\begin{equation}
    \tilde{\mathcal{H}}_t = \text{Span} \big \{ \phi(s), s \in \mathcal{Z}_t \big \},
\end{equation}
where~$\mathcal{Z}_t \subset \{ (x_1, a_1), \dots (x_t, a_t) \}$ is a dictionary of anchor points taken from the previously observed data.
Our approach consists of constructing the dictionaries~$\mathcal Z_t$ \emph{incrementally}, by adding new observed examples~$(x_t, a_t)$ on the fly when deemed important, so that we have~$\mathcal{Z}_1 \subset \mathcal{Z}_2 \dots \subset \mathcal{Z}_t$.
% Anchor points in $\mathcal{Z}_t$ that are used to span a subspace of the original RKHS $\mathcal{H}$ can be chosen in different ways.
We achieve this using the Kernel Online Row Sampling (KORS) algorithm of~\citet{calandriello2017}, shown in Algorithm~\ref{alg:kors}, which decides whether to include a new sample~$s_t = (x_t, a_t)$ by flipping a coin with probability proportional to its \emph{leverage score}~\citep{mahoney2009cur}.
% incremental dictionaries are $\mathcal{Z}_1 \subset \mathcal{Z}_2 \dots \subset \mathcal{Z}_t$ by evaluating at each time $t$ a leverage score $\tilde{\tau}_{t}$ of each sample $s_t = (x_t, a_t)$ to compute a probability $\tilde{p}_t$ so as to draw a Bernouilli random variable to include it or not in the dictionary.
% In our method, we propose to use the KORS algorithm (see Alg. \ref{alg:kors}) at each round to perform incremental approximations of the RKHS and build an efficient algorithm. 
More precisely, an estimate  $\tilde \tau$ of the leverage score that uses the state feature $\varphi_t$ and parameters $\mu, \epsilon$ is used to assess how a given state is useful to characterize the dataset. More details on the KORS algorithm are given in Appendix~\ref{app:implementation}.
\begin{algorithm}[t]
\SetAlgoLined
\KwIn{Time $t$, past dictionary $\mathcal{Z}$, context-action $s_t$, regularization $\mu$, accuracy $\epsilon$, budget $\gamma$}

%  Construct virtual dictionary $\mathcal{Z}_{t, *} = \mathcal{Z} \cup \{s_t\}$\;
 Compute the leverage score $\tilde{\tau}_{t}$ from $\mathcal Z, s_t, \mu, \epsilon$ \;
 Compute $\tilde{p}_t = \min \{ \gamma \tilde{\tau}_{t}, 1 \}$ \;
 Draw $z_t \sim \mathcal{B}(\tilde{p}_t)$ and if $z_t =1$, add $s_t$ to $\mathcal{Z}$\;
 \KwResult{Dictionary $\mathcal{Z}$}
 \caption{Incremental KORS subroutine}
 \label{alg:kors}
\end{algorithm}

We state the following proposition of \citet[Theorem 1, with~$\epsilon = 1/2$]{calandriello2017},
which will be useful for our regret and complexity analyses.
% We recall that $\Vert \Phi^* (I-P_t) \Phi^* \Vert = \Vert (I-P_t) F_t^{1/2} \Vert^2 \eqdef \mu_t$. 

\begin{proposition}
 Let $\delta >0$, $n \geq 1$, $\mu > 0$. Then the sequence of dictionaries $\mathcal{Z}_1 \subset \mathcal{Z}_2 \subset \dots \mathcal{Z}_T$ learned by KORS with parameters $\mu>0, \epsilon=1/2$ and $\gamma=12 \log(T/\delta)$ satisfies with probability $1-\delta$, $\forall t \geq 1$
 \begin{equation*}
     \Vert (I-P_t) F_t^{1/2} \Vert^2 \leq \mu \ \text{ and } \ |\mathcal{Z}_t| \leq 9 \deff(\mu, T) \log(2T/ \delta)^2 \,.
 \end{equation*}
 Additionally, the algorithm runs in $\cO(\deff(\mu, T)^2)$ time and $\cO(\deff(\mu, T)^2 \log(T)^4)$  space  per iteration.
 \label{prop:kors}
\end{proposition}

This result shows that when choosing~$\mu \approx \lambda$, then KORS will maintain dictionaries of size at most~$\deff$ (up to log factors), while guaranteeing that the confidence bounds studied in Section~\ref{sec:conf_proj} are nearly as good as for the case of K-UCB. 

\subsection{Implementation and complexity analysis}
\label{sub:ekucb_complexity}

Here, we analyze the complexity of the algorithm and describe its practical implementation. Recall that at each round $t$ the agent chooses an action $a$ that maximises the UCB rule $\mu_{t, a} + \tilde \beta_t \sigma_{t, a}$ where we use Eq. \eqref{eq:ek_ucb} to reformulate the mean term $\mu_{t, a} = \langle \hat{\theta}_{t-1}, \phi(x_t, a)  \rangle$ and the variance term $\sigma_{t, a}^2 = \Vert \phi(x_t, a) \Vert_{\tilde{V}_{t-1}^{-1}}^2$. We use the representer theorem on the projection space $\mathcal{H}_t$ to derive efficient computations of the latter two terms instead of using a kernel trick with $t \times t$ gram matrices. Indeed, in the next proposition, we prove that the two terms can be expressed with $m_t \times m_t$ matrices instead, where $m_t = |\cZ_t|$ is the size of the dictionary at time $t$.
% We provide a pseudo-code of our method in Alg.~\ref{alg:ek_ucb} and provide the following proposition on the complexity of the algorithm that we detail thoroughly in Appendix \ref{app:implementation}.
We use the notations~$K_{\mathcal{S}_t}(s')$ for the kernel column vector $[k(s_1, s'), \dots, k(s_t, s')]^\top$, where $\mathcal{S}_t=\{ s_i \}_{i=1 \dots t}$ are the past states, and $K_{\mathcal A, \mathcal B}$ for the matrix of kernel evaluations $[k(s, s')]_{s \in \mathcal A, s' \in \mathcal B}$.
% Here,  $K_{\mathcal{S}_t}(s')$ is the kernel column vector $[k(s_1, s'), \dots, k(s_t, s')]^\top$ where $\mathcal{S}_t=\{ s_i \}_{i=1 \dots t-1}$ are the past states. Note that $K_{\cS_t}(s) = \Phi_t \phi(s)$. Similarly, \cite{valko2013} uses the same formulations of the mean and standard deviation $\sigma_t^2(s)$ in their Kernel UCB subroutine used in the phased UCB rules of their finite arms. While inverting the full matrices would induce as full cost of $\mathcal{O}(CT^4)$, using first order updates with Schur complement allows to run the algorithm in $\mathcal{O}(CT^3)$, while using $\mathcal{O}(T^2)$ in space. 

\begin{proposition}
 At any round $t$, by considering $s_{t,a}=~(x_t, a)$, the mean and variance term of the EK-UCB rule can be expressed with:
%  \begin{align*}
%     \Gamma_t &= K_{\cZ_{t}\cS_{t-1}} Y_{t-1} \\
%     \Lambda_t &= \left(K_{\cZ_{t} \cS_{t-1}} K_{\cS_{t-1}\cZ_{t}} + \lambda K_{\cZ_{t}\cZ_{t}}\right)^{-1} \\
%     \mu_{t, a} &= K_{\cZ_{t}}(s_{t,a})^\top \Lambda_t \Gamma_t \\
%     \Delta_{t, a} &= K_{\cZ_{t}}(s_{t,a})^\top \left( \Lambda_t  - \frac{1}{\lambda} K_{\cZ_{t}\cZ_{t}}^{-1} \right) K_{\cZ_{t}}(s_{t,a})
% \\
%     \sigma_{t, a}^2 &= \frac{1}{\lambda} k(s_{t,a}, s_{t,a}) + \Delta_{t, a}.
% \end{align*}
 \begin{align*}
    \Gamma_t &= K_{\cZ_{t-1}\cS_{t-1}} Y_{t-1} \\
    \Lambda_t &= \left(K_{\cZ_{t-1} \cS_{t-1}} K_{\cS_{t-1}\cZ_{t-1}} + \lambda K_{\cZ_{t-1}\cZ_{t-1}}\right)^{-1} \\
    \tilde \mu_{t, a} &= K_{\cZ_{t-1}}(s_{t,a})^\top \Lambda_t \Gamma_t \\
    \Delta_{t, a} &= K_{\cZ_{t-1}}(s_{t,a})^\top \left( \Lambda_t  - \frac{1}{\lambda} K_{\cZ_{t-1}\cZ_{t-1}}^{-1} \right) K_{\cZ_{t-1}}(s_{t,a})
\\
    \tilde \sigma_{t, a}^2 &= \frac{1}{\lambda} k(s_{t,a}, s_{t,a}) + \Delta_{t, a}.
\end{align*}
The algorithm then runs in a space complexity of $\mathcal{O}(Tm)$ and a time complexity of $\mathcal{O}(CTm^2)$.
\label{prop:ek_ucb_implementation}
\end{proposition}

\begin{algorithm}[t]
\SetAlgoLined
\KwIn{$T$ the horizon, $\lambda$ regularization and exploration parameters, $k$ the kernel function, $\epsilon > 0, \gamma > 0$}
% \KwResult{Write here the result }
 Initialization\;
 Context $x_0$, $a_0$ chosen randomly and reward $r_0$ \;
 $\mathcal{S} = \{(x_0, a_0) \}, Y_\mathcal{S} =[r_0]$ $\mathcal{Z} = \{ (x_0, a_0) \}$ \;
 $\Lambda_t = \left(K_{\mathcal{Z} \mathcal{S}}   K_{\mathcal{S}\mathcal{Z}} + \lambda K_{\mathcal{Z}\mathcal{Z}}\right)^{-1}$ 
 $\Gamma_t =  K_{\mathcal{Z} \mathcal{S}}Y_\mathcal{S}$ \;
%  $\Xi = K_{\mathcal{Z}'\mathcal{Z}'}^{-1}$
 
 \For{$t=1$ to $T$}{
 Observe context $x_t$ \; 
 Choose $\tilde{\beta}_t$ (e.g as in Lem. \ref{lemma:beta_delta_tilde}, and $\delta=\frac{1}{T^2}$) \;
Choose $a_t \leftarrow \argmax_{a \in \mathcal{A}} \tilde{\mu}_{t, a} + \tilde{\beta}_t \tilde{\sigma}_{t,a}$ \;
 \Indp{$\tilde{\mu}_{t, a} \leftarrow K_{\mathcal{Z}}(s_{t,a})^\top \Lambda_t \Gamma_t$ \;
 $\Delta_{t,a} = K_{\cZ}(s_{t,a})^\top \left( \Lambda_t  - \frac{1}{\lambda} K_{\cZ\cZ}^{-1} \right) K_{\cZ}(s_{t,a})$ \
 $\tilde{\sigma}_{t,a}^2 \leftarrow \frac{1}{\lambda} k(s_{t,a}, s_{t,a}) + \Delta_{t, a}$ \;}

 \Indm
 Observe reward $r_t$ and $s_t \leftarrow (x_t, a_t)$ \;
 $Y_\mathcal{S} \leftarrow [Y_\mathcal{S}, r_t]^\top, \mathcal{S} \leftarrow \mathcal{S} \cup \{ s_t \}$ \;
 $\mathcal{Z}' \leftarrow \text{KORS}(t, \mathcal{Z}, K_{\mathcal{Z}}(s_t), \lambda, \epsilon, \gamma )$ \;
 \If{$\mathcal{Z}' = \mathcal{Z}$ }{
 Incremental inverse update $\Lambda_t$ with $s_t$\;
 $\Gamma_{t+1} \leftarrow \Gamma_t + r_t K_{\mathcal{Z}}(s_t)$ \;
 }
 \Else{
 $z = \mathcal{Z}' \ \backslash \ \mathcal{Z}$ \;
 Incremental inverse update $\Lambda_t$ with $s_t, z$ \;
 Incremental inverse update $K_{\cZ\cZ}^{-1}$ with $z$\;
 $\Gamma_{t+1} \leftarrow [\Gamma_t + r_t K_{\mathcal{Z}}(s_t), \  K_{\mathcal{S}}(z)^\top Y_\mathcal{S}]^\top$
 }
 }
 \caption{Efficient Kernel UCB}
 \label{alg:ek_ucb}
\end{algorithm}

% In Appendix \ref{app:implementation} we prove the following expressions and detail all the pseudo-codes for practical implementations. 

In our algorithm, the incremental updates of the projections allow us to derive rank-one updates of the expressions $\Lambda_t, \Gamma_t, K_{\cZ_{t}\cZ_{t}}^{-1}$ in all cases. First, when the dictionary does not change (i.e $P_t = P_{t-1}$), the update of the $m_t \times m_t$ matrix $\Lambda_t$ can be performed with Sherman-Morrison updates, and the term $\Gamma_t = K_{\cZ_{t-1}\cS_{t-1}} Y_{t-1}$ can also benefit from a rank-one update given the latest reward and state. Both updates are performed in no more than $\mathcal{O}(m_t^2)$ time and space. Second, when the dictionary changes (i.e $P_t \succ P_{t-1}$), the matrix $\Lambda_t$ can be updated in two stages with a rank-one update using Sherman-Morrison on the states as if the dictionary did not change, in $\mathcal{O}(m_t^2)$ time and space, and second rank-one update on the dictionary using the Schur complement in $\mathcal{O}(t m_t+m_t^2)$ time and space. Similarly, we can update $\Gamma_t =K_{\cZ_{t}\cS_{t-1}} Y_{t-1}$ with a first update on the states and stacking a block of size $1 \times t$ in $\mathcal{O}(t m_t)$ space and time. Eventually, the inverse of the dictionary gram matrix $K_{\cZ_{t}\cZ_{t}}^{-1}$ is updated with Schur complement in $\mathcal{O}(m_t^2)$. Besides, the second case when the projection is updated occurs at most $m$ times and the first case at most $T$ times. When the UCB rule is computed on $C$ discrete actions or when we assume that it can be optimized using $\mathcal{O}(C)$ evaluations, given that the KORS algorithm runs in $\mathcal{O}(m^2)$ time and space, our algorithm has a total complexity of $\mathcal{O}(CT \deff ^2)$ in time and $\mathcal{O}(T \deff)$ in space, using that $m \approx \deff$. 
Note that, as in all UCB algorithms, including ours, the theoretical value for $\tilde{\beta}_t$ in Lemma~\ref{lemma:beta_delta_tilde} is hard to estimate and often too pessimistic and leads to over-exploration, as discussed by~\citet{calandriello20a}. In practice, choosing a fixed value has shown to perform well in our experiments.
%Note also that the computation of $\tilde{\beta}_t$ with $\delta= 1/T^2$ is possible with upper-bounding the approximation errors $\mu_t$ with the KORS parameter~$\mu$, and by estimating $\deff$ with $\lambda = \mu$ and then taking $\deff \approx m$. In practice, choosing a fixed value for $\tilde{beta}_t$ seems to perform well in our experiments.

In contrast, the non-incremental approach of~\citet{calandriello20a} in the BBKB algorithm needs to recompute a new dictionary about $\deff$ times. Each update involves the computation of a new covariance matrix $K_{\cZ \cS} K_{\cS \cZ}$ which costs $\mathcal{O}(t m_t^2)$ operations for its contextual variant\footnote{The original BBKB algorithm does not involve contexts and consider a finite set of actions, allowing to compute the covariance matrix in $\mathcal{O}(\min(t,|\mathcal A|) m_t^2)$.}, yielding an overall $\mathcal{O}( T \deff^3)$ with $m \approx \deff$, as illustrated in Table~\ref{table:comparison_algos}.

\subsection{Regret analysis}

We now analyze the regret of the EK-UCB algorithm, using Proposition \ref{prop:kors} as well as Lemma \ref{lemma:beta_delta_tilde}.

\begin{theorem}
Let $T\geq 1$ and $\theta^* \in \cH$. Assume that $|\langle \phi(x,a), \theta^* \rangle| \leq 1$ for all $a \in \bigcup_{t=1}^{T} \mathcal{A}_t \subset \cA$ and $x \in \mathcal{X}$. 
% Then the EK-UCB rule in Eq.~\eqref{eq:solution_ekucb} with $\tilde{\mathcal{C}}_t$ defined in Eq.~\eqref{eq:ellispoid_tilde} with parameter $\lambda >0$ and a dictionary learnt by KORS as in Prop.~\ref{prop:kors}, with $m\defeq |\mathcal{Z}_T|$ dictionary updates, satisfies the pseudo-regret bound 
Then, the EK-UCB algorithm with regularization~$\lambda$ along with KORS updates with parameter~$\mu$ satisfies the regret bound
\begin{align*}
    % R_T \leq \sqrt{T} \left[  \dfrac{\sqrt{2m \mu}}{\lambda}  +  \sqrt{ \left( \log \left( e + \dfrac{e T \kappa^2}{\lambda}  \right) \deff \right)} \right] \\
    % \left[ (\sqrt{\lambda} + \sqrt{\mu}) \Vert \theta^* \Vert + \sqrt{2 \log (T) +   \log \left( e + \dfrac{e T \kappa^2}{\lambda}  \right) \deff} \right] 
    R_T \lesssim \sqrt{T} \left( \sqrt{\dfrac{\mu m}{\lambda}} \!+\! \sqrt{\deff} \right) \left( \|\theta^*\| (\sqrt{\lambda} \!+\! \sqrt{\mu}) \!+\! \sqrt{\deff} \right),
\end{align*}
where~$m\defeq |\mathcal{Z}_T|$.
In particular, the choice $\mu = \lambda$ yields $m \lesssim \deff$ and the bound
\[
    R_T \lesssim \sqrt{T}  \big( \|\theta^*\| \sqrt{\lambda \deff }  + \deff \big) \,.
\]
Furthermore, the algorithm runs in $O(Tm)$ space complexity  and $O(CTm^2)$ time complexity.
\label{thm:regret_bound_projected}
\end{theorem}

The regret bound is again given up to logarithmic factors and we detail the proof as well as the precise bound in Appendix \ref{app:proofEKUCB}.
As for K-UCB, one may analyze the resulting regret under a capacity condition, and when~$\mu \approx \lambda$, we obtain the same guarantees as in Corollary~\ref{cor:bound_capacity_condition}.
Note that our analysis leverages the fact that the dictionary is constructed incrementally, in particular using a condition~$P_t \succeq P_{t-1}$, which yields the approximation term~$\sqrt{\mu m / \lambda}$.
% Note that in our analysis, we also used the incremental assumption on projections $P_t \succeq P_{t-1}$ which allowed us to derive the second approximation term $\sqrt{{\mu m}/{\lambda}}$.
Had we used fixed projections with some operator~$P$, this approximation term would instead be~$\sqrt{\mu/\lambda}$ with~$\mu = \|(I - P) F_T^{1/2}\|^2$.

As a consequence of this theorem, the following corollary analyzes when the approximation terms dominate the regret, i.e when the dictionary size does not suffice to recover the original regret bound.

\begin{corollary}
Assuming the capacity condition $\deff \leq (T/\lambda)^{\alpha}$ for $0 \leq \alpha \leq 1$ . Let $m \geq 1$, under the assumptions of Thm.~\ref{thm:regret_bound_projected}, the regret of EK-UCB satisfies
\begin{equation*}
    R_{T} \lesssim \left\{
    \begin{array}{ll}
         & T m^{\frac{\alpha-1}{2\alpha}} \mbox{ if } m \leq T^{\frac{\alpha}{1+\alpha}} \\
         & T^{\frac{1+3\alpha}{2+2\alpha}} \mbox{ otherwise}
    \end{array}
\right.
\end{equation*}
for the choice $\lambda = \mu = T m^{-1/\alpha}$. 
\label{cor:bound_capacity_condition_projected}
\end{corollary}

\begin{figure*}[h!]
    \centering
\begin{subfigure}
  \centering
    \includegraphics[height=0.3\linewidth]{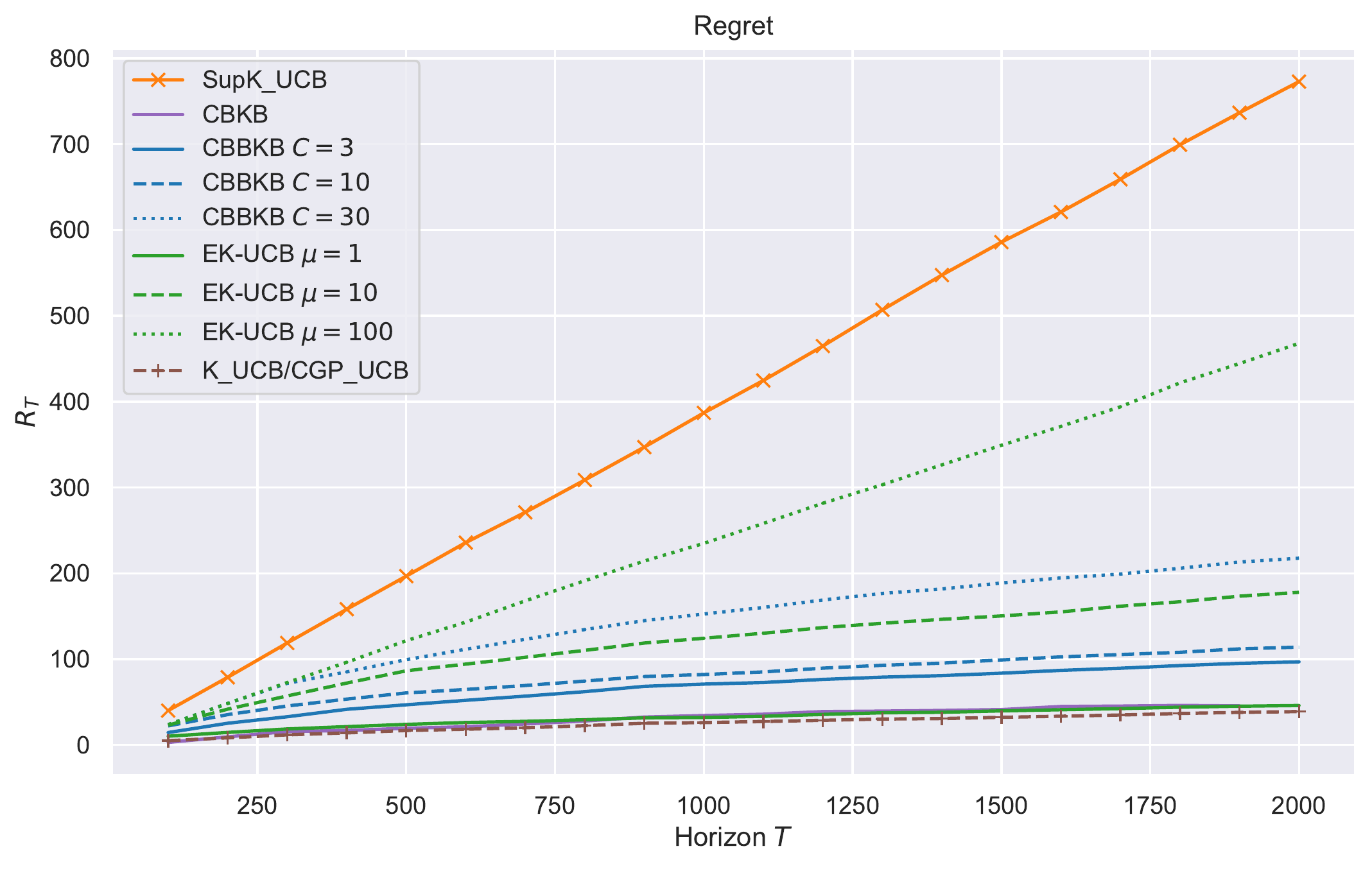}
  \label{fig:regret}
\end{subfigure}
\begin{subfigure}
  \centering
    \includegraphics[height=0.3\linewidth]{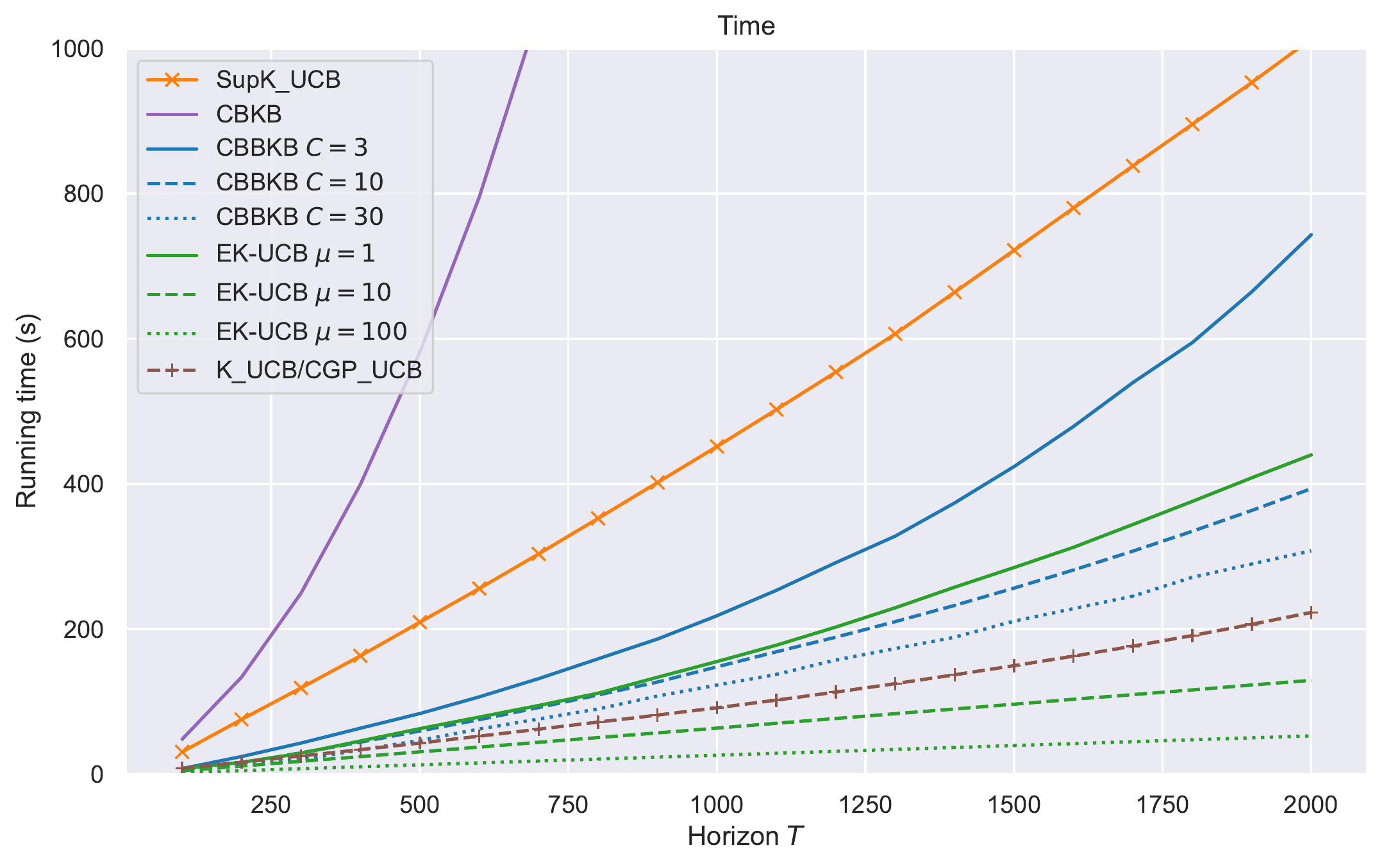}
  \label{fig:time}
\end{subfigure}

\caption{'Bump' setting: Regret and running times of EK-UCB, CBBKB, CBKB, SupKUCB and K-UCB, with $T=1000$ and $\lambda=10$ (see Corollary \ref{cor:bound_capacity_condition} and \ref{cor:bound_capacity_condition_projected}). EK-UCB matches the best theoretical regret-time compromise when the projection parameter $\mu=\lambda$. We show other values of $\mu$: higher $\mu$ ($\mu=100$) leads to faster computational time but worse regret, and reciprocally ($\mu=1$) leads to worse computational time and better regret. Additional results where $\lambda$ and $\mu$ change simultaneously are available in the Appendix \ref{app:experimental_details}.}
\label{fig:experiment}
\end{figure*}

The proof is postponed to Appendix~\ref{app:proofEKUCB}. In a practical setting, the dictionary size is controlled  by the choice of the projection parameter $\mu$. When $\mu$ is too high, it induces a smaller dictionary size  $m$ but thus linear regret as indicated in the previous corollary. However, by choosing a low $\mu$, we still recover the original regret but increase the size of the dictionary and thus pay a higher computation time. To recover the original regret, the regularization parameter $\lambda$ must be set to $\mu$ in all cases to recover the original regret, and both values have a theoretical optimal value which depends on the horizon to recover the best convergence rate under the capacity condition assumption. 

% Note that for clarity, we used the capacity condition assumption to illustrate the convergence rate of the algorithm, but the same analysis would apply to recover the original regret with the condition $\mu \approx \lambda$. We prove this point in Appendix \ref{app:proofEKUCB}. As a matter of fact, what is notable in this corollary is that the algorithm requires to choose $\mu \approx \lambda$, because it will generate a dictionary of large size when choosing a low $\mu$ KORS parameter, and reciprocally. All algorithm implementations were carefully optimized.

\section{Numerical Experiments}
\label{sec:exps}

We now evaluate our proposed EK-UCB approach empirically on a synthetic scenario, in order to illustrate its performance in practice. All algorithms  have been carefully optimized for fair comparisons.\footnote{The code with open-source implementations for experimental reproducibility is available at \codeurl.}  More experimental details, discussions, and additional experimental results are provided in Appendix~\ref{app:experimental_details}.

\paragraph{Experimental setup.}
We consider a 'Bump' synthetic environment with contexts uniformly distributed in~$[0, 1]^p$, with~$p=5$, and actions in~$[0, 1]$. The rewards are generated using the
% The synthetic environment is made of a context-action space $\cX \ =[0, 1]^p$ (we take $p=5$ and finite arms for $\cA$). The agent has to learn to take actions $a$ that maximize the reward
function~$r(x,a) =\max(0, 1 - \Vert a - a^* \Vert_1 - \langle w^*, x - x^* \rangle)$ for some $a^*, w^*$ and $x^*$ picked randomly and fixed. We also consider additional 2D synthetic settings 'Chessboard' and 'Step Diagonal' presented in Appendix \ref{app:additional-settings}. We use a Gaussian kernel in this setting.
% Unlike smooth environments where the choice of anchor points for the dictionary is less determinant, for e.g using fixed projections, we use an environment that requires finer inducing point strategies as in real-world applications. 
We run our algorithms for $T=2000$ steps and average our results over different 3 random runs. 
%We provide the code for reproducibility in the supplementary materials. 

\paragraph{Baselines.} In our experiments, we chose to compare to K-UCB, SupK-UCB and to works which focus on improving the $\mathcal{O}(T^3)$ time-complexity for the kernel case. We implemented K-UCB, SupK-UCB (SupKernelUCB, \cite{valko2013}), EK-UCB (our efficient version of the K-UCB algorithm) as well as our contextual adaptation of the BKB~\citep{calandriello19a} and BBKB~\citep{calandriello20a} algorithms; we will refer to these respectively as CBKB and CBBKB. Specifically, we use the same accumulation criteria as~\citet{calandriello20a} for the ``resparsification'' strategies (i.e., the resampling of the dictionary) with a threshold parameter~$C$. We also proceed to the same sampling and equation updates as the original algorithms while using our joint kernel on context-action pairs. Note also that CGP-UCB/K-UCB only differ from their parameter $\beta_t$ and match the same algorithm in our implementation (see second last paragraph in Sec. 4.3).

\begin{figure*}[h!]
    \centering
\begin{subfigure}
  \centering
    \includegraphics[height=0.3\linewidth]{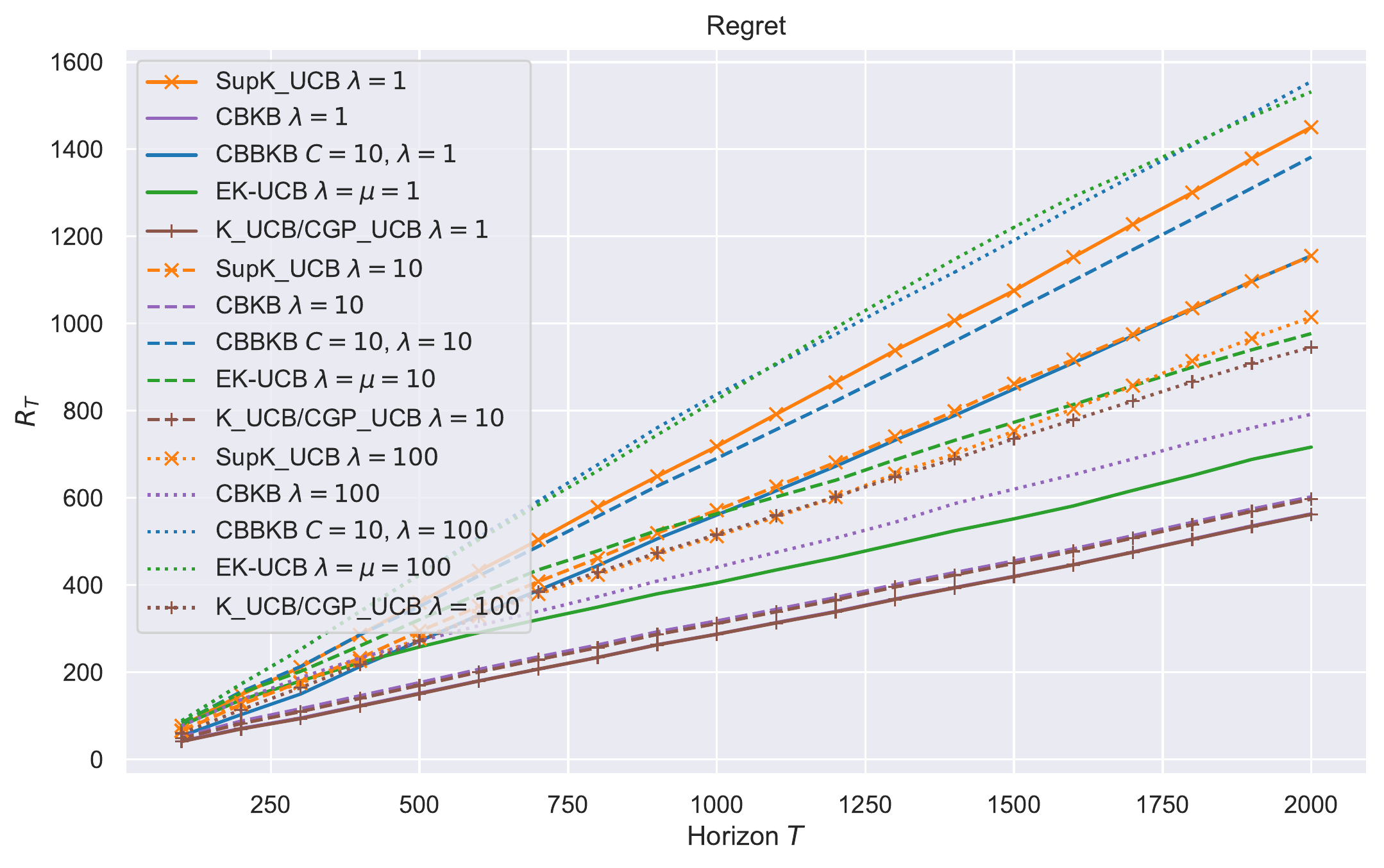}
  \label{fig:regret2}
\end{subfigure}
\begin{subfigure}
  \centering
    \includegraphics[height=0.3\linewidth]{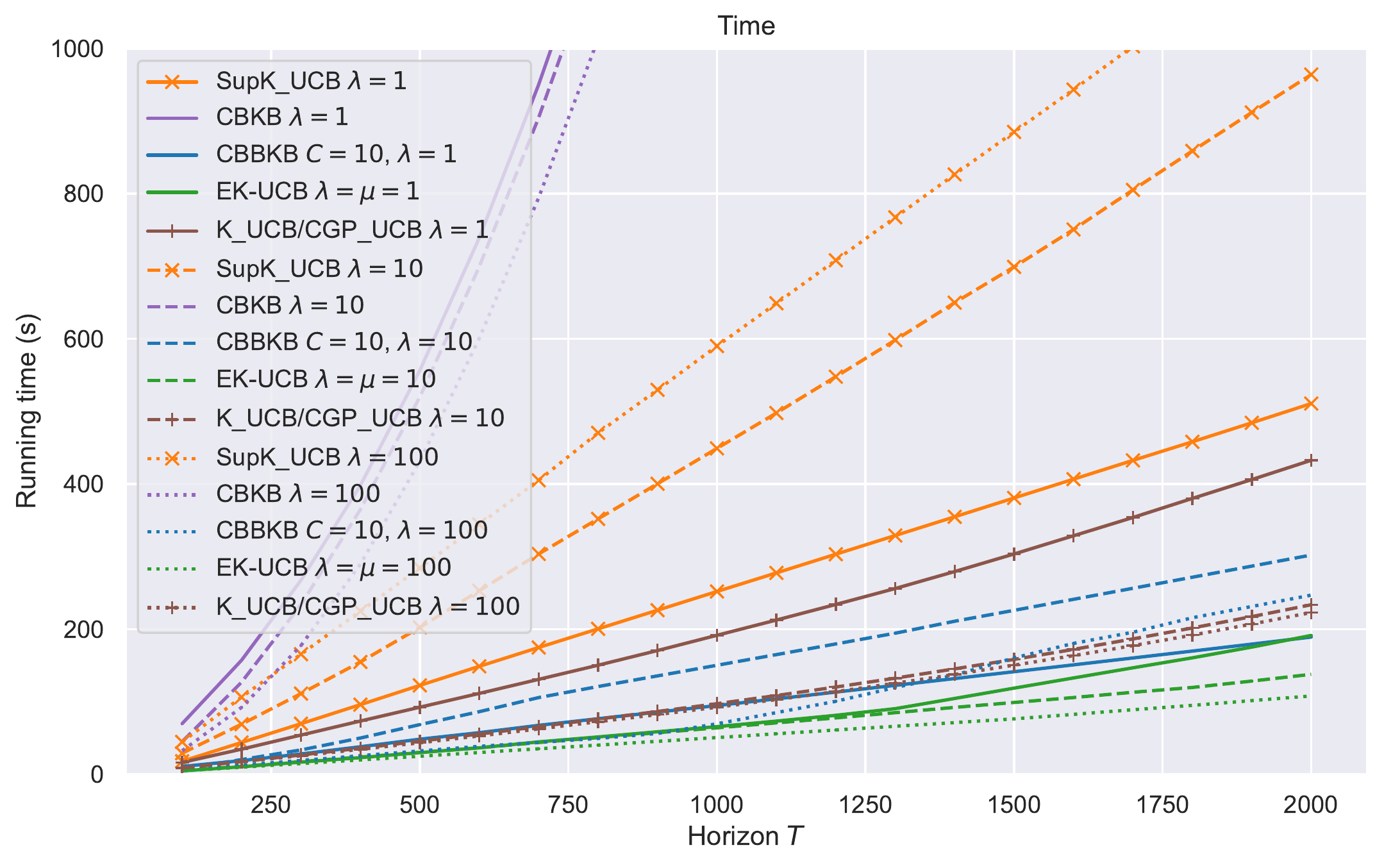}
  \label{fig:time2}
\end{subfigure}

\caption{'Chessboard' setting: Regret and running times of EK-UCB ($\lambda=\mu$), CBBKB ($C=10$) and CBKB, with $T=2000$ and  with varying $\lambda$. We notice that low $\lambda$ values have better regrets but higher computatinal times. Overall EK-UCB achieves the best regret-time compromise for all parameters of $\lambda$ while CBBKB sometimes improves upon the K-UCB complexity but has both higher regret than EK-UCB and higher computational time.}
\label{fig:experiment2}
\end{figure*}

\paragraph{Results.} We report the average regret and running times of the algorithms over different runs in Fig. \ref{fig:experiment} and Fig. \ref{fig:experiment2} to analyze how the the different algorithms perform. In particular, our algorithm (EK-UCB) achieves low regret while running in low computational time.  

In the first example for the 'Bump' environment in Fig. \ref{fig:experiment}, for $T=2000$, we have set~$\lambda~=~10$ (of the order of $\sqrt{T}$) and see that the value of $\mu = \lambda$ indeed achieves a good tradeoff between regret and time. The parameter~$\mu$ determines the quality of the projection required in the algorithm. Thus, for a smaller~$\mu$, the algorithm achieves a better regret but pays a higher time complexity. We note that a similar role is played by the parameter~$C$ in the BBKB algorithm. The smaller~$C$, the more frequent the dictionary updates, and thus the slower is the algorithm.  While the CGP-UCB/K-UCB obtains the best regret, we  note also, that EK-UCB ($\mu=1$), CBKB (which is CBBKB with $C=1$) essentially take the full dictionary $m\approx T$ and thus also match K-UCB, but with dictionary building computational overheads which make them more computationally intensive than K-UCB itself. In the Appendix~\ref{app:experimental_details} we provide additional results that show that consistently EK-UCB provides the best time-regret compromise with regards to K-UCB.  

Second, in Fig. \ref{fig:experiment2} we show for the 'Chessboard' setting the influence of varying $\lambda$ for all methods (fixing $\mu=\lambda$ for EK-UCB). Both CBBKB and EK-UCB improve upon the K-UCB computational time in this case, but EK-UCB achieves lower computational times while also having lower regrets than CBBKB for all settings. We also notice that the CBKB algorithm runs much slower than the CBBKB algorithm in all experiments, as expected due to its costly dictionary update at every round which requires processing all previous points. The computational overheads of its dictionary building therefore makes it not practical despite its theoretical guarantees. Note also that CBBKB uses scores based on the variance estimates on past states for its ``resparfication'' strategy and EK-UCB uses leverage scores to build its dictionary thus looking for directions that are orthogonal to the previous anchor points; both approaches are more effective than updating the dictionary at each round. Eventually, recall that our incremental projection scheme allows us to perform rank-one updates of the dictionary. This also contributes to the practical speedup of our EK-UCB algorithm, as compared to the CBBKB strategy.

Moreover, SupK-UCB performs poorly in our experiments due to its over-exploring elimination strategy that might be beneficial only for large $T$ and makes it unpractical in its current time-complexity. Note that the main author of SupK-UCB co-authored \cite{calandriello19a} where it is mentioned that it indeed has "tighter analysis than GP-UCB [but] does not work well in practice". 

\section{Discussions}

In this work, we proposed a method for contextual kernel UCB algorithms in large-scale problems. The EK-UCB algorithm runs in $\cO(T \deff)$ space and $\cO(C T\deff^2)$ time complexity, which significantly improves over the standard contextual kernel UCB method. Note that while previous efficient Gaussian process algorithms allow to scale up the learning problems in non contextual and discrete action environments, we have shown how the incremental projection updates were crucial to perform efficient approximations in the joint context-action space, providing the same regret guarantees for a smaller computational cost.
We note that the batching strategy of BBKB may still be useful even under our incremental updates, and thus provides an interesting avenue for future work.
Another natural question is whether we may obtain algorithms with better regret guarantees similar to~\citet{valko2013} in the finite action case, while also achieving gains in computational efficiency as in our work.

\acknowledgments{JM was supported by the ERC grant number 714381 (SOLARIS project)
and by ANR 3IA MIAI@Grenoble Alpes, (ANR-19-P3IA-0003).}

\bibliographystyle{abbrvnat}
\bibliography{references}

% \documentclass[twoside]{article}

% \usepackage{aistats2022}
% If your paper is accepted, change the options for the package
% aistats2022 as follows:
%
%\usepackage[accepted]{aistats2022}
%
% This option will print headings for the title of your paper and
% headings for the authors names, plus a copyright note at the end of
% the first column of the first page.

% If you set papersize explicitly, activate the following three lines:
%\special{papersize = 8.5in, 11in}
%\setlength{\pdfpageheight}{11in}
%\setlength{\pdfpagewidth}{8.5in}

% If you use natbib package, activate the following three lines:
%\usepackage[round]{natbib}
%\renewcommand{\bibname}{References}
%\renewcommand{\bibsection}{\subsubsection*{\bibname}}

% If you use BibTeX in apalike style, activate the following line:
%\bibliographystyle{apalike}

% \begin{document}



% Supplementary material: To improve readability, you must use a single-column format for the supplementary material.
\onecolumn
\appendix
\hrule
\begin{center}
    {\huge \bfseries Appendix}
\end{center}
\hrule

\vspace*{0.5cm}

This appendix is organized as follows: 
\begin{itemize}[nosep, label={--}]
    \item Appendix~\ref{app:notations}: summary of the notations used in the analysis
    \item Appendix~\ref{app:proofKUCB}: proofs of Section~\ref{sec:kernelUCB} -- Kernel-UCB
    \item Appendix~\ref{app:proofEKUCB}: proofs of Section~\ref{sec:EKUCB} -- Efficient Kernel-UCB
    \item Appendix~\ref{app:implementation}: details on the implementation of the algorithms
    \item Appendix~\ref{app:experimental_details}: additional experiment details, discussions and results 
\end{itemize}

\label{appendix:proofs}

\section{List of notations}
\label{app:notations}
In this appendix, we recall useful notations that are used throughout the paper.

\smallskip
Below are generic notations and notations related to RKHS:
\begin{itemize}[nosep, label={--},topsep=-4pt]
    \item $[n] \defeq \{1,\dots,n\}$
    \item $\lesssim$ denotes an approximate inequality up to logarithmic multiplicative or additive terms
    \item $\cX$ is the input space
    \item $\cA$ is the action set
    \item $k: (\cX \times \cA) \times (\cX \times \cA) \to \R$ is a bounded positive definite Kernel
    \item $\kappa >0$ is an upper-bound on the kernel $\kappa^2 \geq \sup_{s \in \cX \times \cA} k(s,s)$.
    \item $\cH$ is the reproducing kernel Hilbert space associated to $k$
    \item $\phi: \cX \times \cA \to \cH$ is the feature map such that $k(s,s') = \langle \phi(s), \phi(s')\rangle$ for any $s,s'\in \cX \times \cA$
    \item $\langle \varphi , \varphi'\rangle \defeq \varphi ^\top \varphi'$ denotes the inner product for any {$\varphi, \varphi' \in \cH$}%$\phi, \phi' \in \cH$ 
    \item $\Vert \cdot \Vert$ denotes the norm associated to $\cH$. It is the one induced by the inner product, i.e., $\|\varphi\|^2  = \left<\varphi,\varphi\right>$
    \item $\Vert \cdot \Vert_{V}$ denotes for any symmetric positive semi-definite operator $V : \cH \to \cH$ the norm such that $\Vert \varphi \Vert_{V} = \Vert V^{1/2} \varphi \Vert$ for all $\varphi \in \cH$ 
    \item $L \preccurlyeq L'$ means that  $L-L'$ is positive semi-definite for two operators $L, L'$ on $\mathcal{H}$
    \item $\varphi \otimes \varphi': \cH \to \cH$ is the tensor product of $\varphi$ and $\varphi' \in \cH$
\end{itemize}

\smallskip
Below are notations related to the sequential setting. Here, $t\in [T]$ denotes the index of the round:
\begin{itemize}[nosep,label={--},topsep=-4pt]
    \item $\theta^* \in \cH$ is the unknown parameter
    \item $x_t \in \cX, a_t \in \cA$ are the context and action played at round $t$ 
    \item $s_t \defeq (x_t,a_t) \in \cX \times \cA$
    \item $\cS_{t} \defeq \{s_1,\dots, s_t\}$ denotes the history
    \item $\epsilon_1,\dots,\epsilon_T$ are independent centered sub-Gaussian noise
    \item $H_t \defeq (\epsilon_1,\dots,\epsilon_t)^\top$
    \item $\cF_t \defeq \sigma( \epsilon_1, \dots, \epsilon_t)$ is the natural filtration with respect to $(\epsilon_i)_{i\geq 1}$
    \item $r_t \defeq \langle \theta^*, \phi(x_t,a_t)\rangle + \epsilon_t$ is the reward
    \item $Y_t \defeq (r_1,\dots,r_t)^\top \in \R^t$ is the vector of rewards
    \item $\varphi_t \defeq \phi(x_t,a_t) \in \cH$
    \item $\lambda >0$ is the regularization parameter
    \item $F_t \defeq \sum_{s=1}^t\varphi_s \otimes \varphi_s$ is the covariance operator
    \item $V_t \defeq \sum_{s=1}^t \varphi_s \otimes \varphi_s + \lambda I:  \cH \to \cH$ is the regularized covariance operator
    \item $\Phi_t: \cH \to \R^t$ is the operator such that $[\Phi_t \varphi]_i = \varphi(x_i,a_i) = \langle \varphi, \phi(x_i,a_i)\rangle$ for any $\varphi \in \cH$ and $i \in [t]$
    \item $\Phi^*$ denotes the conjugate transpose of a linear operator $\Phi$ on $\cH$
    \item $K_t \defeq \Phi_t \Phi_t^*: \R^t \to \R^t$ is the kernel matrix at time $t\geq 1$. Note that $[K_t]_{ij} = k((x_i,a_i),(x_j,a_j))$.
    \item $\lambda_i(K_t)$ is the $i$-th largest eigenvalue of $K_t$
    \item $\deff(\lambda, t) \defeq \Tr (K_t (K_t + \lambda I_t)^{-1})$ is the effective dimension of the matrix $K_t$
\end{itemize}

\smallskip
Below are notations related to the Kernel-UCB algorithm without projections:
\begin{itemize}[nosep, label={--},topsep=-4pt]
    \item $\hat \theta_t \defeq V_t^{-1} \Phi_t^* Y_t$ is the estimator of the algorithm
    \item $\delta >0$ is the confidence level
    \item $\beta_t(\delta)$ is the radius of the confidence ellipsoid of the algorithm
    \item $\cC_t \defeq \big\{ \theta \in \cH: \big\|\theta - \hat \theta_{t-1}\big\|_{V_{t-1}} \leq \beta_t(\delta) \big\}$ is the confidence ellipsoid played by the algorithm
\end{itemize}

\smallskip
Below are notations related to the Kernel-UCB algorithm with projections. All along the analysis, the notation $\tilde x$ corresponds to the projected version of the object $x$. 
\begin{itemize}[nosep,label={--},topsep=-4pt]
    \item $\cZ_t \subset \big\{(x_1,a_1),\dots,(x_t,a_t)\big\}$ is a dictionary
    \item $\tilde \cH_t \defeq \mathrm{Span}\big\{\phi(s), s \in \cZ_t\big\}$ is a linear subspace of $\cH$ and is used at round $t$. 
    \item $P_t: \cH \to \tilde \cH_t$ is the Euclidean projection onto $\cH$ so that $\tilde \cH_t = \{P_t \varphi, \varphi \in \cH\}$
    \item $\tilde V_t \defeq \sum_{s=1}^t (P_t \varphi_s) \otimes (P_t \varphi_s) + \lambda I = P_t F_t P_t + \lambda I$ is the regularized projected covariance operator
    \item $\tilde \theta_t \defeq P_t \tilde V_t^{-1} P_t \Phi_t^* Y_t$ is the projected estimator of the algorithm
    \item $\tilde C_t \defeq  \big\{\theta \in \cH: \|\theta - \tilde \theta_{t-1}\|_{\tilde V_{t-1}} \leq \tilde \beta_t(\delta)\big\}$ is the confidence ellipsoid related to the projected estimator
    \item $\smash{\mu_t \defeq \big\Vert (I - P_t)F_t^{1/2}\big\Vert}^2$ is the approximation error of the projection
\end{itemize}
Eventually, we provide notations related to the kernel matrix computations when we write the update rules of the efficient algorithm.
\begin{itemize}[nosep,label={--},topsep=-4pt]
    \item $K_{\mathcal{S}}(s')$ is the kernel column vector $[k(s_1, s'), \dots, k(s_l, s')]^\top$ of size $|\mathcal{S}|=l$. Note that $K_{\cS_t}(s) = \Phi_t \phi(s)$\,.
    \item $K_{\mathcal{Z}\mathcal{S}}$ is the kernel matrix vector $[k(z, s)]_{z \in \mathcal{Z}, s \in \mathcal{S}}$ of size $|\mathcal{Z}| \times |\mathcal{S}|$. 
    \item $s_{t,a} = (x_t, a)$ refers to the pair of context $x_t$ and any action $a \in \cA_t$ that can be chosen in the UCB rule.
\end{itemize}

\section{Proofs of Section~\ref{sec:kernelUCB}: Kernel UCB}
\label{app:proofKUCB}
In this appendix we prove of Lemma \ref{lemma:beta_delta} and Theorem~\ref{thm:regret_bound}.

\label{appx:effective_dimension_proof}

% \begin{proposition}
%  For all $n \geq 1,  \lambda> 0$ and all input sequences $(x_1, a_1), \dots, (x_t, a_t)$ 
%  \begin{equation}
%      \sum_{k=1}^{t} \log \left( 1+\dfrac{\lambda_k (K_t)}{\lambda} \right) \leq \log \left( e + \dfrac{e t \kappa^2}{\lambda}  \right) \deff(\lambda, t)
%  \end{equation}
%  where $\lambda_k (K_t)$ denotes the $k$-th largest eigenvalue of $K_t$.
% \end{proposition}

% \begin{proof}
% Using that for $x>0$

% \begin{equation*}
%     \log(1+x) \leq \dfrac{x}{x+1} (1+ \log(1+x))
% \end{equation*}

% and denoting by $a(\lambda)$ the quantity $a(s, \lambda) = 1 + \log(1 +s/\lambda) = \log(e + es/\lambda)$, we get for any $n \geq 1$

% \begin{equation*}
%     \log(1+\dfrac{\lambda_k(K_{t})}{\lambda}) \leq \dfrac{\lambda_k(K_{t})}{\lambda_k(K_{t})+\lambda} a(\lambda_k(K_{t}), \lambda)
% \end{equation*}

% Therefore, summing over $k \geq 1$ and denoting by $\lambda_1$ the largest eigenvalue of $K_{t}$

% \begin{align*}
%     \sum_{k=1}^\top \log \left( 1 + \dfrac{\lambda_k(K_{t})}{\lambda} \right) &\leq a(\lambda_1, \lambda) \sum_{k=1}^\top \dfrac{\lambda_k (K_{t})}{\lambda + \lambda_k (K_{t})} \\
%     &\leq a(\lambda_1, \lambda) \Tr(K_{t}(K_{t}+\lambda I)^{-1}) \\
%     &\leq a(\lambda_1, \lambda) \deff(\lambda)
% \end{align*}

% Now using that $\lambda_1(K_{t}) \leq \Tr(K_{t}) = \sum_{t=1}^\top \Vert \phi(x_t) \Vert^2 \leq t \kappa^2$, we obtain that:

% \begin{equation*}
%     a(\lambda_1, \lambda) \leq \log(e + e \frac{t\kappa^2}{\lambda})
% \end{equation*}

% which concludes the proof. 
% \end{proof}

\subsection{Proof of Lemma~\ref{lemma:beta_delta}}
\label{appx:regret_bound_proof}

We first prove Lemma~\ref{lemma:beta_delta}, which controls the size of the confidence intervals considered by the algorithm. It states that with probability $1-\delta$, for all $t\geq 1$:
\begin{equation}
    \theta^* \in C_t, \quad \text{ where } \quad C_t = \big\{ \theta \in \mathbb{R}^d,  \Vert \theta - \hat \theta_{t-1}  \Vert_{V_{t-1}} \leq \beta_t(\delta) \big\} \,.
\end{equation}

\begin{customlemma}{\ref{lemma:beta_delta}}
Let $\delta \in (0,1)$. Assume $\kappa^2 \geq \sup_{s \in \cX \times \cA} k(s,s)$. Then with probability at least $1-T\delta$, for all $t \in [T]$
\begin{align*}
    \Vert \hat{\theta}_t - \theta^* \Vert_{V_t} & \leq  \sqrt{\lambda} \Vert \theta^* \Vert + \sqrt{2 \log\frac{1}{\delta} + \log \left(  \det \left( \frac{1}{\lambda} (K_t+ \lambda I) \right) \right)}  \\
    & \leq \sqrt{\lambda} \Vert \theta^* \Vert + \sqrt{2 \log\frac{1}{\delta} + \log \left( e + \dfrac{e t \kappa^2}{\lambda}  \right) \deff(\lambda,T) } \quad \eqdef \quad \beta_{t+1}(\delta).
\end{align*}
% where $\Vert \theta \Vert_{V_t}^2 = \theta^\top V_t \theta$.
% \label{lemma:beta_delta}
\end{customlemma}

\begin{proof}

The analysis is inspired by the one of \cite{oful2011} for linear bandits and uses inequality tails on vector valued martingales. We introduce $M_t= \sum_{s=1}^t \varphi_s \epsilon_s \in \cH$, which is a martingale with regards to the natural filtration $\cF_t \defeq \sigma( \epsilon_1, \dots, \epsilon_t)$. Solving the least-square optimization problem~\eqref{eq:theta_hat}, $\hat \theta_t$ equals
\begin{equation*}
    \hat{\theta_t} = V_{t}^{-1} \sum_{s=1}^t \varphi_s Y_s =  V_t^{-1} \sum_{s=1}^t \varphi_s ( \varphi_s^\top \theta^* + \epsilon_s ) = V_{t}^{-1} {\left((V_t - \lambda I_d) \theta^* + M_t\right)} = \theta^* - \lambda V_t^{-1} \theta^* + V_t^{-1} M_t \,.
\end{equation*}
Multiplying by the square root of $V_t$ and using the triangle inequality
\begin{equation*}
    \Big \Vert V_t^{1/2} \big(\hat{\theta}_t - \theta^*  \big) \Big \Vert = \Big\Vert -\lambda V_t^{-1/2}\theta^* + V_t^{-1/2} M_t \Big\Vert \leq \lambda \big\Vert V_t^{-1/2} \theta^* \big\Vert + \big\Vert V_t^{-1/2}M_t \big\Vert \,.
\end{equation*}
On the other hand, given that $V_t = F_t + \lambda I$ where $F_t$ is positive semi-definite, $V_t^{-1/2} \preccurlyeq \lambda^{-1/2} I$ and thus
\begin{equation*}
    \lambda \Vert V_t^{-1/2} \theta^* \Vert \leq \lambda \frac{1}{\sqrt{\lambda}} \Vert \theta^* \Vert = \sqrt{\lambda} \Vert \theta^* \Vert \,.
\end{equation*}
We now prove for the other term that with probability at least $1-\delta$ 
\begin{equation*}
    \Vert V_t^{-1/2} M_t \Vert \leq \sqrt{2 \log\frac{1}{\delta} + \log \det \frac{1}{\lambda}(K_t +\lambda I) } \,.
\end{equation*}

\textit{Step 1: Martingales} For all $\nu \in  \cH$, we define the random-variable
\begin{equation*}
    S_{t, \nu} = \exp \Big(\nu^\top M_t - \frac{1}{2} \nu^\top V_t \nu\Big) 
\end{equation*}
and now show that it is a $\cF_t$-super-martingale. First, note that the common distribution of the $\epsilon_1, \dots, \epsilon_t$ is 1-sub Gaussian, i.e., for all $\cF_{t-1}$-measurable real-valued random variable $\nu_{t-1}$, we have
\begin{equation}
     \E \big[ \exp(\nu_{t-1}\epsilon_t) | \cF_{t-1} \big] \leq \exp\Big(\frac{\nu_{t-1}^2}{2}\Big) \,.
     \label{eq:subGaussian}
\end{equation}
Thus, using that $M_t = M_{t-1} + \varphi_t \epsilon_t$ and $V_{t} = V_{t-1} + \varphi_t \otimes \varphi_t$, 
\begin{align*}
     \E \left[ S_{t, \nu} | \cF_{t-1} \right] 
        &  = \E\Big[  \exp \big(\nu^\top M_t - \frac{1}{2} \nu^\top V_t \nu\big) | \cF_{t-1}\Big] \\
        & = \E\Big[  S_{t-1,\nu} \exp\big(\nu^\top  \varphi_t \epsilon_t -\frac{1}{2} \nu^\top (\varphi_t \otimes \varphi_t) \nu \big) | \cF_{t-1}\Big] \\
        & = S_{t-1,\nu} \E\Big[   \exp\big(\nu^\top  \varphi_t \epsilon_t -\frac{1}{2} (\nu^\top \varphi_t)^2 \big) \big| \cF_{t-1}\Big]  \leq S_{t-1, \nu} \,,
\end{align*}
where the last inequality is by applying~\eqref{eq:subGaussian} with $\nu_{t-1} = \nu^\top \varphi_t$ since $\varphi_t = \phi(x_t,a_t)$ is $\cF_{t-1}$-measurable. Therefore, $S_{t,\nu}$ is a $\cF_t$-super-martingale for any $\nu \in \cH$, and
\begin{equation}
    \E\big[S_{t,\nu}\big] \leq \E\big[S_{0,\nu}\big] = \exp\Big(- \frac{\lambda}{2} \|\nu\|^2\Big) \,.
    \label{eq:supermart}
\end{equation}
Rewriting $S_{t, \nu}$ in its vertex form with $m=V_{t-1}M_t$ yields
\begin{equation*}
    S_{t, \nu} = \exp \left( - \frac{1}{2} (\nu -m)^\top V_t (\nu -m) \right) \times \exp\Big(\frac{1}{2} \big\Vert V_t^{-1/2} M_t \big\Vert^2\Big)\,,
\end{equation*}
which substituted into~\eqref{eq:supermart} entails
\begin{equation}
    \E\Bigg[ \exp \Big(- \frac{1}{2} (\nu -m)^\top V_t (\nu -m) \Big) \times \exp\Big(\frac{1}{2} \big\Vert V_t^{-1/2} M_t \big\Vert^2\Big)\bigg] \leq \exp\Big(- \frac{\lambda}{2} \|\nu\|^2\Big), \qquad \forall \nu \in \cH  \,.
    \label{eq:supermart_}
\end{equation}

\textit{Step 2: Laplace's method integrating}

Now, following Laplace's method which is standard for the proof of LinUCB, the goal is to integrate both sides of the above expression. Let us first rewrite it in order to consider finite dimensional objects thanks to the Kernel trick. 

Recalling $V_t \defeq \Phi_t^* \Phi_t + \lambda I$ and $K_t \defeq \Phi_t \Phi_t^*$, following \citep{valko2013}, we will use the following identities:
\begin{align}
    ( \Phi_t^* \Phi_t  + \lambda I)  \Phi_t^* &=  \Phi_t^* (  \Phi_t  \Phi_t^*  + \lambda I) \label{eq:kerneltrick1}\\
  \Rightarrow \hspace*{2.28cm}  V_t  \Phi_t^* &=  \Phi_t^* (K_t + \lambda I) \label{eq:kerneltrick2}\\
  \Rightarrow \qquad    \Phi_t^* (K_t + \lambda I)^{-1} &= V_t^{-1}  \Phi_t^*  \label{eq:kerneltrick3} \,.
\end{align}

Let $x \in \mathbb{R}^t$ and write $\nu = V_t^{-1} \Phi_t^* x \in \cH$ and recall that $m = V_t^{-1} M_t = V_t^{-1} \Phi_t^* H_t$, where $H_t = (\epsilon_1,\dots,\epsilon_t)^\top$. We have
\begin{align}
    \exp \Big(- \frac{1}{2} (\nu -m)^\top V_t & (\nu -m) \Big) 
         = \exp \Big(- \frac{1}{2} (x -H_t)^\top \Phi_t V_t^{-1} V_t V_t^{-1} \Phi_t^* (x - H_t) \Big) \nonumber \\
        & = \exp \Big(- \frac{1}{2} (x -H_t)^\top \Phi_t \Phi_t^* (K_t +\lambda I)^{-1} (x - H_t) \Big) \qquad \leftarrow \text{ by~\eqref{eq:kerneltrick3}}\nonumber \\
        & = \exp \Big(- \frac{1}{2} (x -H_t)^\top K_t (K_t +\lambda I)^{-1} (x - H_t) \Big) \hspace*{1.1cm} \leftarrow K_t = \Phi_t \Phi_t^* \nonumber\\
        & = \exp \Big(- \frac{1}{2} (x -H_t)^\top K_t^{1/2} (K_t +\lambda I)^{-1} K_t^{1/2} (x - H_t) \Big) \,, \label{eq:laplace1}
\end{align}
where the last equality is because $(K_t +\lambda I)^{-1}$ and $K_t^{1/2}$ commute. Similarly,
\[
    \exp\Big(- \frac{\lambda}{2} \|\nu\|^2\Big) = \exp\Big(- \frac{\lambda}{2} x^\top K_t^{1/2} (K_t + \lambda I)^{-2}  K_t^{1/2} x\Big) \,.
\]
Combining with~\eqref{eq:supermart} and~\eqref{eq:laplace1} thus gives for any $x \in \R^t$,
\begin{multline}
    \E \bigg[ \exp \Big(- \frac{1}{2} (x -H_t)^\top K_t^{1/2} (K_t +\lambda I)^{-1} K_t^{1/2} (x - H_t) \Big)  \times \exp\Big(\frac{1}{2} \big\Vert V_t^{-1/2} M_t \big\Vert^2\Big)\bigg] \\
    \leq \exp\Big(- \frac{\lambda}{2} x^\top K_t^{1/2} (K_t + \lambda I)^{-2}  K_t^{1/2} x\Big) \,.
    \label{eq:laplace2}
\end{multline}
Now, that we are back to finite dimensional space, the idea would consists in integrating both parts over $x \in \R^t$. But the matrix $K_t$ may be non-invertible, we thus need a few more steps to integrate over $\Im(K_t)$ only. 

\smallskip
Let $d_t = \mathrm{rank}(K_t)$ and $Q_t \in \R^{t \times d_t}$ the matrix formed by the orthonormal eigenvectors of $K_t$ with non-zero eigenvalues. 
Let $u \in \R^{d_t}$ then $Q_tu \in \Im(K_t)$ and there exists $x \in \R^t$ such that $K_t^{1/2} x = Q_t u$.  Defining $z \in \R^{d_t}$ such that $Q_t z =  K_t^{1/2} H_t$ and substituting into Inequality~\eqref{eq:laplace2} yields, for any $u \in \R^{d_t}$
\begin{multline}
    \E \bigg[ \exp \Big(- \frac{1}{2} (u - z)^\top Q_t^\top (K_t +\lambda I)^{-1} Q_t (u - z) \Big)  \times \exp\Big(\frac{1}{2} \big\Vert V_t^{-1/2} M_t \big\Vert^2\Big)\bigg] \\
    \leq \exp\Big(- \frac{\lambda}{2} u^\top Q_t^\top  (K_t + \lambda I)^{-2}  Q_t u \Big) \,.
    \label{eq:laplace3}
\end{multline}
Now, we integrate both sides over $u \in \R^{d_t}$, recognizing a multidimensional Gaussian density, we have
\begin{multline*}
\int_{\R^d} \exp \Big(- \frac{1}{2} (u - z)^\top Q_t^\top (K_t +\lambda I)^{-1} Q_t (u - z) \Big) d\mu(u) 
    = \sqrt{\det\big( 2 \pi  (Q_t^\top (K_t + \lambda I)^{-1} Q_t)^{-1} \big)} \\
    = \sqrt{ (2\pi)^{d_t} \prod_{i=1}^{d_t} \big(\lambda_i(K_t) + \lambda\big)} \,,
\end{multline*}
where $\lambda_i(K_t)$ is the $i$-th largest eigenvalue of $K_t$. Similarly 
\[
\int_{\R^d} \exp\Big(- \frac{\lambda}{2} u^\top Q_t^\top  (K_t + \lambda I)^{-2}  Q_t u \Big) d\mu(u) =  \sqrt{\det \Big(2 \pi \lambda^{-1} \big(Q_t^\top (K_t + \lambda I)^{-2} Q_t\big)^{-1} \Big) } = \sqrt{\Big(\frac{2\pi}{\lambda}\Big)^{d_t} \prod_{i=1}^{d_t} \big(\lambda_i(K_t) + \lambda\big)^2 } \,.
\]
Therefore, by the Fubini-Tonelli theorem, plugging the last two equations into Inequality~\eqref{eq:laplace3} entails
\[
   \sqrt{ (2\pi)^{d_t} \prod_{i=1}^{d_t} \big(\lambda_i(K_t) + \lambda\big)}  \E \bigg[\exp\Big(\frac{1}{2} \big\Vert V_t^{-1/2} M_t \big\Vert^2\Big)\bigg] \leq \sqrt{\Big(\frac{2\pi}{\lambda}\Big)^{d_t} \prod_{i=1}^{d_t} \big(\lambda_i(K_t) + \lambda\big)^2 } \,,
\]
which, after reorganizing the terms, yields
\begin{align*}
  \E \bigg[\exp\Big(\frac{1}{2} \big\Vert V_t^{-1/2} M_t \big\Vert^2\Big)\bigg] \leq \sqrt{\prod_{i=1}^{d_t} \Big(1 + \frac{\lambda_i(K_t)}{\lambda}\Big)} = \sqrt{\frac{\det (K_t+\lambda I)}{\lambda^t}} \,.
\end{align*}

\textit{Step 3: Markov-Chernov bound}. It remains to upper-bound the above expectation using concentration inequalities. For $u>0$,
\begin{align}
    P\left(\Vert V_t^{-1/2} M_t \Vert > u \right) = P\left(\dfrac{\Vert V_t^{-1/2} M_t \Vert^2}{2} > \dfrac{u^2}{2} \right) &\leq \exp \left( -\frac{1}{2} u^2 \right) \displaystyle \E \left[ \exp \left( \frac{1}{2} \Vert V_t^{-1/2} M_t \Vert^2 \right) \right] \nonumber \\
    &\leq \exp \left( -\dfrac{u^2}{2} + \dfrac{1}{2} \log \dfrac{\det(K_t + \lambda I)}{\lambda^t} \right )= \delta 
    \label{eq:martingale}
\end{align}
for the claimed choice 
\begin{equation*}
    u = \sqrt{2 \log \frac{1}{\delta} + \log \det \frac{1}{\lambda} (K_t+ \lambda I)}\,.
\end{equation*}
The proof then concludes by using Prop.~\ref{prop:d_eff} on the $\log \det \frac{1}{\lambda} (K_t+ \lambda I)$ term and by applying a union bound.
\end{proof}

\subsection{Proof of Theorem~\ref{thm:regret_bound}}

We are now ready to prove Theorem~\ref{thm:regret_bound}, which upper-bounds the regret of K-UCB. 

\begin{customthm}{\ref{thm:regret_bound}}
Let $T\geq 2$ and $\theta^* \in \mathcal{H}$. Assume that $|\langle \phi(x,a), \theta^* \rangle| \leq 1$ and $\|\phi(x,a)\| \leq \kappa$ for all $a \in \bigcup_{t=1}^{T} \mathcal{A}_t \subset \cA$ and $x \in \mathcal{X}$. Then, the K-UCB rule defined in Eq. \eqref{eq:argmax} for the choice $\mathcal{C}_t$ as in \eqref{eq:ellispoid} satisfies the pseudo-regret bound
\begin{align*}
    R_T & \leq 2 + 2 \sqrt{T \left( \log \left( e + \dfrac{e T \kappa^2}{\lambda}  \right) \deff(\lambda, T) \right)}  \left[ \sqrt{\lambda} \Vert \theta^* \Vert + \sqrt{2 \log (T) +   \log \left( e + \dfrac{e T \kappa^2}{\lambda}  \right) \deff(\lambda)} \right]  \\
        & \lesssim \sqrt{T} \Big(  \Vert \theta^* \Vert \sqrt{ \lambda  \deff(\lambda, T)}  + \deff(\lambda, T) \Big)   \,.
\end{align*}
\end{customthm}

\begin{proof}
Let $\delta\in (0,1/2)$. By Lemma \ref{lemma:beta_delta}, with probability $1-T\delta$, 
\begin{equation}
    \forall t \in [T], \quad \theta^* \in C_t \,.
    \label{eq:event_ucb}
\end{equation}

\textit{Step 1: Small instantaneous regrets under the event (\ref{eq:event_ucb}}). Assume that (\ref{eq:event_ucb}) holds. Let 
\begin{equation*}
    a_{t}^* \defeq \max_{a \in \mathcal{A}_t} \langle \phi(x_t,a), \theta^* \rangle \quad \text{and} \quad  \Delta_{t} \defeq  \langle \phi(x_t, a_t^{*}) - \phi(x_t, a_t), \theta^* \rangle
\end{equation*}
be respectively the optimal decision and the instantaneous regret at round t. We also define
\begin{equation*}
    \rho_t \in \argmax_{\theta \in C_t} \big\{ \langle \phi(x_t,a_t), \theta \rangle \big\} \,.
\end{equation*}
Since $\theta^* \in C_t$, we have
\begin{equation*}
    \langle \phi(x_t, a_t^*), \theta^* \rangle \leq \max_{\theta \in C_t} \{ \langle \phi(x_t, a_t^*), \theta \rangle \} = \text{K-UCB}_t(a_t^*) \leq \text{K-UCB}_t(a_t) = \max_{\theta \in C_t} \{ \langle \phi(x_t, a_t), \theta \rangle \} = \langle \phi(x_t, a_t), \rho_t \rangle\,,
\end{equation*}
which entails because $\theta^*$ and $\Tilde{\theta}_{t-1}$ belong to $C_t$,
\begin{equation*}
    \Delta_t = \langle \phi(x_t, a_t^{*}) - \phi(x_t, a_t), \theta^* \rangle \leq \langle \phi(x_t, a_t), \rho_t - \theta^* \rangle \leq \Vert \phi(x_t, a_t) \Vert_{V_{t-1}^{-1}} \Vert \rho_t - \theta^* \Vert_{V_{t-1}} \leq 2 \Vert \phi(x_t, a_t) \Vert_{V_{t-1}^{-1}} \beta_t(\delta) \,.
\end{equation*}
Recall that $\varphi_t \defeq \phi(x_t, a_t)$. Then, summing over $t=1, \dots, T$ and using that by assumption
\[
    |\Delta_t| \leq \big| \langle \phi(x_t, a_t^{*}), \theta^* \rangle \big|  + \big| \langle \phi(x_t, a_t), \theta^* \rangle\big| \leq   2 \sup_{x \in \cX, a\in \cA_t} |\langle \phi(x,a), \theta^* \rangle| \leq 2 \,,
\]
we can write the cumulative regret as
\begin{align}
    \sum_{t=1}^T \Delta_t
    & \textstyle \leq  \sqrt{ T \sum_{t=1}^T \Delta_t^2}  \hspace*{3.8cm} \leftarrow \quad \text{Jensen's inequality} \nonumber \\
    & \textstyle\leq 2 \sqrt{ T \sum_{t=1}^T \min \{ \Vert \varphi_t \Vert_{V_{t-1}^{-1}}^2 \beta_t(\delta)^2, 1 \} }   \nonumber \\
    & \textstyle\leq 2 \beta_T(\delta) \sqrt{T \sum_{t=1}^T \min \{ \Vert \varphi_t \Vert_{V_{t-1}^{-1}}^2, 1 \}}  \hspace*{.38cm} \leftarrow \quad 1 \leq \beta_t(\delta) \leq \beta_T(\delta)  \nonumber \\
    &\textstyle \leq  2 \beta_T(\delta) \sqrt{T \sum_{t=1}^T \log \Big( 1+ \Vert \varphi_t \Vert_{V_{t-1}^{-1}}^2 \Big)}  \hspace*{.2cm} \leftarrow \quad \min (u,1) \leq 2 \log(1+u), \ \forall u >0\,. \label{eq:regretbound1}
\end{align}

Now we will use the kernel trick to obtain a formulation of $\varphi_t^\top V_{t-1}^{-1}\varphi_t$ using gram matrices. 
Define $s_t := (x_t, a_t)$ and  $\cS_t := (s_i)_{ 1 \leq i \leq t}$ the historical data. For any $l \geq 1$ and $\cS \in (\cX \times \cA)^l$, we also denote by $K_{\mathcal{S}}(s')$ the kernel column vector $[k(s_1, s'), \dots, k(s_l, s')]^\top$ of size $|\mathcal{S}|=l$. Specifically, we have $K_{\cS_{t-1}}(s_t) := [k(s_1, s_t), \dots, k(s_{t-1}, s_t)]^\top =  \Phi_{t-1} \varphi_t \in \R^t$. When multiplying $V_{t-1} \defeq \Phi_{t-1}^* \Phi_{t-1} + \lambda I$ by $\varphi_t$ on the right, we can express
\begin{align*}
    V_{t-1} \varphi_t & =  \Phi_{t-1}^* K_{\cS_{t-1}}\left( s_t \right) + \lambda \varphi_t, \\
    \Rightarrow \hspace*{1cm} 
     \varphi_t & =  V_{t-1}^{-1}  \Phi_{t-1}^* K_{\cS_{t-1}}\left( s_t \right) + \lambda V_{t-1}^{-1} \varphi_t \\
    \Rightarrow \hspace*{1cm} \varphi_t &= \Phi_{t-1}^* (K_{t-1} + \lambda I)^{-1} K_{\cS_{t-1}}\left(s_t\right) + \lambda V_{t-1}^{-1} \varphi_t \,,
\end{align*}
where the last equation is by Eq.~\eqref{eq:kerneltrick3}. Thus, multiplying now by $\varphi_t^\top$ on the left and using $\varphi_t^\top \Phi_{t-1}^* = K_{\cS_{t-1}}(s_t)$ entails
\begin{align*}
    \varphi_t^\top \varphi_t &= K_{\cS_{t-1}}\left( s_t \right)^\top (K_{t-1} + \lambda I)^{-1} K_{\cS_{t-1}}\left(s_t\right) + \lambda \varphi_t^\top V_{t-1}^{-1} \varphi_t  \,.
\end{align*}
Therefore, reorganizing the terms ans recognizing $\smash{\Vert \varphi_t \Vert_{V_{t-1}^{-1}}^2 = \varphi_t^\top V_{t-1}^{-1} \varphi_t}$ and $\smash{k(s_t,s_t) = \varphi_t^\top \varphi_t}$, we can write
\begin{align*}
    1+& \Vert \varphi_t \Vert_{V_{t-1}^{-1}}^2 = 1 + \varphi_t^\top V_{t-1}^{-1} \varphi_t \\
    &= \dfrac{\lambda + k(s_t,s_t)}{\lambda}  - \dfrac{1}{\lambda} K_{\cS_{t-1}}(s_t)^\top (K_{t-1} + \lambda I)^{-1} K_{\cS_{t-1}}(s_t) \\
    &=  \dfrac{\lambda + k(s_t,s_t)}{\lambda} \bigg(1 - K_{\cS_{t-1}}(s_t)^\top (K_{t-1} + \lambda I)^{-1} K_{\cS_{t-1}}(s_t) \big(\lambda +k(s_t,s_t)\big)^{-1} \bigg) \\
    & = \dfrac{\lambda + k(s_t,s_t)}{\lambda} \det \bigg(1 - K_{\cS_{t-1}}(s_t)^\top (K_{t-1} + \lambda I)^{-1} K_{\cS_{t-1}}(s_t) \big(\lambda +k(s_t,s_t)\big)^{-1} \bigg) \\
    & = \dfrac{\lambda + k(s_t,s_t)}{\lambda} \det \bigg(I - (K_{t-1} + \lambda I)^{-1/2} K_{\cS_{t-1}}(s_t) \big(\lambda +k(s_t,s_t)\big)^{-1} K_{\cS_{t-1}}(s_t)^\top  (K_{t-1} + \lambda I)^{-1/2}   \bigg) \,,
\end{align*}
where the last equality follows by the matrix determinant lemma  $\det(I + AB^\top) = \det(I + B^\top A)$ if $A$ and $B$ are $n$-by-$m$ matrices. Then, $1+ \Vert \varphi_t \Vert_{V_{t-1}^{-1}}^2$ equals
\begin{multline*}
      \dfrac{\lambda + k(s_t,s_t)}{\lambda}  \det \bigg( (K_{t-1} + \lambda I)^{-1/2} \Big( K_{t-1} + \lambda I - K_{\cS_{t-1}}(s_t) \big(\lambda +k(s_t,s_t)\big)^{-1} K_{\cS_{t-1}}(s_t)^\top \Big) (K_{t-1} + \lambda I)^{-1/2}   \bigg) \\
      =  \dfrac{\lambda + k(s_t,s_t)}{\lambda } \dfrac{\det \Big( K_{t-1} + \lambda I - K_{\cS_{t-1}}(s_t) \big(\lambda +k(s_t,s_t)\big)^{-1} K_{\cS_{t-1}}(s_t)^\top \Big)  }{ \det(K_{t-1} + \lambda I)} \,.
\end{multline*}
Now, using that 
\[
    K_t + \lambda I = 
    \begin{bmatrix}
        K_{t-1} + \lambda I & K_{S_{t-1}}(s_t) \\
        K_{S_{t-1}}(s_t)^\top & k(s_t,s_t) + \lambda
    \end{bmatrix} \,,
\]
by the block matrix determinant formula
\[
    \det\big(K_t + \lambda I\big) = (k(s_t,s_t) + \lambda) \det\Big( K_{t-1} + \lambda I - K_{S_{t-1}}(s_t) (k(s_t,s_t) + \lambda)^{-1} K_{S_{t-1}}(s_t)^\top \Big)
\]
we finally get 
\begin{equation}
    1+ \Vert \varphi_t \Vert_{V_{t-1}^{-1}}^2  = \dfrac{1}{\lambda} \dfrac{\det(K_t + \lambda I)}{\det(K_{t-1} + \lambda I)} \,.
    \label{eq:telescoping_term}
\end{equation}

Note here that contrary to the proof in \cite{lattimore_szepesvari_2020}, we used here computations using the gram matrix $K_t$ instead of the $V_t$ which lives in the feature space that can be infinite dimensional. 

Taking the $\log$ and summing over $t=1,\dots,T$ telescopes
\begin{equation*}
    \sum_{t=1}^T \log \left( 1+ \Vert \varphi_t \Vert_{V_{t-1}^{-1}}^2 \right) = \log \left(  \det \left( \frac{1}{\lambda} (K_t+ \lambda I) \right) \right) 
    \leq \log \left(e + \dfrac{e T \kappa^2}{\lambda}  \right) \deff(\lambda, T) \,,
\end{equation*}
where we used the Proposition~\ref{prop:d_eff} for the last inequality and that . Substituting into the regret bound~\eqref{eq:regretbound1} together with $\beta_{T}(\delta) \leq \beta_{T+1}(\delta)$ entails with probability at least $1-T\delta$
\[
    \sum_{t=1}^T \Delta_t \leq 2 \beta_{T+1}(\delta) \sqrt{T \left(e + \dfrac{e T \kappa^2}{\lambda}  \right) \deff(\lambda, T)} \,.
\]
Choosing $\delta = 1/T^2$, taking the expectation $R_T = \E\big[\sum_{t=1}^T \Delta_t\big]$ and using $|\Delta_t| \leq 2$ concludes. 
\end{proof}

We now provide a proof for the Corollary that gives out the convergence speed of the K-UCB algorithm with the capacity condition assumption. 

\subsection{Proof of Corollary~\ref{cor:bound_capacity_condition}}

\begin{customcor}{\ref{cor:bound_capacity_condition}}
Assuming the capacity condition $\deff \leq (T/\lambda)^{\alpha}$ for $0 \leq \alpha \leq 1$, the regret of K-UCB is bounded as $R_{T} \lesssim T^{\frac{1+ 3\alpha}{2 + 2\alpha}}$ with an optimal $\lambda \approx T^{\frac{\alpha}{1+\alpha}}$.
\end{customcor}

\begin{proof}

Starting from $R_T \lesssim \sqrt{T} \Big( \sqrt{\lambda  \deff(\lambda)} + \deff(\lambda) \Big)$ and assuming the capacity condition $\deff(\lambda) \lesssim \Big( \frac{T}{\lambda}\Big)^\alpha$ for some $\alpha \in (0,1)$,
\[
R_T \lesssim  \sqrt{T} \Big( \sqrt{T^\alpha \lambda^{1-\alpha}  } + T^\alpha \lambda^{-\alpha} \Big)\,.
\]
Minimizing in $\lambda >0$ entails
\[
\sqrt{T^\alpha \lambda^{1-\alpha} } = T^\alpha \lambda^{-\alpha} \quad \Rightarrow \quad \lambda^* = T^{\frac{\alpha}{1+\alpha}} \,,
\]
which yields for $\lambda = \lambda^*$
\[
    R_T \lesssim T^{\frac{1}{2}+\alpha -\frac{\alpha^2}{1+\alpha}} = T^{\frac{1+ 3\alpha}{2 + 2\alpha}}\,.
\]

\end{proof}

\section{Proofs of Section~\ref{sec:EKUCB}: Efficient Kernel-UCB}
\label{app:proofEKUCB}

Let us start by recalling the setting and the notation of this section. Let $\cZ_t \subseteq \cS_{t}$, $\smash{\tilde \cH_t \defeq \mathrm{Span}\big\{\phi(s), s \in \cZ_t\big\}}$ be the corresponding linear subspace of $\cH$, and $P_t: \cH \to \tilde \cH_t$ be the Euclidean projection onto $\cH$ so that $\smash{\tilde \cH_t = \{P_t \varphi, \varphi \in \cH\}}$. The EK-UCB algorithm also builds an estimator
\begin{equation}
    \tilde{\theta}_{t-1} \in \argmin_{\theta \in \tilde{\mathcal H}_{t-1}} \bigg \{ \sum_{s=1}^{t-1} \left( \langle \theta, \phi(x_s, a_s) \rangle - r_s \right)^2 + \lambda \Vert \theta \Vert^2 \bigg \} \in {\tilde \cH}_{t-1} \,,
\end{equation}
and uses the confidence set $\smash{\tilde C_t \defeq  \big\{\theta \in \cH: \|\theta - \tilde \theta_{t-1}\|_{\tilde V_{t-1}} \leq \tilde \beta_t(\delta)\big\}}$. We define $\smash{\tilde{V}_t \defeq \sum_{s=1}^{t} (P_t \varphi_s) \otimes (P_t \varphi_s)  + \lambda I}$, that we rewrite $\tilde{V}_t = P_t F_t P_t + \lambda I$ where $\Phi_t^* = [\varphi_1, \dots, \varphi_t]$ and $F_t = \Phi_t^* \Phi_t$. Recalling the notation, $\smash{Y_t \defeq (r_1, \dots, r_t)^\top}$, we then obtain that $\smash{\tilde{\theta}_t = P_t \tilde{V}_t^{-1} P_t \Phi_t^* Y_t}$. We recall the definition $\smash{\mu_t \defeq \big\Vert  (I -P_t) F_t^{1/2} \big\Vert^2}$.

\subsection{Proof of Lemma~\ref{lemma:beta_delta_tilde}}

The following lemma serves to compute the distance of the center $\tilde{\theta}_t$ to any point in the ellipsoid in the projected space $\tilde{\mathcal H}_t$. Note that the norm uses the geometry induced by the direction matrix $\tilde{V}_t$.

\begin{customlemma}{\ref{lemma:beta_delta_tilde}}
 Let $\delta \in (0,1)$. Assume that  $\sup_{s \in \cX \times \cA} k(s,s) \leq \kappa^2$. Then, with probability $1-\delta$, for all $t \geq 1$
\begin{align*}
    \Vert \tilde{\theta}_t - \theta^* \Vert_{\tilde{V}_t} 
        & \leq \left(\sqrt{\lambda} +\sqrt{\mu_t}  \right) \Vert \theta^* \Vert + \sqrt{4 \log \frac{1}{\delta} + 2 \log \det \Big( \dfrac{{K}_t+\lambda I}{\lambda}} \Big) \\
        & \leq \left(\sqrt{\lambda} +\sqrt{\mu_t}  \right) \Vert \theta^* \Vert + \sqrt{4 \log\frac{1}{\delta} + 2 \log \left( e + \dfrac{e t \kappa^2}{\lambda}  \right) \deff(\lambda,T)} 
 \quad \defeq \quad \tilde{\beta}_{t+1}(\delta) \,,
\end{align*}
where $\Vert \theta \Vert_V^2 = \theta^\top V \theta$.
\end{customlemma}

\begin{proof}
Let $t\geq 1$. 
Note that $P_t V_t P_t = P_t (F_t + \lambda I) P_t = P_t \tilde{V}_t = \tilde{V}_t P_t$ and consequently as well $P_t \tilde{V}_t^{-1} = \tilde{V}_t^{-1} P_t$. We can write with $H_t \defeq (\epsilon_1, \dots, \epsilon_t)^\top$,
\begin{align*}
    \tilde{\theta}_t 
        &  = P_t \tilde{V}_t^{-1} P_t \Phi_t^* Y_t \\
        & = \tilde{V}_t^{-1} P_t \Phi_t^* Y_t \hspace{7cm} \leftarrow \quad  P_t \tilde{V}_t^{-1} = \tilde{V}_t^{-1} P_t \\
        & = \tilde{V}_t^{-1} P_t \Phi_t^* (\Phi_t \theta^* + H_t) \\
        &= \tilde{V}_t^{-1} P_t F_t P_t \theta^* + \tilde{V}_t^{-1} P_t F_t (I - P_t) \theta^* + \tilde{V}_t^{-1} P_t \Phi_t^* H_t \\
        &= \theta^* - \lambda \tilde{V}_t^{-1}\theta^* + \tilde{V}_t^{-1} P_t F_t (I - P_t) \theta^* + \tilde{V}_t^{-1} P_t \Phi_t^* H_t \,.
\end{align*}
To obtain later on the norm $\Vert \tilde{\theta}_t - \theta^* \Vert_{\tilde{V}_t}$, we multiply by $\tilde{V}_t^{1/2}$ on the left
\begin{equation}
    \tilde{V}_t^{1/2}(\tilde{\theta}_t - \theta^*) = - \underbrace{\lambda  \tilde{V}_t^{-1/2}\theta^*}_{\ref{item1}} + \underbrace{\tilde{V}_t^{-1/2} P_t F_t (I - P_t) \theta^*}_{\ref{item2}} + \underbrace{\tilde{V}_t^{-1/2} P_t \Phi_t^* H_t}_{\ref{item3}} \,.
    \label{eq:three_terms}
\end{equation}

We then compute each norm separately.
\begin{enumerate}[label=(\roman*),wide=0pt]
    \item \label{item1} Since $\tilde V_t = P_tF_tP_t + \lambda I$, all its eigenvalues are larger than $\lambda$. Thus, $ \tilde{V}_t^{-1/2} \preccurlyeq \lambda^{-1/2} I$, which implies 
    \begin{equation}
        \big\Vert \lambda  \tilde{V}_t^{-1/2} \theta^* \big\Vert \leq \sqrt{\lambda} \Vert \theta^* \Vert \,.
        \label{eq:item1}
    \end{equation}
    \item \label{item2} We write $\big\Vert \tilde{V}_t^{-1/2} P_t F_t (I - P_t) \theta^* \big\Vert = \big\Vert \tilde{V}_t^{-1/2} P_t F_t^{1/2} F_t^{1/2} (I - P_t) \theta^* \big\Vert$ and recall $\tilde{V}_t = P_t F_t^{1/2} F_t^{1/2} P_t + \lambda I$ therefore $\tilde{V}_t^{1/2} \succcurlyeq P_t F_t^{1/2}$, which entails
    \begin{equation}
        \big\Vert \tilde{V}_t^{-1/2} P_t F_t (I - P_t) \theta^* \big\Vert \leq \big\Vert F_t^{1/2} (I-P_t) \theta^* \big\Vert 
        \leq \sqrt{\mu_t} \Vert \theta^* \Vert\,,
        \label{eq:item2}
    \end{equation}
    where we recall that $\mu_t \defeq \big\|(I-P_t)F_t^{1/2}\big\|^2$ \,.
    \item \label{item3} Let us upper-bound the norm of the last term 
    \begin{align}
       \big\| \tilde V_t^{-1/2} P_t \Phi_t^* H_t \big\| 
            & \leq \big\| \tilde V_t^{-1/2} P_t V_t^{1/2} \big\| \big\| V_t^{-1/2} \Phi_t^* H_t \big\| \nonumber \\
            & \leq \big\| \tilde V_t^{-1/2} P_t V_t^{1/2} \big\| \sqrt{2 \log \frac{1}{\delta} + \log \det \Big( \frac{1}{\lambda}(K_t + \lambda I) \Big) } \,,
            \label{eq:term3}
    \end{align}
    with probability at least $1-\delta$, where the last inequality follows from the same analysis as~\eqref{eq:martingale}. 
    Then, using that $P_tV_tP_t = P_tF_tP_t + \lambda P_t = \tilde V_t + \lambda (P_t - I)$, we have
    \begin{multline*}
        \big\| \tilde V_t^{-1/2} P_t V_t^{1/2} \big\|^2 
            = \big\| \tilde V_t^{-1/2} P_t V_t P_t \tilde V_t^{-1/2} \big\|  
            = \big\| \tilde V_t^{-1/2} \big(\tilde V_t + \lambda (P_t -I)\big) \tilde V_t^{-1/2} \big\|  \\
            = \big\| I +  \lambda  \tilde V_t^{-1/2} (P_t -I) \tilde V_t^{-1/2} \big\| 
            \leq 1 + \lambda \big\|\tilde V_t^{-1/2}\big\|^2 \big\|P_t - I\big\| \leq 2 \,,
    \end{multline*}
    where the last inequality is because $\big\|P_t - I\big\| \leq 1$ and $\big\|\tilde V_t^{-1/2}\big\| \leq \lambda^{-1/2}$. Therefore, substituting into Inequality~\eqref{eq:term3} yields 
    \begin{equation}
        \big\| \tilde V_t^{-1/2} P_t \Phi_t^* H_t \big\| \leq \sqrt{4 \log \frac{1}{\delta} + 2 \log \det \Big( \frac{1}{\lambda}(K_t + \lambda I) \Big) } \,,
        \label{eq:item3}
    \end{equation}
    with probability at least $1-\delta$.
\end{enumerate}
Finally, combining~\eqref{eq:item1},~\eqref{eq:item2}, and~\eqref{eq:item3} with Equation~\eqref{eq:three_terms} concludes
\begin{align*}
    \big\| \tilde \theta_t - \theta \big\|_{\tilde V_t} 
        & \leq \lambda \big\|\tilde V_t^{-1/2} \theta^* \big\| +  \big\|\tilde{V}_t^{-1/2} P_t F_t (I - P_t) \theta^* \big\| + \big\| \tilde{V}_t^{-1/2} P_t \Phi_t^* H_t\big\| \\
        & \leq \big( \sqrt{\lambda} + \sqrt{\mu_t}\big) \big\|\theta^*\big\|  + \sqrt{4 \log \frac{1}{\delta} + 2 \log \det \Big( \frac{1}{\lambda}(K_t + \lambda I) \Big) } \,.
\end{align*}

The second line of the statement follows from Proposition~\ref{prop:d_eff}.
\end{proof}

\subsection{Proof of Theorem~\ref{thm:regret_bound_projected}}

\begin{customthm}{\ref{thm:regret_bound_projected}}
Let $T\geq 1$ and $\theta^* \in \cH$. Assume that $|\langle \phi(x,a), \theta^* \rangle| \leq 1$ for all $a \in \bigcup_{t=1}^{T} \mathcal{A}_t \subset \cA$ and $x \in \mathcal{X}$ then the EK-UCB rule in Eq.~\eqref{eq:solution_ekucb} with $\tilde{\mathcal{C}}_t$ defined in Eq.~\eqref{eq:ellispoid_tilde}, with $m=|\cZ_t|$ dictionary updates, satisfies the pseudo-regret bound 
\begin{align*}
    % R_T \leq \sqrt{T} \left[  \dfrac{\sqrt{2m \mu}}{\lambda}  +  \sqrt{ \left( \log \left( e + \dfrac{e T \kappa^2}{\lambda}  \right) \deff \right)} \right] \\
    % \left[ (\sqrt{\lambda} + \sqrt{\mu}) \Vert \theta^* \Vert + \sqrt{2 \log (T) +   \log \left( e + \dfrac{e T \kappa^2}{\lambda}  \right) \deff} \right] 
    R_T \lesssim \sqrt{T} \left( \sqrt{\dfrac{\mu m}{\lambda}} + \sqrt{\deff} \right) \left( \sqrt{\lambda} + \sqrt{\mu} + \sqrt{\deff} \right).
\end{align*}
In particular, the choice $\mu = \lambda$ yields $m \lesssim \deff$ and
\[
    R_T \lesssim \sqrt{T}  \big( \|\theta^*\| \sqrt{\lambda \deff }  + \deff \big) \,.
\]
\end{customthm}

\begin{proof}
Let $\delta > 0$. By Lemma \ref{lemma:beta_delta_tilde}, with probability $1-\delta$, 
\begin{equation}
    \forall t \geq 1, \quad \theta^* \in \tilde{C_t} \,.
    \label{eq:event_ucb_tilde}
\end{equation}

Let us recall and start from the definition of the regret 
\begin{align*}
    R_T \defeq \E\bigg[\sum_{t = 1}^T \Delta_t\bigg], \qquad \text{where}  \quad  \Delta_{t} \defeq \langle \phi(x_t, a_t^{*}) - \phi(x_t, a_t), \theta^* \rangle \quad \text{and} \quad a_{t}^* \defeq \max_{a \in \mathcal{A}_t} \langle \phi(x_t,a), \theta^* \rangle \,.
\end{align*}

\textit{Step 1: Small instantaneous regrets under the event (\ref{eq:event_ucb_tilde}}). Assume that (\ref{eq:event_ucb_tilde}) holds and define
\begin{equation*}
    \Tilde{\rho}_t \in \argmax_{\theta \in \tilde{C}_t} \{ \langle \phi(x_t,a_t), \theta \rangle \} \,.
\end{equation*}
Note here that the use of the original feature map allows us to not have any misspecified term that would have been incurred if  the projected feature map was used instead instead with $\langle \phi(x_t, a_t^*), \theta^* \rangle = \langle P_t \phi(x_t, a_t^*), \theta^* \rangle + \langle (I-P_t)\phi(x_t, a_t^*), \theta^* \rangle$) in the upper bound expression. 

Now given that $\theta^* \in \tilde{C}_t$ and $a_t \in \argmax_{a \in \mathcal{A}} \text{EK-UCB}_t(a)$, we have 
\begin{equation*}
    \langle \phi(x_t, a_t^*), \theta^* \rangle \leq \max_{\theta \in \tilde{C}_t} \{ \langle \phi(x_t, a_t^*), \theta \rangle \} = \text{EK-UCB}_t(a_t^*) \leq \text{EK-UCB}_t(a_t) = \max_{\theta \in \tilde{C}_t} \{ \langle \phi(x_t, a_t), \theta \rangle \} = \langle \phi(x_t, a_t), \Tilde{\rho}_t \rangle \,.
\end{equation*}
Therefore,
\begin{equation}
    \Delta_t \defeq \langle \phi(x_t, a_t^{*}) - \phi(x_t, a_t), \theta^* \rangle \leq \langle \phi(x_t, a_t), \Tilde{\rho}_t - \theta^* \rangle 
     \leq \Vert \varphi_t \Vert_{\tilde{V}_{t-1}^{-1}} \Vert \Tilde{\rho}_t - \theta^* \Vert_{\tilde{V}_{t-1}} \leq 2 \Vert \varphi_t \Vert_{\tilde{V}_{t-1}^{-1}} \tilde{\beta}_{t}(\delta) \,.
     \label{eq:deltat}
\end{equation}

Then, summing over $t=1, \dots, T$ and using $|\Delta_t| \leq 2$ and $\tilde \beta_{T}(\delta) \geq  \beta_t(\delta) \geq 1$, we get
\begin{align}
     \sum_{t=1}^T \Delta_t &\leq \sqrt{T \sum_{t=1}^T \Delta_t^2}  \hspace*{3.8cm} \leftarrow \quad \text{Jensen's inequality} \nonumber \\
    &\leq 2 \sqrt{T \sum_{t=1}^T \min \Big\{ \Vert \varphi_t \Vert_{\tilde{V}_{t-1}^{-1}}^2 \tilde{\beta}_t(\delta)^2, 1 \Big\} } \hspace*{.38cm} \leftarrow \quad |\Delta_t| \leq 2 \text{ and } \eqref{eq:deltat}\\
    & \leq 2 \tilde  \beta_T(\delta) \sqrt{T \sum_{t=1}^T \min \big\{ \Vert \varphi_t \Vert_{\tilde{V}_{t-1}^{-1}}^2, 1 \big\}} \hspace*{.38cm} \leftarrow \quad 1 \leq \beta_t(\delta) \leq \beta_T(\delta)   \,. \label{eq:regretboundphi}
\end{align}
Note now that 
\begin{align}
    \min \big\{ \Vert \varphi_t \Vert_{\tilde{V}_{t-1}^{-1}}^2, 1 \big\} 
        & \leq 2 \min \big\{  \Vert P_t \varphi_t \Vert_{\tilde{V}_{t-1}^{-1}}^2 + \Vert (I-P_t) \varphi_t \Vert_{\tilde{V}_{t-1}^{-1}}^2, 1 \big\} \nonumber \\
        & \leq 2 \min \big\{ \Vert P_t \varphi_t \Vert_{\tilde{V}_{t-1}^{-1}}^2, 1 \big\} +  2\Vert (I-P_t) \varphi_t \Vert_{\tilde{V}_{t-1}^{-1}}^2 \nonumber \\
        & \leq 4 \log \left( 1+ \Vert P_t \varphi_t \Vert_{\tilde{V}_{t-1}^{-1}}^2 \right) +  2\Vert (I-P_t) \varphi_t \Vert_{\tilde{V}_{t-1}^{-1}}^2. \label{eq:min_tilde}
\end{align}
The first term can be upper-bounded similarly to~\eqref{eq:telescoping_term}. First, note that since $P_s = P_s P_{t-1}$ for any $1\leq s\leq t-1$,
\[
    \tilde V_{t-1} \defeq \sum_{s=1}^{t-1} (P_{t-1} \varphi_s) \otimes (P_{t-1}\varphi_s) + \lambda I \succcurlyeq \sum_{s=1}^{t-1} (P_{s} \varphi_s) \otimes (P_s\varphi_s) + \lambda I \eqdef \tilde W_{t-1}
\]
which implies $\tilde V_{t-1}^{-1} \preccurlyeq \tilde W_{t-1}^{-1}$ and thus
\begin{equation}
   \log \left( 1+ \Vert P_t \varphi_t \Vert_{\tilde{V}_{t-1}^{-1}}^2\right) \leq \log \left( 1+ \Vert P_t \varphi_t \Vert_{\tilde{W}_{t-1}^{-1}}^2 \right)\,.
   \label{eq:firstterm}
\end{equation}
Now, recalling that $V_{t-1} \defeq \sum_{s=1}^{t-1} \varphi_s \otimes \varphi_s + \lambda I$, following the same analysis as for~\eqref{eq:telescoping_term}, replacing $\varphi_s$ with $P_{s} \varphi_s$ for all $s=1,\dots,t$, we get
\begin{equation*}
    1+ \Vert P_t \varphi_t \Vert_{\tilde W_{t-1}^{-1}}^2  = \dfrac{1}{\lambda} \dfrac{\det(\tilde K_t + \lambda I)}{\det(\tilde K_{t-1} + \lambda I)} \,,
\end{equation*}
where $\tilde K_{t} \in \R^{t\times t}$ is the kernel matrix such that $\big[\tilde K_{t}\big]_{ij} = \langle P_{i} \varphi_i, P_{j} \varphi_j\rangle$ for all $1\leq i,j\leq t$. Together with Inequalities~\eqref{eq:min_tilde} and~\eqref{eq:firstterm}, and summing over $t=1,\dots, T$, it yields
\begin{align}
    \sum_{t=1}^T \min \big\{ \Vert \varphi_t \Vert_{\tilde{V}_{t-1}^{-1}}^2, 1 \big\} 
        & \leq 4 \sum_{t=1}^T \log \left(\dfrac{1}{\lambda} \dfrac{\det(\tilde K_t + \lambda I)}{\det(\tilde K_{t-1} + \lambda I)} \right) + 2 \sum_{t=1}^T \Vert (I-P_t) \varphi_t \Vert_{\tilde{V}_{t-1}^{-1}}^2 \nonumber \\
        & \leq  4 \log \left(\dfrac{\det(\tilde K_T + \lambda I)}{\lambda^t} \right) + 2 \sum_{t=1}^T \Vert (I-P_t) \varphi_t \Vert_{\tilde{V}_{t-1}^{-1}}^2 \,.
\end{align}

We now upper-bound the second term in the right-hand-side. 
% Using that $(I-P_{t+1})$ and $(P_{t+1} - P_t)$ are orthogonal projections such that 
% \[
%     \langle (I-P_{t+1})\varphi_t, (P_{t+1} - P_t) \varphi_t\rangle =  \langle (I-P_{t+1})\varphi_t, P_{t+1} (P_{t+1} - P_t) \varphi_t\rangle = 0\,,
% \]
% we can write
% \begin{equation*}
%     \Vert (I-P_t)\varphi_t \Vert^2 = \Vert (I-P_{t+1}) \varphi_t \Vert^2 +  \Vert (P_{t+1} - P_t) \varphi_t \Vert^2 \,.
% \end{equation*}
% Thus, summing over $t=1,\dots,T$, and 
Denoting by $1= \tau_1 < \tau_2< \dots < \tau_m \leq T$ the indexes in time when the projection is updated, i.e., $P_{t} = P_{\tau_i}$ for all $t \in \{\tau_i,\dots, \tau_{i+1} - 1\}$, we can write
\begin{align*}
    \sum_{t=1}^T \big\Vert (I-P_{t}) \varphi_t \big\Vert^2 
    &= \sum_{i=1}^m \sum_{t=\tau_i}^{\tau_{i+1}-1} \big\|(I - P_{\tau_i}) \varphi_t \big\|^2 \\
    & = \sum_{i=1}^m \sum_{t = \tau_{i}}^{\tau_{i+1}- 1}  \Tr\big( (I-P_{\tau_i})  \varphi_t \otimes \varphi_t (I-P_{\tau_i}) \big) \\
    &  = \sum_{i=1}^m   \Tr\bigg( (I-P_{\tau_i}) \Big( \sum_{t = \tau_{i}}^{\tau_{i+1}- 1}  \varphi_t \otimes \varphi_t \Big) (I-P_{\tau_i}) \bigg) \\
    & = \sum_{i=1}^m   \Tr\bigg( (I-P_{\tau_{i+1}-1}) \Big( \sum_{t = \tau_{i}}^{\tau_{i+1}- 1}  \varphi_t \otimes \varphi_t \Big) (I-P_{\tau_{i+1}-1}) \bigg) \\
    & \leq  \sum_{i=1}^m   \Tr\bigg( (I-P_{\tau_{i+1}-1}) \Big( \sum_{t = 1}^{\tau_{i+1}- 1}  \varphi_t \otimes \varphi_t \Big) (I-P_{\tau_{i+1}-1}) \bigg) \\
    & \leq \sum_{l=1}^m \mu_{\tau_{i+1}-1} \leq m \mu \,,
\end{align*}
where the last inequality follows from Prop.~\ref{prop:kors}. Therefore, using that $\tilde V_{t-1}^{-1} \preccurlyeq \lambda^{-1} I$, from~\eqref{eq:min_tilde} we get
\[
   \sum_{t=1}^T \min \big\{ \Vert \varphi_t \Vert_{\tilde{V}_{t-1}^{-1}}^2, 1 \big\} \leq 4 \log \left(\dfrac{\det(\tilde K_T + \lambda I)}{\lambda^t} \right) + \frac{2 m \mu}{\lambda} \,.
\]
Substituting into Inequality~\eqref{eq:regretboundphi} entails
\begin{align*}
    \sum_{t=1}^T \Delta_t
        & \leq 2 \tilde  \beta_T(\delta) \sqrt{T  \bigg( 4 \log \left(\dfrac{\det(\tilde K_T + \lambda I)}{\lambda^t} \right) + \frac{2 m \mu}{\lambda}\bigg) } \\
        & \leq 2 \tilde  \beta_T(\delta) \sqrt{T  \bigg( 4 \log \det \left(\dfrac{K_T + \lambda I}{\lambda}  \right) + \frac{2 m \mu}{\lambda}\bigg) }\\
        & \leq 2 \tilde  \beta_T(\delta) \sqrt{T  \bigg( 4 \log \left( e + \dfrac{e T \kappa^2}{\lambda}  \right) \deff  + \frac{2 m \mu}{\lambda}\bigg) }
\end{align*}
where the last inequality is by Prop.~\ref{prop:d_eff} and where we recall
\begin{align*}
    \tilde \beta_T(\delta) \leq  \left(\sqrt{\lambda} +\sqrt{\mu}  \right) \Vert \theta^* \Vert + \sqrt{4 \log\frac{1}{\delta} + 2 \log \left( e + \dfrac{e T \kappa^2}{\lambda}  \right) \deff}\,.
\end{align*}
Choosing $\delta = 1/T$ and taking the expectation concludes
\begin{align*}
    R_T 
        & \leq 2 + 2 \tilde  \beta_T(1/T) \sqrt{T  \bigg( 4 \log \left( e + \dfrac{e T \kappa^2}{\lambda}  \right) \deff  + \frac{2 m \mu}{\lambda}\bigg)} \\
        & \lesssim \Big( (\sqrt{\lambda} + \sqrt{\mu}) \|\theta^*\| + \sqrt{\deff}\Big) \sqrt{T \Big(\deff + \frac{m\mu}{\lambda}\Big)} \,.
\end{align*}

In particular, for the choice $\mu = \lambda$, by Prop.~\ref{prop:kors}, the dictionary is at most of size $m \lesssim \deff$ with high probability. 
\end{proof}

\subsection{Proof of Cor.~\ref{cor:bound_capacity_condition_projected}}

\begin{customcor}{\ref{cor:bound_capacity_condition_projected}}
Assuming the capacity condition $\deff \leq (T/\lambda)^{\alpha}$ for $0 \leq \alpha \leq 1$. Let $1\leq m \leq  T^{\alpha/(1+\alpha)}$, under the assumptions of Thm.~\ref{thm:regret_bound_projected}, the regret of EK-UCB satisfies
\begin{equation*}
    R_{T} \lesssim \left\{
    \begin{array}{ll}
         & T m^{\frac{\alpha-1}{2\alpha}} \mbox{ if } m \leq T^{\frac{\alpha}{1+\alpha}} \\
         & T^{\frac{1+3\alpha}{2+2\alpha}} \mbox{ otherwise}
    \end{array}
\right.
\end{equation*}
for the choice $\lambda = \mu = T m^{-1/\alpha}$. Furthermore, the algorithm runs in $O(Tm)$ space complexity  and $O(CTm^2)$ time complexity.
\end{customcor}

We start from the regret bound of Theorem~\ref{thm:regret_bound_projected}, which, forgetting all dependencies that do not depend on $T$,  for the choice $\mu = \lambda$ yields
\begin{equation*}
    R_T \lesssim \sqrt{T}  \big( \sqrt{\lambda \deff }  + \deff \big) \,.
\end{equation*}
Under the capacity condition $\deff(\lambda, T) \leq \left({T}/{\lambda}\right)^\alpha$, it entails
\begin{equation*}
    R_T \lesssim \sqrt{T}  \big( \lambda^{\frac{1-\alpha}{2}}T^{\frac{\alpha}{2}}  + \lambda^{-\alpha}T^{\alpha} \big)  = T^{\frac{1}{2}} \big(T^{1/2} m^{\frac{\alpha - 1}{2\alpha}} +  m \big) = Tm^{\frac{\alpha - 1}{2\alpha}} + \sqrt{T} m  \,,
\end{equation*}
where we replaced $\lambda = Tm^{-1/\alpha}$. Optimizing in $m$, we retrieve the original rate $R_T \lesssim T^{\frac{1+3\alpha}{2+2\alpha}}$ for a dictionary of size $m = T^{\frac{\alpha}{1+\alpha}} \ll T$. Note that a larger dictionary is not necessary in theory since it only hurts both the theoretical rate and the computational complexity.   For a smaller dictionary, the first term is predominant and yields a regret of order $\cO\big(Tm^{\frac{\alpha-1}{2\alpha}}\big)$, highlighting a trade-off between the complexity which increases with $m$ and the regret which decreases.

\section{Details on the comparison of the regret bounds of CGP-UCB, SupKernelUCB, and K-UCB}
\label{app:comparison-regret}
In this appendix, we first detail why we can compare the regrets of  CGP-UCB \citep{krause2011contextual}, SupKernelUCB \citep{valko2013} and K-UCB as shown in Table \ref{table:comparison_algos}. We compare the quantities $\tilde{d}, \gamma$ and $\deff$ that appear in the regret bound of the literature \citep{valko2013,calandriello19a,krause2011contextual}. We show that they are essentially equivalent up to logarithmic factors. We recall first their definitions: for any $t \geq 0$ and $\lambda >0$
\begin{align*}
    \gamma(\lambda, t) &= \dfrac{1}{2} \log \left(\det \left( I + \dfrac{1}{\lambda} K_t \right)\right) \\
    \tilde{d}(\lambda, t) &= \min \{j:  j\lambda \log T \geq \sum_{k>j} \lambda_k(K_t) \} \\
    \deff(\lambda,T) &= \Tr (K_T (K_T + \lambda I_T)^{-1}) \,.
\end{align*}
We start by proving the first equality (up to logarithmic factors) $\deff(\lambda, t) \lesssim \gamma(\lambda, t) \lesssim \deff(\lambda, t)$. We first obtain that $\gamma(\lambda, t)  \lesssim \deff$ with Proposition \ref{prop:d_eff}. Next, to prove that $\deff  \lesssim \gamma(\lambda, t) $, we prove that for all $x>-1$ $\dfrac{x}{x+1} \leq \log(1+x)$ by writing for $x>-1$, $h(x) = \dfrac{x}{x+1} - \log(1+x)$ studying $h'$ and $h(0)$. Therefore,
\begin{equation*}
    \deff(\lambda, t) = \Tr(K_{t}(K{t}+ \lambda I)^{-1}) = \sum_{k=1}^t \dfrac{\frac{\lambda_k}{\lambda}}{\frac{\lambda_k}{\lambda} + 1} \leq \sum_{k=1}^t \log (1+ \frac{\lambda_k}{\lambda}) = \gamma(\lambda, t)
\end{equation*}
Next we detail that $\tilde{d}(\lambda, t) \lesssim \gamma(\lambda, t) \lesssim \tilde{d}(\lambda, t)$. First, \cite{valko2013} shows that $\tilde{d}(\lambda, t) \lesssim \gamma(\lambda, t)$. Second, to prove $\gamma(\lambda, t) \lesssim \tilde{d}(\lambda, t)$, we write 
\begin{equation*}
    \sum_{k=1}^t \log \left( 1 + \dfrac{\lambda_k}{\lambda}  \right)  \leq \sum_{k> \tilde{d}} \dfrac{\lambda_k}{\lambda} + \sum_{k \leq \tilde{d}} \log(1 + \dfrac{\lambda_k}{\lambda}) \leq \tilde{d}(\lambda, t) \log(t) + \tilde{d}(\lambda, t) \log \left( \dfrac{\lambda_1}{\lambda}\right),
\end{equation*}
where we used $\log(1+x) \leq x$ on the first term of the sum decomposition, and $\lambda_1$ the first and larger eigenvalue of the matrix $K_t$. Then, using $\lambda_1(K_{t}) \leq \Tr(K_{t}) = \sum_{k=1}^t \Vert \varphi_k \Vert^2 \leq t \kappa^2$, we subsequently obtain $\gamma(\lambda, t) \leq \tilde{d} \left(\log(T) + \log\left(\dfrac{t \kappa^2}{\lambda}\right) \right)$ which concludes the inequality.

\section{Algorithm Implementations}
\label{app:implementation}

Here we give details on the implementations of the contextual kernel UCB algorithms as well as our EK-UCB.

\subsection{Kernel UCB algorithm -- Implementation details}

Let us write  $s_{t, a} \defeq (x_t, a)$ and by abbreviation $s_i \defeq (x_i, a_i)$, let us write the historical data $\cS_t = (s_i)_{ 1 \leq i \leq t}$. Let us recall $\Phi_t^* = [ \varphi_1, \dots, \varphi_{t}]$ where $\varphi_i = \phi(x_i, a_i) = \phi(s_i)$ and $K_{\cS_{t}} \left(s \right) = \Phi_t \phi(s) = [k(s_1, s), \dots k(s_{t}, s)]^\top$. We write $F_t =\Phi_t^* \Phi_t$ and the gram matrix $K_t = \Phi_t  \Phi_t^* $. As in \citep{valko2013}:
\begin{align*}
    ( \Phi_t^* \Phi_t  + \lambda I)  \Phi_t^* &=  \Phi_t^* (  \Phi_t  \Phi_t^*  + \lambda I) \\
    (F_t + \lambda I)  \Phi_t^* &=  \Phi_t^* (K_t + \lambda I) \\
      \Phi_t^* (K_t + \lambda I)^{-1} &= (F_t + \lambda I)^{-1}  \Phi_t^*.
\end{align*}

\paragraph{Expression of the mean $\hat{\mu}_{t, a} = \langle \hat{\theta}_t, \varphi_{t, a} \rangle$.} For the mean expression recall that we have: $\hat{\mu}_{t, a} = \langle \hat{\theta}_{t-1}, \varphi_{t, a} \rangle = \varphi_{t, a}^\top \hat{\theta}_{t-1}$ and $\hat{\theta}_t = V_t^{-1} \Phi_t^* Y_t$. Therefore,
\begin{align*}
     \hat{\mu}_{t, a} = \varphi_{t, a}^\top \hat{\theta}_{t-1} = \varphi_{t, a}^\top \Phi_{t-1}^* (K_{t-1} + \lambda I)^{-1} Y_{t-1} = K_{\cS_{t-1}}\left(s_{t,a}\right)^\top (K_{t-1} + \lambda I)^{-1} Y_{t-1}.
\end{align*}

\paragraph{Expression of the standard deviation $\hat{\sigma}_{t, a} = \Vert \varphi_{t, a} \Vert_{V_{t-1}^{-1}}$.} When multiplying by $\varphi_{t, a} \defeq \phi(x_t, a)$ on the right and then by $\varphi_{t, a}^\top$ on the left 
\begin{align*}
    ( \Phi_{t-1}^* \Phi_{t-1}  + \lambda I) \varphi_{t, a} &=  \Phi_{t-1}^* K_{\cS_{t-1}}\left( s_{t,a} \right) + \lambda \varphi_{t, a} \\
    \varphi_{t, a} &= \Phi_{t-1}^* (K_{t-1} + \lambda I)^{-1} K_{\cS_{t-1}}\left(s_{t,a}\right) + \lambda ( \Phi_{t-1}^* \Phi_{t-1}  + \lambda I)^{-1} \varphi_{t, a} \\
        \varphi_{t, a}^\top \varphi_{t, a} &= K_{\cS_{t-1}}\left( s_{t,a} \right)^\top (K_{t-1} + \lambda I)^{-1} K_{\cS_{t-1}}\left(s_{t,a}\right) + \lambda \varphi_{t, a}^\top V_{t-1}^{-1} \varphi_{t, a} \\
    \hat{\sigma}_{t, a} =\Vert \varphi_{t, a} \Vert_{V_{t-1}^{-1}} &= \frac{1}{\lambda} k\left(s_{t,a}, s_{t,a} \right) - \frac{1}{\lambda} K_{\cS_{t-1}}\left(s_{t,a}\right)^\top (K_{t-1}  + \lambda I)^{-1} K_{\cS_{t-1}} \left(s_{t,a} \right)
\end{align*}
This allows to compute the UCB rule with kernel representations as illustrated in Alg.~\ref{alg:kernelucb}.
\begin{algorithm}[H]
\SetAlgoLined
\KwIn{$T$ the horizon, $\lambda$ regularization and exploration parameters, $k$ the kernel function}
% \KwResult{Write here the result }
 initialization\;
 $K_\lambda = \lambda, Y_0 = [r_0]$ where $r_0 = r(x_0, a_0)$ and $a_0$ is chosen randomly \;
%  $K_t \leftarrow [k\left( (x_i, a_i), (x_j, a_j) \right)]_{(i,j) \in [1, t]} + \lambda I$ \;
 \For{$t=1$ to $T$}{
 Observe context $x_t$ \; 
  Compute $\beta_t$ \;
 Choose $a_t \leftarrow \argmax_{a \in \mathcal{A}} \hat{\mu}_{t, a} + \beta_t
  \hat{\sigma}_{t,a}$ \;
  \Indp
  $\hat{\mu}_{t, a} \leftarrow K_{\cS_{t-1}}\left(s_{t,a}\right)^\top K_\lambda^{-1} Y_{t-1}$ \;
 $\hat{\sigma}_{t,a}^2 \leftarrow \frac{1}{\lambda} k\left(s_{t,a}, s_{t,a} \right) - \frac{1}{\lambda} K_{\cS_{t-1}}\left(s_{t,a}\right)^\top K_\lambda^{-1} K_{\cS_{t-1}} \left(s_{t,a} \right)$ \;
\Indm
 Observe reward $r_t$ and update $Y_t \leftarrow [r_1, \dots r_t]$ \;
 Update the translated gram matrix $K_\lambda \leftarrow [k(s_i,s_j)]_{1 \leq i,j \leq t} + \lambda I$ \;
 }
 \caption{Kernel UCB}
 \label{alg:kernelucb}
\end{algorithm}
Since the kernel matrices are used instead of estimating and computing directly $\hat{\theta}_t$ and $\phi(x_t, a)$, we can use first-rank updates of the matrices $K_t$, since:
% We then note with $b_t = K_{\cS_t}(s_{t,a})$ 
\begin{align*}
    K_t = \begin{bmatrix}
K_{t-1} & K_{\cS_{t-1}}\left(s_{t,a}\right) \\
K_{\cS_{t-1}}\left(s_{t,a}\right)^\top & K\left(s_{t,a}, s_{t,a} \right) 
\end{bmatrix}.
% = \begin{bmatrix}
% K_{t-1} & b_t \\
% b_t^\top & K\left(s_{t,a}, s_{t,a} \right) 
% \end{bmatrix}
\end{align*}
It is then easy to use the Schur complement on the inverse $K_\lambda^{-1}$. Specifically, the update is performed as the following, with 
\begin{align*}
    s &\leftarrow k\left(s_{t,a}, s_{t,a} \right) + \lambda - K_{\cS_{t-1}}(s_{t,a})^\top K_\lambda^{-1} K_{\cS_{t-1}}(s_{t,a}) \\
    Z_{12} &\leftarrow - \frac{1}{s} K_{\cS_{t-1}}(s_{t,a})^\top K_\lambda^{-1} \\
    Z_{21} &\leftarrow - \frac{1}{s} K_\lambda^{-1} K_{\cS_{t-1}}(s_{t,a})  \\
    Z_{11} &\leftarrow K_\lambda^{-1} + \frac{1}{s} K_\lambda^{-1} K_{\cS_{t-1}}(s_{t,a}) K_{\cS_{t-1}}(s_{t,a})^\top  K_\lambda^{-1} \\
    K_\lambda^{-1} &\leftarrow [Z_{11}, Z_{12}, Z_{21}, \frac{1}{s}].
\end{align*}
Therefore, while inverting the full matrices would induce as full cost of $\mathcal{O}(CT^4)$, using first order updates with Schur complement allows to run the algorithm in $\mathcal{O}(CT^3)$, while using $\mathcal{O}(T^2)$ in space. 
\\

% \begin{algorithm}[H]
% \SetAlgoLined
% \KwIn{$T$ the horizon, $\lambda$ regularization and exploration parameters, $k$ the kernel function}
% % \KwResult{Write here the result }
%  initialization\;
%  $Z = 1/\lambda, Y_0 = [r_0]$ where $r_0 = r(x_0, a_0)$ and $a_0$ is chosen randomly \;
% %  $K_t \leftarrow [k\left( (x_i, a_i), (x_j, a_j) \right)]_{(i,j) \in [1, t]} + \lambda I$ \;
%  \For{$t=1$ to $T$}{
%  Observe context $x_t$ \; 
%  Compute $b_t = K_{\cS_t}\left(s_{t,a}\right)$ \;
%   $\hat{\mu}_{t, a} \leftarrow b_t^\top Z Y_t$ \;
%  $\hat{\sigma}_{t,a}^2 \leftarrow K\left(s_{t,a}, s_{t,a} \right) -  b_t^\top Z b_t$ \;
%   Compute $\beta_t$ \;
%  Choose $a_t \leftarrow \argmax_{a \in \mathcal{A}} \hat{\mu}_{t, a} + \dfrac{\beta_t}{\lambda^{1/2}} \hat{\sigma}_{t,a}$ \;
%  Observe reward $r_t$ and update $Y_t \leftarrow [r_1, \dots r_t]$ \;
%  Update the inverse of the translated gram matrix $Z$ \; 
%  $s \leftarrow  K\left(s_{t,a}, s_{t,a} \right) + \lambda - b_t^\top Z b_t$ \;
%  $Z_{12} \leftarrow - \frac{1}{s} b_t^\top Z $\;
%  $Z_{21} \leftarrow - \frac{1}{s} Z b_t  $ \;
%  $Z_{11} \leftarrow Z + \frac{1}{s} Z b_t b_t^\top Z$ \;
%  $Z \leftarrow [Z_{11}, Z_{12}, Z_{21}, \frac{1}{s}]$
%  } 
%  \caption{Online-Kernelized UCB}
%  \label{alg:kernelucb}
% \end{algorithm}

\subsection{Efficient Kernel UCB algorithm -- Implementation details}

% We recall  $s_{t, a} = (x_t, a)$ and by abbreviation for $ 0 \leq i \leq t-1$, $s_i = (x_i, a_i)$. Let us now write $\Phi_t P_t = [ (P_t \varphi_1)^\top, \dots, (P_t\varphi_{t-1})^\top]^\top$ where $P_t\varphi_i = P_t\phi(x_i, a_i) = \phi(s_i)$ and $k_t \left(s \right) = \Phi_t \phi(s) = [k(s_1, s), \dots k(s_{t-1}, s)]$. We now write $\tilde{F}_t = P_t \Phi_t^* \Phi_t P_t = P_t F_t P_t$ and the projected gram matrix $\tilde{K}_t = \Phi_t P_t \Phi_t^* P_t$. 

Instead of using the kernel trick as in the standard algorithm, the efficient Kernel UCB algorithm uses computations in the projected feature space. The key high-level idea is to use as much as possible computations in the projected space $\mathcal{H}_t = \text{span} \{ \phi(z) \}_{z \in \mathcal{Z}_t}$ which is of dimension $m_t$ and does not use implicit kernel representation of the whole data which are of size $t \times t$. Here, we detail the computations of the predicted mean and variance bound in the projected space.

At the time $t$ we define the dictionary $\cZ_t = \{ z_1, \dots, z_{m_t} \}$ of size $|\cZ_t| = m_t$ and the $m_t \times m_t$ kernel matrix $K_{\cZ_t}~=~[k(z_i, z_j)]_{ 1 \leq i,j \leq m_t}$, we also write $K_{\cZ_t \cS_t} = [k(z_i, s_j)]_{1 \leq i \leq m_t, 1 \leq j \leq t}$ the $m_t \times t$ matrix on anchor points and historical data $\cS_t = \{s_i \}_{ 1 \leq i \leq t}$.

The following proposition provides closed-form formulas to implement EK-UCB (Alg.~\ref{alg:ek_ucb}). 

\begin{customprop}{\ref{prop:ek_ucb_implementation}}
  At any round $t$, by considering $s_{t,a}=~(x_t, a)$, the mean and variance term of the EK-UCB rule (Alg~\ref{alg:ek_ucb}) can be expressed as\footnote{Erratum: Note that the proposition slightly differs from the original one in the main document due to typos in the indexes that will be corrected in the final version of the manuscript.}
 \begin{align*}
    \Gamma_t &= K_{\cZ_{t-1}\cS_{t-1}} Y_{t-1} \\
    \Lambda_t &= \left(K_{\cZ_{t-1} \cS_{t-1}} K_{\cS_{t-1}\cZ_{t-1}} + \lambda K_{\cZ_{t-1}\cZ_{t-1}}\right)^{-1} \\
    \tilde \mu_{t, a} &= K_{\cZ_{t-1}}(s_{t,a})^\top \Lambda_t \Gamma_t \\
    \Delta_{t, a} &= K_{\cZ_{t-1}}(s_{t,a})^\top \left( \Lambda_t  - \frac{1}{\lambda} K_{\cZ_{t-1}\cZ_{t-1}}^{-1} \right) K_{\cZ_{t-1}}(s_{t,a})
\\
    \tilde \sigma_{t, a}^2 &= \frac{1}{\lambda} k(s_{t,a}, s_{t,a}) + \Delta_{t, a}.
\end{align*}
The algorithm then runs in a space complexity of $\mathcal{O}(Tm)$ and a time complexity of $\mathcal{O}(CTm^2)$.
\end{customprop}

\paragraph{Expression of the mean $\tilde{\mu}_{t+1, a} = \langle \tilde{\theta}_t, \varphi_{t+1, a} \rangle$.} At a time $t+1$, we look for $\tilde \theta \in \tilde \cC_{t+1}$ that we write $\tilde \theta = \alpha^\top K_{\cZ_t}$ where $\alpha \in \R^{m_t}$. We can rewrite the optimization process in Eq.~\eqref{eq:theta_tilde} as 
\begin{equation*}
    \argmin_{\alpha \in \mathbb{R}^{m_t}} \Big\{ (K_{\cS_t\cZ_t} \alpha - Y_t)^\top (K_{\cS_t\cZ_t} \alpha - Y_t) + \lambda \alpha^\top K_{\cZ_t\cZ_t} \alpha \Big\} \,
\end{equation*}
which can be rewritten as
\begin{equation*}
    \argmin_{\alpha \in \mathbb{R}^{m_t}} \Big\{  \alpha^\top K_{\cZ_t \cS_t} K_{\cS_t \cZ_t} \alpha - 2 \alpha^\top K_{\cZ_t \cS_t}Y_t + \lambda \alpha^\top K_{\cZ_t \cZ_t} \alpha \Big\},
\end{equation*}
and can be solved in closed-form as
\begin{equation*}
    \alpha^* = (K_{\cZ_t \cS_t} K_{\cS_t \cZ_t} + \lambda K_{\cZ_t \cZ})^{-1} K_{\cZ\cS_t} Y_t.
\end{equation*}
This eventually gives the expression $\tilde{\mu}_{t+1,a} = \alpha^\top K_{\cZ_ts_{t+1}}$
\begin{equation*}
\tilde{\mu}_{t+1,a} = K_{\cZ_t}(s_{t+1,a})^\top (K_{\cZ_t\cS_t} K_{\cS_t\cZ_t} + \lambda K_{\cZ_t\cZ_t})^{-1} K_{\cZ_t\cS_t} Y_t \,.
\end{equation*}

\paragraph{Expression of the standard deviation $\tilde{\sigma}_{t+1, a} = \Vert \varphi_{t+1, a} \Vert_{\tilde{V}_t^{-1}}$.}

When we look for the value of EK-UCB in Eq.~\eqref{eq:ek_ucb}, it is equivalent to have:
\begin{equation*}
    \text{EK-UCB}_{t+1}(a) = \max_{\theta \in \mathcal{H}, \text{ s.t } \Vert \theta - \tilde{\theta}_{t} \Vert_{\tilde{V}_{t}} \leq \beta} \langle\theta, \phi(s_{t+1,a}) \rangle = \tilde \mu_{t+1,a} + \beta \tilde \sigma_{t+1,a}.
\end{equation*}
where the variance term $\tilde \sigma_{t+1,a}$ is solution to 
\begin{align*}
    \max_{\theta \in \mathcal{H}} \ \theta^\top \phi(s_{t+1,a})  \,. \\
    \text{s.t } \Vert \theta \Vert_{\tilde{V}_{t}} \leq 1.
\end{align*}
% by substituting $\theta' = \tilde{V}_t^{1/2} \theta$, we can write the solution as $\theta = \dfrac{}{}\tilde{V}_t^{-1/2} \phi(s)$

% $\max_{\Vert \theta \Vert_{\tilde{V}_t} \leq \beta_t} \langle \theta, \phi(s) \rangle$
Below, we abbreviate $s = s_{t+1,a} \defeq (x_{t+1},a)$ for simplicity of notation. We advocate that at each time $t$ when we solve this maximization problem, $\theta$ lives in the finite dimensional space
\begin{equation*}
    \theta \in \mathcal{H}_{t+1, s} \eqdef \text{Span} \left( K_{z_{1}}, \dots K_{z_{m_t}}, K_{s} \right)\,,
\end{equation*}
where $K_z, K_s \in \cH$  such that $K_z(z') = k(z,z')$ and $K_s(s') = k(s, s')$.  To prove the above statement, following the Representer theorem proof, and $\mathcal{H}_{t+1, s}$ be the linear span of $K_{z_{1}}, \dots K_{z_{m_t}}, K_{s} \in \cH$. $\mathcal{H}_{t+1, s}$ is a finite dimensional subspace of $\cH$, therefore any $\theta \in \mathcal{H}$ can be uniquely decomposed as
\begin{equation*}
    \theta = \theta_{\mathcal{H}_{t+1, s}} + \theta_{\perp}
\end{equation*}
with $\theta_{\mathcal{H}_{t+1, s}} \in \mathcal{H}_{t+1, s}$ and $\theta_{\perp} \perp \mathcal{H}_{t+1, s}$. $\mathcal{H}$ being a RKHS it holds that $\langle \theta_{\perp}, \phi(s) \rangle = \langle \theta_{\perp}, K_{s} \rangle = 0$ because $K_s \in \mathcal{H}_{t+1, s}$. Therefore, $\langle \theta, K_s \rangle = \langle \theta_{\mathcal{H}_{t+1, s}}, K_s \rangle$. 

Now writing $\tilde{V}_t = P_t V_t P_t + \lambda (I-P_t)$, we have that $\Vert \theta \Vert_{\tilde{V}_t}$ can be written as
\begin{equation*}
    \Vert \theta \Vert_{\tilde{V}_t} = \theta_{\mathcal{H}_{t+1, s}}^\top P_t V_t P_t  \theta_{\mathcal{H}_{t+1, s}} + \lambda \theta_{\mathcal{H}_{t+1, s}}^\top (I-P_t) \theta_{\mathcal{H}_{t+1, s}} + \lambda \theta_{\perp}^\top (I-P_t) \theta_{\perp}.
\end{equation*}

Therefore, $\Vert \theta_{\mathcal{H}_{t+1, s}} \Vert_{\tilde{V}_t} \leq \Vert \theta \Vert_{\tilde{V}_t} \leq 1$. The maximization domain $\{ \theta \in \mathcal{H} \text{ s.t } \Vert \theta \Vert_{\tilde{V}_t} \leq 1\}$ is thus included in $\{ \theta \in \mathcal{H}_{t+1,s} \text{ s.t } \Vert \theta_{\mathcal{H}_{t+1, s}} \Vert_{\tilde{V}_t} \leq 1\}$ , while  $\langle \theta, K_s \rangle = \langle \theta_{\mathcal{H}_{t+1, s}}, K_s \rangle$. Therefore, $\max_{\theta \in \mathcal{H}} \langle \theta, K_s \rangle = \max_{\theta \in \mathcal{H}_{t+1, s}} \langle \theta_{\mathcal{H}_{t+1, s}}, K_s \rangle$. Hence we can write the solution of the problem from Eq.~\eqref{eq:solution_ekucb} as 

\begin{equation*}
    \theta_{\mathcal{H}_{t+1, s}}=  \sum_{i=1}^{m_t} \alpha_i K_{z_i} + \alpha_{m_t + 1} K_{s}, \ \alpha \in \mathbb{R}^{m_t}, \quad \alpha_{m_t + 1} \in \mathbb{R}.
\end{equation*}

We will write $\bar{K}_{\cZ_t} \alpha =  \sum_{i=1}^{m_t} \alpha_i K_{z_i}$ and therefore $\bar{K}_{\cZ_t}^\top \bar{K}_{\cZ_t} = K_{\cZ_t, \cZ_t}$ or even $\bar{K}_{\cZ_t}^\top K_{s} = K_{\cZ_t s} \in \mathbb{R}^{m_t}$. 

Using this notation allows us to write $P_t \varphi_{t+1} = \sum_{i=1}^{m_t} \beta_i(s_{t+1,a}) K_{z_i} = \bar{K}_{\cZ_t} (K_{\cZ_t, \cZ_t}^{-1} K_{\cZ_t}(s_{t+1,a}))$ where the $\beta$ coefficient is obtained by solving with the minimization problem defined in the Nystr\"om projection. Therefore when taking the projection $P_t : \mathcal{H} \rightarrow \mathbb{R}^{m_t}$ and the operator $\Phi_t : \mathbb{R}^t \rightarrow \mathcal{H}$ we can write $P_t \Phi_t = \bar{K}_{\cZ_t} (K_{\cZ_t, \cZ_t}^{-1}) K_{\cZ_t \cS_t}$. 

Therefore when writing $\tilde{V}_t = P_t F_t P_t + \lambda I$ we can express $\Vert \theta \Vert_{\tilde{V}_t}$ as
\begin{equation*}
    \Vert \theta \Vert_{\tilde{V}_t} = [\bar{K}_{\cZ_t} \alpha + \alpha_{m_t +1} K_{s_t}]^\top[\bar{K}_{\cZ_t} K_{\cZ_t, \cZ_t}^{-1} K_{\cZ_t, \cS_t}  K_{\cS_t, \cZ_t} K_{\cZ_t, \cZ_t}^{-1} \bar{K}_{\cZ_t}^\top + \lambda I] [\bar{K}_{\cZ_t} \alpha + \alpha_{m_t +1} K_{s}].
\end{equation*}
This can be reformulated as 
\begin{equation*}
\begin{bmatrix}
    \alpha & \alpha_{m_t+1}
    \end{bmatrix}
    Q_t
    \begin{bmatrix}
    \alpha \\ \alpha_{m_t+1}
    \end{bmatrix} \,,
\end{equation*}
where  $Q_t = \begin{bmatrix}
A_t  & b_t \\
b_t^\top & c_t
\end{bmatrix}$
and for which we have $A_t= K_{\cZ_t\cS_t} K_{\cS_t\cZ_t} + \lambda K_{\cZ_t\cZ_t}$, $b^\top_t = K_{s \cZ_t} K_{\cZ_t\cZ_t}^{-1} K_{\cZ_t\cS_t} K_{\cS_t\cZ_t} + \lambda K_{s \cZ_t}$ and eventually $c_t= K_{s \cZ_t} K_{\cZ_t\cZ_t}^{-1} K_{\cZ_t\cS_t} K_{\cS_t\cZ_t} K_{\cZ_t\cZ_t}^{-1} K_{ \cZ_t s} +\lambda K_{ss} $

Next to find the variance term, we note $q_t = [K_{\cZ_t s}, K_{ss}]^\top$ and reformulate the optimization process above as
\begin{align*}
    \max_{\alpha \in \mathbb{R}^{m_t+1}}  \ \alpha^\top q_t \\
    \text{s.t } \alpha^\top Q_t \alpha \leq 1
\end{align*}
% which by changing variable $\alpha' = Q^{1/2} \alpha$ gives 

% \begin{align*}
%     \max_{\alpha' \in \mathbb{R}^{m_t+1}}  \ \alpha'^\top Q^{-1/2} q \\
%     \text{s.t } \alpha'^\top \alpha' \leq 1
% \end{align*}

gives the solution $\alpha' = \dfrac{Q_t^{-1/2} q_t}{\Vert Q_t^{-1/2} q_t \Vert}$ which gives $\sigma_{t,a}$ the maximum value: $\sqrt{q_t^\top Q_t^{-1} q_t}$. We will now express the squared maximum $\sigma_{t+1,a}^2 = q_t^\top Q_t^{-1} q_t$ using the Schur complement on the $Q_t$ matrix.

Defining $A_t = K_{\cZ_t\cS_t} K_{\cS_t\cZ_t} + \lambda K_{\cZ_t\cZ_t}$ and the Schur complement $l_t= c_t - b_t^\top A_t^{-1} b_t $.

\smallskip
We start by simplifying the expression of the Schur complement. For this we reformulate 
\begin{align*}
    A_t &= K_{\cZ_t\cS_t} K_{\cS_{t}\cZ_t} + \lambda K_{\cZ_t\cZ_t} \\
    b_t^\top &= K_{s \cZ_t} K_{\cZ_t\cZ_t}^{-1} (A_t - \lambda K_{\cZ_t\cZ_t}) + \lambda K_{s\cZ_t} \\
    &= K_{s \cZ_t} K_{\cZ_t\cZ_t}^{-1} A_t \\
    c_t &= K_{s \cZ_t} K_{\cZ_t\cZ_t}^{-1} (A_t - \lambda K_{\cZ_t\cZ_t}) K_{\cZ_t\cZ_t}^{-1} K_{ \cZ_t s} +\lambda K_{ss} \\
    &=  K_{s \cZ_t} K_{\cZ_t\cZ_t}^{-1} A_t K_{\cZ_t\cZ_t}^{-1} K_{ \cZ_t s} - \lambda K_{s \cZ_t} K_{\cZ_t\cZ_t}^{-1}K_{ \cZ_t s} +\lambda K_{ss}.
\end{align*}
Thus we obtain:
\begin{align*}
    l_t &= K_{s \cZ_t} K_{\cZ_t\cZ_t}^{-1} A_t K_{\cZ_t\cZ_t}^{-1} K_{ \cZ_t s} - \lambda K_{s \cZ_t} K_{\cZ_t\cZ_t}^{-1}K_{ \cZ_t s} +\lambda K_{ss} - K_{s \cZ_t} K_{\cZ_t\cZ_t}^{-1} A_t A_t^{-1} A_t K_{\cZ_t\cZ_t}^{-1} K_{ \cZ_t s} \\
    &= \lambda (K_{ss} - K_{s \cZ_t} K_{\cZ_t\cZ_t}^{-1}K_{ \cZ_t s}).
\end{align*}
Then we write the product between $Q_t^{-1}$ and $q_t$ as:
\begin{align*}
\tilde \sigma_{t+1,a}^2 &= 
\begin{bmatrix}
K_{s \cZ_t} & K_{ss}
\end{bmatrix}
 \begin{bmatrix}
A_t^{-1} +  \frac{1}{l} A_t^{-1} b_t b_t^\top A_t^{-1}  & - \frac{1}{l} A_t^{-1} b_t \\
- \frac{1}{l}  b_t^\top A_t^{-1} & \frac{1}{l} 
\end{bmatrix} 
\begin{bmatrix}
K_{\cZ_t s} \\ K_{ss}
\end{bmatrix}
\\
&= \begin{bmatrix}
K_{s \cZ_t}A_t^{-1} + \frac{1}{l} K_{s \cZ_t} A_t^{-1} b_t b_t^\top A_t^{-1} - \frac{1}{l} K_{ss} b_t^\top A_t^{-1} & - \frac{1}{l}\lambda K_{s \cZ_t} A_t^{-1} b_t +  \frac{1}{l} K_{ss}
\end{bmatrix}
\begin{bmatrix}
K_{\cZ_t s} \\ K_{ss}
\end{bmatrix} \\
&=K_{s \cZ_t}A_t^{-1}K_{\cZ_t s} + \frac{1}{l} K_{s \cZ_t} A_t^{-1} b_t b_t^\top A_t^{-1} K_{\cZ_t s} - \frac{1}{l} K_{ss} b_t^\top A_t^{-1}K_{\cZ_t s}  - \frac{1}{l} K_{s \cZ_t} A_t^{-1} b_t K_{ss} +  \frac{1}{l} K_{ss}^2
\\
&= K_{s \cZ_t} A_t^{-1}K_{\cZ_t s} + \frac{1}{l} \left(K_{s \cZ_t} A_t^{-1} b_t - K_{ss} \right)^2 \\
&= K_{s \cZ_t}A_t^{-1}K_{\cZ_t s} + \frac{1}{l} \left(K_{s \cZ_t} K_{\cZ_t\cZ_t}^{-1}K_{ \cZ_t s} - K_{ss} \right)^2 \\
&=  K_{s \cZ_t}A_t^{-1}K_{\cZ_t s} + \frac{1}{\lambda} K_{ss} - \frac{1}{\lambda} K_{s \cZ_t}K_{\cZ_t\cZ_t}^{-1}K_{ \cZ_t s}\\
& = \frac{1}{\lambda} k(s_{t,a}, s_{t,a}) + \Delta_{t+1, a}\,,
\end{align*}
where $\Delta_{t+1, a} \defeq K_{\cZ_{t}}(s_{t+1,a})^\top \left( \Lambda_{t+1}  - \frac{1}{\lambda} K_{\cZ_{t}\cZ_{t}}^{-1} \right) K_{\cZ_{t}}(s_{t+1,a})$ and $\Lambda_{t+1} \defeq A_{t+1}^{-1}$.

This proves the first of Prop.~\ref{prop:ek_ucb_implementation}.

\begin{algorithm}[h]
\SetAlgoLined
\KwIn{$T$ the horizon, $\lambda$ regularization and exploration parameters, $k$ the kernel function, $\epsilon > 0, \gamma > 0$}
% \KwResult{Write here the result }
 Initialization\;
 Context $x_0$, $a_0$ chosen randomly and reward $r_0$ \;
 $\mathcal{S} = \{(x_0, a_0) \}, Y_\mathcal{S} =[r_0]$ \;
 $\mathcal{Z} = \{ (x_0, a_0) \}$ \;
 $\Lambda_t = \left(K_{\mathcal{Z} \mathcal{S}}   K_{\mathcal{S}\mathcal{Z}} + \lambda K_{\mathcal{Z}\mathcal{Z}}\right)^{-1}$ 
 $\Gamma_t =  K_{\mathcal{Z} \mathcal{S}}Y_\mathcal{S}$ \;
%  $\Xi = K_{\mathcal{Z}'\mathcal{Z}'}^{-1}$
 
 \For{$t=1$ to $T$}{
 Observe context $x_t$ \; 
 Choose $\tilde{\beta}_t$ (e.g as in Lem. \ref{lemma:beta_delta_tilde}, and $\delta=\frac{1}{T^2}$) \;
Choose $a_t \leftarrow \argmax_{a \in \mathcal{A}} \tilde{\mu}_{t, a} + \tilde{\beta}_t \tilde{\sigma}_{t,a}$ \;
 \Indp{$\tilde{\mu}_{t, a} \leftarrow K_{\mathcal{Z}}(s_{t,a})^\top \Lambda_t \Gamma_t$ \;
 $\Delta_{t,a} = K_{\cZ}(s_{t,a})^\top \left( \Lambda_t  - \frac{1}{\lambda} K_{\cZ\cZ}^{-1} \right) K_{\cZ}(s_{t,a})$ \;
 $\tilde{\sigma}_{t,a}^2 \leftarrow \frac{1}{\lambda} k(s_{t,a}, s_{t,a}) + \Delta_{t, a}$ \;}
  
 \Indm
 Observe reward $r_t$ and $s_t \leftarrow (x_t, a_t)$ \;
 $Y_\mathcal{S} \leftarrow [Y_\mathcal{S}, r_t]^\top, \mathcal{S} \leftarrow \mathcal{S} \cup \{ s_t \}$ \;
 $\mathcal{Z}' \leftarrow \text{KORS}(t, \cZ, K_{\mathcal{Z}}(s_t), \lambda, \epsilon, \gamma )$ \;
 \If{$\mathcal{Z}' = \mathcal{Z}$ }{
 Incremental inverse update $\Lambda_t$ with $s_t$\;
 $\Gamma_{t+1} \leftarrow \Gamma_t + r_t K_{\mathcal{Z}}(s_t)$ \;
 }
 \Else{
 $z = \mathcal{Z}' \ \backslash \ \mathcal{Z}$ \;
 Incremental inverse update $\Lambda_t$ with $s_t, z$ \;
 Incremental inverse update $K_{\cZ\cZ}^{-1}$ with $z$\;
 Update $\Gamma_{t+1} \leftarrow [\Gamma_t + r_t K_{\mathcal{Z}}(s_t), \  K_{\mathcal{S}}(z)^\top Y_\mathcal{S}]^\top$
 }
 }
 \caption{Efficient Kernel UCB}
%  \label{alg:ek_ucb}
\end{algorithm}

\paragraph{Discussion on practical implementation and time and space complexities}

% We have at time $t$ given a context $x_t$ for each action $a \in \mathcal{A}$, using the notations on $\Lambda_t, \Gamma_t$ in Prop. \ref{prop:ek_ucb_implementation} that the expressions of the mean and variance are given by:
% \begin{align*}
%     \mu_{t, a} &= K_{\cZ_{t-1}}(s_{t,a})^\top (K_{\cZ_{t-1} \cS_{t-1}} K_{\cS_{t-1}\cZ_{t-1}} + \lambda K_{\cZ_{t-1} \cZ_{t-1}})^{-1} K_{\cZ_{t-1} \cS_{t-1}} Y_{t-1} \\
%     \sigma_{t, a}^2 &= \frac{1}{\lambda} k(s_{t,a}, s_{t,a}) + K_{\cZ_{t-1}}(s_{t,a})^\top \left[(K_{\cZ_{t-1}\cS_{t-1}} K_{\cS_{t-1}\cZ_{t-1}} + \lambda K_{\cZ_{t-1}\cZ_{t-1}})^{-1} - \frac{1}{\lambda} K_{\cZ_{t-1}\cZ_{t-1}}^{-1} \right] K_{\cZ_{t-1}}(s_{t,a}).
% \end{align*}

The efficient implementation of the algorithm requires to perform efficient updates of the quantities (defined in Prop~\ref{prop:ek_ucb_implementation}) $\Lambda_t = (K_{\cZ_{t-1} \cS_{t-1}} K_{\cS_{t-1}\cZ_{t-1}} + \lambda K_{\cZ_{t-1} \cZ_{t-1}})^{-1}$ and $\Gamma_t = K_{\cZ_{t-1}\cS_{t-1}} Y_{t-1}$.

(i) When the dictionary is not updated $\cZ_{t} = \cZ_{t-1}$. For the matrix $\Gamma_t$ we can perform the update $\Gamma_{t+1}  \leftarrow \Gamma_t + r_t K_{\cZ_t}(s_t) $ which requires $m_t$ kernel evaluations. As for the matrix $\Lambda_t$ we can use the first rank Shermann-Morrison formula on it by adding updates on $s_t$ in $\mathcal{O}(m_t^2)$ operations where $\Lambda_{t+1} = (K_{\cZ_t\cS_t} K_{\cS_t \cZ_t} + \lambda K_{\cZ_t\cZ_t})^{-1}$. Here we only store $ K_{\cZ_t\cZ_t}^{-1}$ and do not update it.

(ii) When the dictionary is updated $\cZ_{t} \neq \cZ_{t-1}$ and we can write $\cZ_t = \cZ_{t-1} \cup \{ z_{m_t} \}$, 

Regarding $\Gamma_t$, we do two updates, one on the state $s_t$ by adding $r_t K_{\cZ_{t-1}}(s_t)$ and a second on the new anchor point $z_{m_t}$ so that we have
\begin{equation*}
    \Gamma_{t+1} \leftarrow [\Gamma_t + r_t K_{\cZ_{t-1}}(s_t), K_{\cS_t}(z_{m_t})^\top Y_t]^\top \,.
\end{equation*}
The first update is performed in $\mathcal{O}(m_t)$ kernel evaluations as in the (i) case, and the second update requires $\mathcal{O}(t)$ kernel evaluations and then $\mathcal{O}(t)$ computations. Note that the (ii) is only visited at most $m$ times which is the size of the dictionary at $t=T$. 

Regarding $\Lambda_t$, we note that we can write $ K_{\cZ_t \mathcal{S}_t} K_{\mathcal{S}_t \cZ_t} + \lambda K_{\cZ_t \cZ_t}$ as
\begin{equation*}
    \begin{bmatrix}
    K_{\cZ_{t-1} \cS_t} K_{\cS_t\cZ_{t-1}} + \lambda K_{\cZ_{t-1} \cZ_{t-1}} & K_{\cZ_{t-1} \cS_t} K_{\cS_t}(z) + \lambda K_{\cZ_{t-1}}(z) \\
    K_{\cS_t}(z)^\top K_{\cS_t \cZ_{t-1}}  + \lambda K_{\cZ_{t-1}}(z)^\top & K_{\cS_t}(z)^\top K_{\cS_t}(z) + \lambda k(z, z)
    \end{bmatrix}.
\end{equation*}

We perform the update in two stages by first computing the inverse $(K_{\cZ_{t-1} \cS_t} K_{\cS_t \cZ_{t-1}} + \lambda K_{\cZ_{t-1} \cZ_{t-1}})^{-1}$ by using a first-rank Sherman Morrison on the state update $s_t$, as if the dictionary did not change, and we then perform a Schur complement update using the latter inverse. Both updates are done in $\mathcal{O}(m_t^2)$  operations. 

As for the inverse of the projection gram matrix, we use a Schur complement update in $\mathcal{O}(m_t^2)$ operations that we detail here for $K_{\cZ_{t+1}\cZ_{t+1}}^{-1}$:
\begin{align*}
K_{\cZ_t\cZ_t}^{-1} = \begin{bmatrix}
K_{\cZ_{t-1}\cZ_{t-1}}^{-1} + \frac{1}{\omega} w_t w_t^\top & -\frac{1}{\omega} w_t \\
-\frac{1}{\omega} w_t^\top & \frac{1}{\omega}
\end{bmatrix}
\end{align*}

where $\omega = k(z_{m_t}, z_{m_t}) - K_{\cZ_{t-1}}(z_{m_t})^\top K_{\cZ_{t-1} \cZ_{t-1}}^{-1} k_{\cZ_{t-1}}(z_{m_t})$ and with $w_t = K_{\cZ_{t-1} \cZ_{t-1}}^{-1} k_{\cZ_{t-1}}(z_{m_t})$.

\subsection{Kernel Online Row Sampling (KORS) Subroutine}

As in \cite{pmlr-calandriello2017}, let us define a projection dictionary $\cZ_t$ as a collection of indexed anchor points $\{(z_{t_i} \}_{1\leq i \leq m_t}$ where $m_t = |\cZ_t|$ as well as the rescaling diagonal matrix $S_{\cZ_t}$ with $1/\sqrt{\tilde{p}_{z_s}}$ corresponding to the past sampling probabilities of points $z \in \cZ_t$, this matrix is of size $m_t \times m_t$. At each time step, KORS temporarily adds $t$ with weight 1 to the temporary dictionary $\mathcal{Z}_{t}^{*}$ and accordingly augments the corresponding matrix $S_{\cZ_t^*}$. The augmented dictionary is then used to compute the ridge leverage score (RLS) estimator:
\begin{equation}
    \tilde{\tau}_{t} = \dfrac{1+\epsilon}{\mu} \left( k(s_t, s_t) - K_{\cZ_t^*}(s_t)^\top S_{\cZ_t^*} (S_{\cZ_t^*}^\top K_{\cZ_t^* \cZ_t^*} S_{\cZ_t^*} + \mu I)^{-1} S_{\cZ_t^*}^\top  K_{\cZ_t^*}(s_t) \right)\,.
\label{eq:rls_score}
\end{equation}
Afterward, it draws a Bernoulli random variable $z_t$ proportionally to $\tilde{\tau}_t$, if it succeeds, ($z_t = 1$) the point is deemed relevant and added to the dictionary, otherwise it is discarded and never added.

% \begin{algorithm}[H]
% \SetAlgoLined
% \KwIn{Time $t$, past dictionary $\mathcal{Z}$, feature vector $\varphi_t$, regularization $\mu$, accuracy $\epsilon$, budget $\beta$}

%  Construct temporary dictionary $\mathcal{I}_{t, *} = \mathcal{Z} \cup \{ (t,1)\}$\;
%  Compute $\tilde{\tau}_{t}$ using Eq.~\eqref{eq:rls_score} \;
%  Compute $\tilde{p}_t = \min \{ \beta \tilde{\tau}_{t, t}, 1 \}$ \;
%  Draw $z_t \sim \mathcal{B}(\tilde{p}_t)$ and if $z_t =1$, add $(t, 1/\tilde{p}_t)$ to $\mathcal{Z}$\;
%  \KwResult{Dictionary $\mathcal{Z}$}

%  \caption{}
%  \label{alg:kors_}
% \end{algorithm}

\begin{algorithm}[h]
\SetAlgoLined
\KwIn{Time $t$, past dictionary $\mathcal{Z}$, context-action $s_t$, regularization $\mu$, accuracy $\epsilon$, budget $\gamma$}

%  Construct virtual dictionary $\mathcal{Z}_{t, *} = \mathcal{Z} \cup \{s_t\}$\;
 Compute the leverage score $\tilde{\tau}_{t}$ from $\mathcal Z, s_t, \mu, \epsilon$ \;
 Compute $\tilde{p}_t = \min \{ \gamma \tilde{\tau}_{t}, 1 \}$ \;
 Draw $z_t \sim \mathcal{B}(\tilde{p}_t)$ and if $z_t =1$, add $s_t$ to $\mathcal{Z}$\;
 \KwResult{Dictionary $\mathcal{Z}$}
 \caption{Incremental Kernel Online Row Sampling (KORS) subroutine}
%  \label{alg:kors}
\end{algorithm}

Here, all rows and columns for which $S_{t, *}$ is zero (all points outside the temporary dictionary $\mathcal{I}_{t, *}$) do not influence the estimator, so they can be excluded from the computation. As a consequence, the RLS score $\tilde{\tau}_{t}$ can be computed efficiently in $\cO((m_t + 1)^2)$ space and $\cO((m_t+1)^2)$ time, using an incremental update in Eq.~\eqref{eq:rls_score}.

As a side note, the quantity $\tilde{\tau}$ is an estimator of the exact RLS quantity $\tau_{t}$ (see \cite{pmlr-calandriello2017}): 
\begin{equation}
    \tau_{t} =   \varphi_t^\top (K_t + \mu I)^{-1} \varphi_t \,.
\label{eq:rls_score1}
\end{equation}
Here, leverage scores are used to measure the correlation between the new point $\varphi_t$ w.r.t. the previous $t-1$ points $\{ \varphi_i \}_{i \leq t-1}$, and therefore how essential it is in characterizing the dataset. In particular, if $\varphi_t$ is completely orthogonal to the other points, its RLS is maximized, while in the opposite case it would be minimal. In the incremental strategy of the Nystr\"om dictionary building, we use the RLS estimates to add anchor points that are as informative as possible.

\section{Experiment details}

\label{app:experimental_details}

In this section we provide further details as well as additional discussions and numerical results on our proposed method. 

\subsection{Reproducibility and Implementations}

We provide code that is accessible at the link \url{https://github.com/criteo-research/Efficient-Kernel-UCB}. All experiments were run on a single CPU core (2 x Intel(R) Xeon(R) Gold 6146 CPU@ 3.20GHz). 

\paragraph{Baseline implementations} We implemented the BKB and BBKB algorithms in \citep{calandriello19a} and \citep{calandriello20a} by introducing modifications in their implementation to handle contextual information. For both methods, in the contextual variant, each update involves the computation of a new covariance matrix $K_{\cZ \cS} K_{\cS \cZ}$ while the original algorithms do not involve contexts and consider a finite set of actions, allowing to compute the covariance matrix on the finite set of actions (which is done for computational efficiency and is impossible in the joint context-action space). The baselines were carefully optimized using the Jax library (\url{https://github.com/google/jax}) to allow for just in time compilations of similar blocks in every methods. 

\paragraph{Empirical setting} In our empirical setting we aimed at showing the regret/computational complexity compromise that is achieved by each method. In particular, both the CBBKB method \citep{calandriello20a} and our EK-UCB algorithm use additional hyperparameters than the CBKB. As a matter of fact, CBBKB uses an accumulation threshold $C$ and is used for the 'resparsification' step, with dictionary updates based on all historical states. EK-UCB also uses the hyperparameter $\mu$ in KORS that is set to $\lambda$ for optimal regret-time compromise (see Theorem~\ref{thm:regret_bound_projected}). The KORS algorithm uses a budget parameter $\gamma$, for which we found empirically good performances when $\gamma \approx \lambda$. We tried our method with a grid on hyperparameters and discuss their influence in the next subsection. 

\subsection{Additional Results}

In this section we provide additional numerical experiment discussions. 

\subsubsection{Additional discussions on the setting of Section \ref{sec:exps}}

\begin{figure*}[h!]
    \centering
\begin{subfigure}
  \centering
    \includegraphics[height=0.3\linewidth]{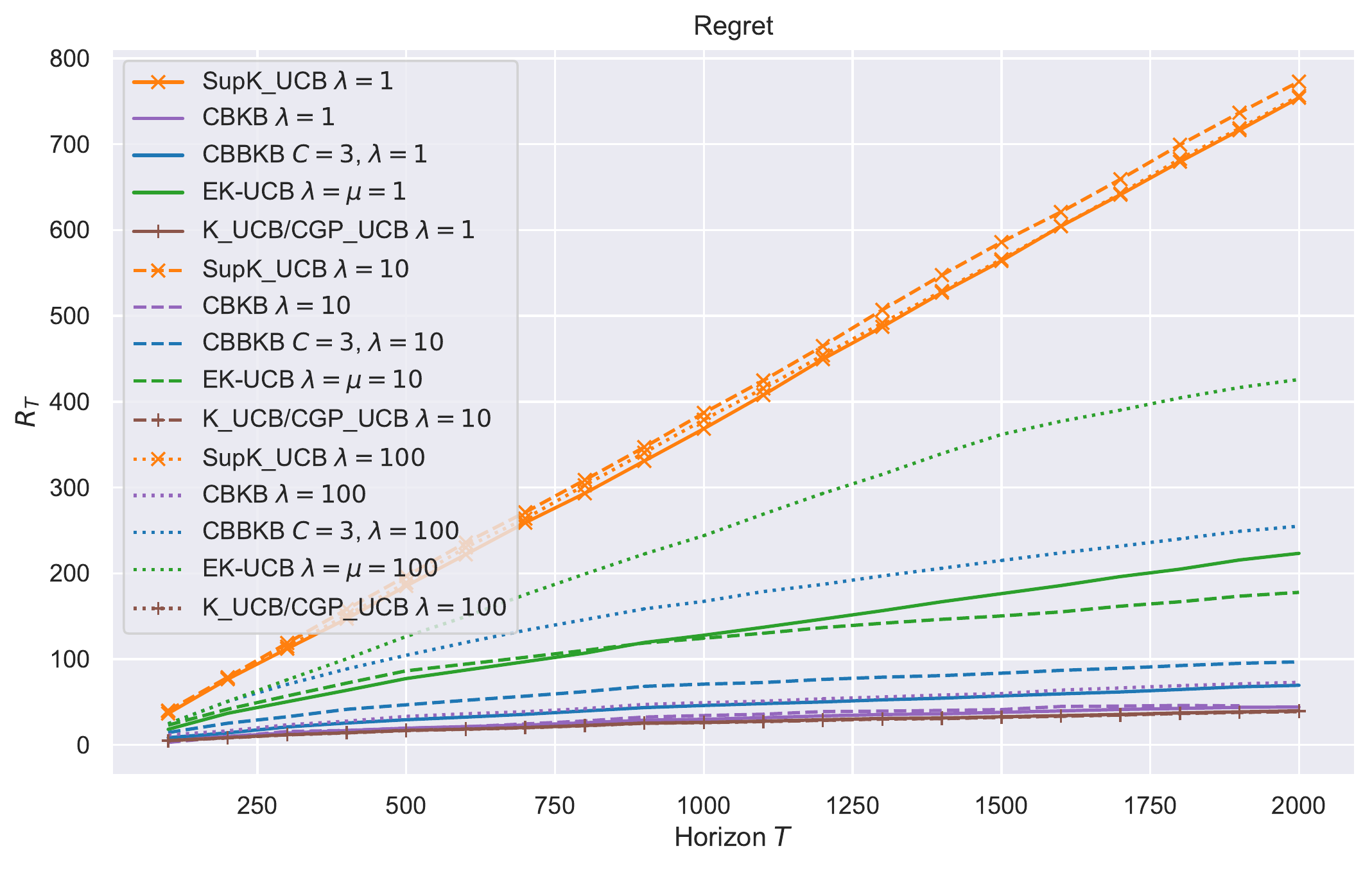}
  \label{fig:regret}
\end{subfigure}
\begin{subfigure}
  \centering
    \includegraphics[height=0.3\linewidth]{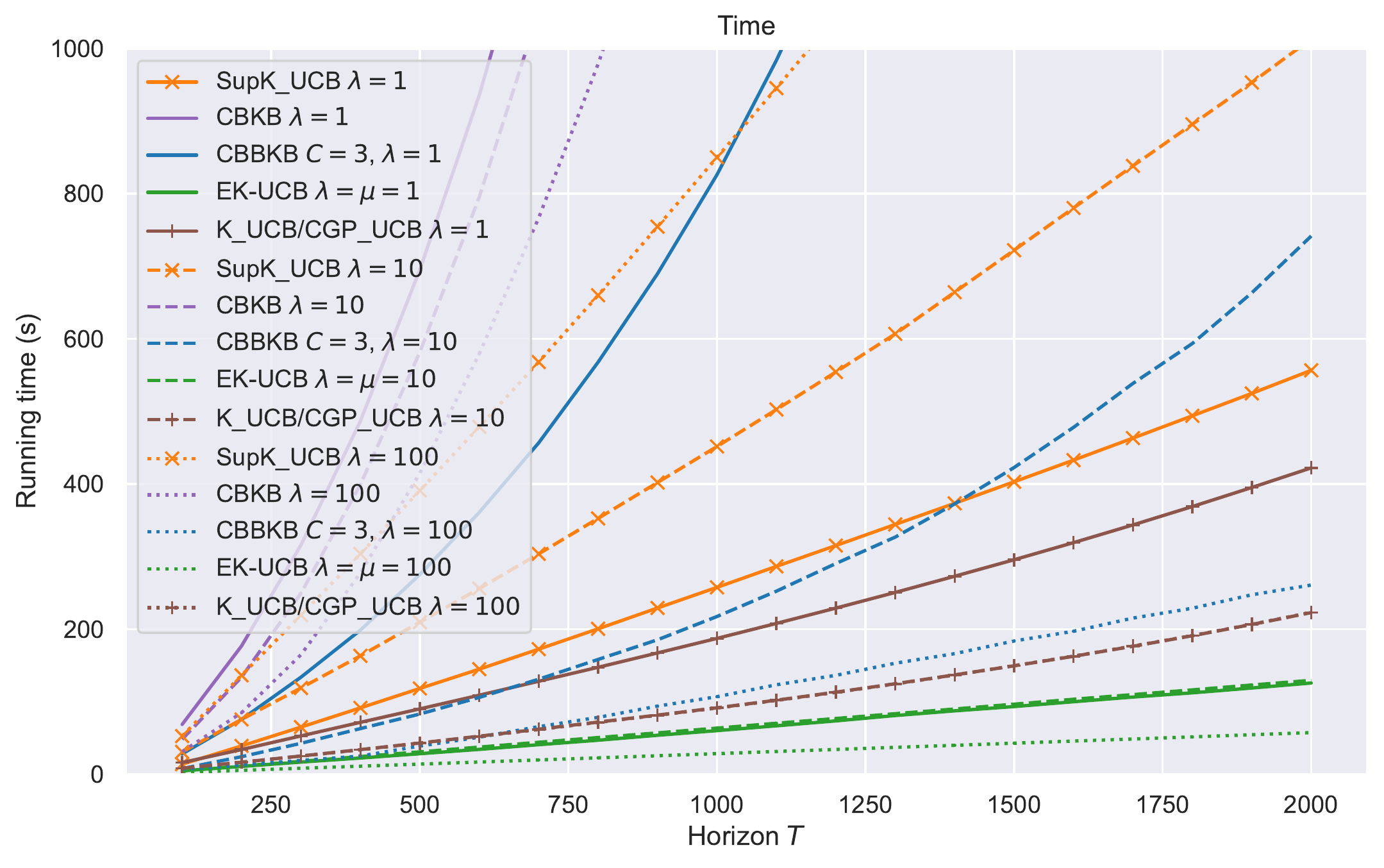}
  \label{fig:time}
\end{subfigure}

\caption{'Bump' setting: Regret and running times of EK-UCB, CBBKB and CBKB, with $T=2000$ and $\lambda=\mu$ (see Corollary \ref{cor:bound_capacity_condition} and \ref{cor:bound_capacity_condition_projected}) with varying $\lambda$ and $C=3$ for CBBKB. EK-UCB matches the best regret-time compromise.}
\label{fig:experiment_bump_appx1}
\end{figure*}

\begin{figure*}[h!]
    \centering
\begin{subfigure}
  \centering
    \includegraphics[height=0.3\linewidth]{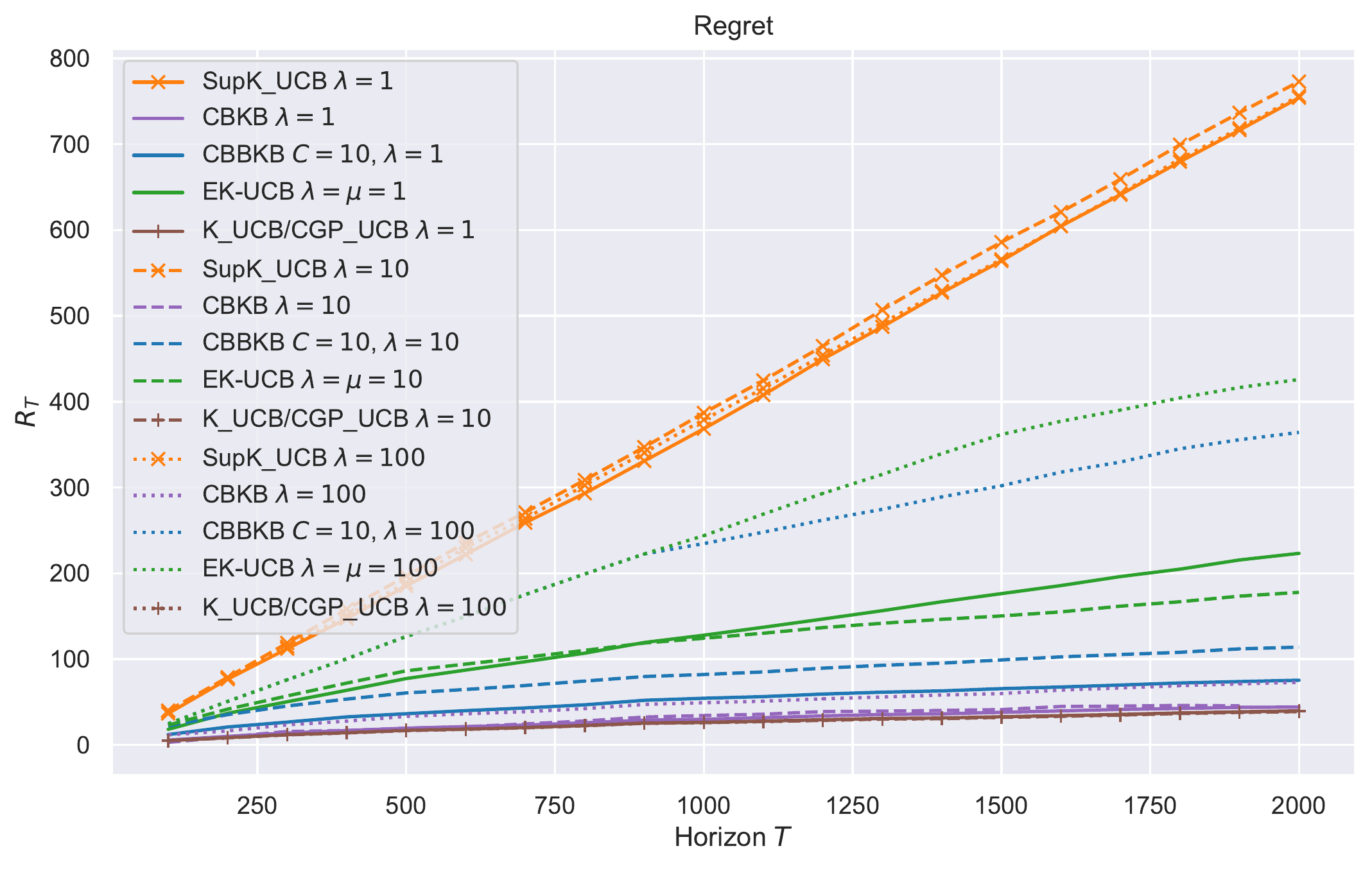}
  \label{fig:regret}
\end{subfigure}
\begin{subfigure}
  \centering
    \includegraphics[height=0.3\linewidth]{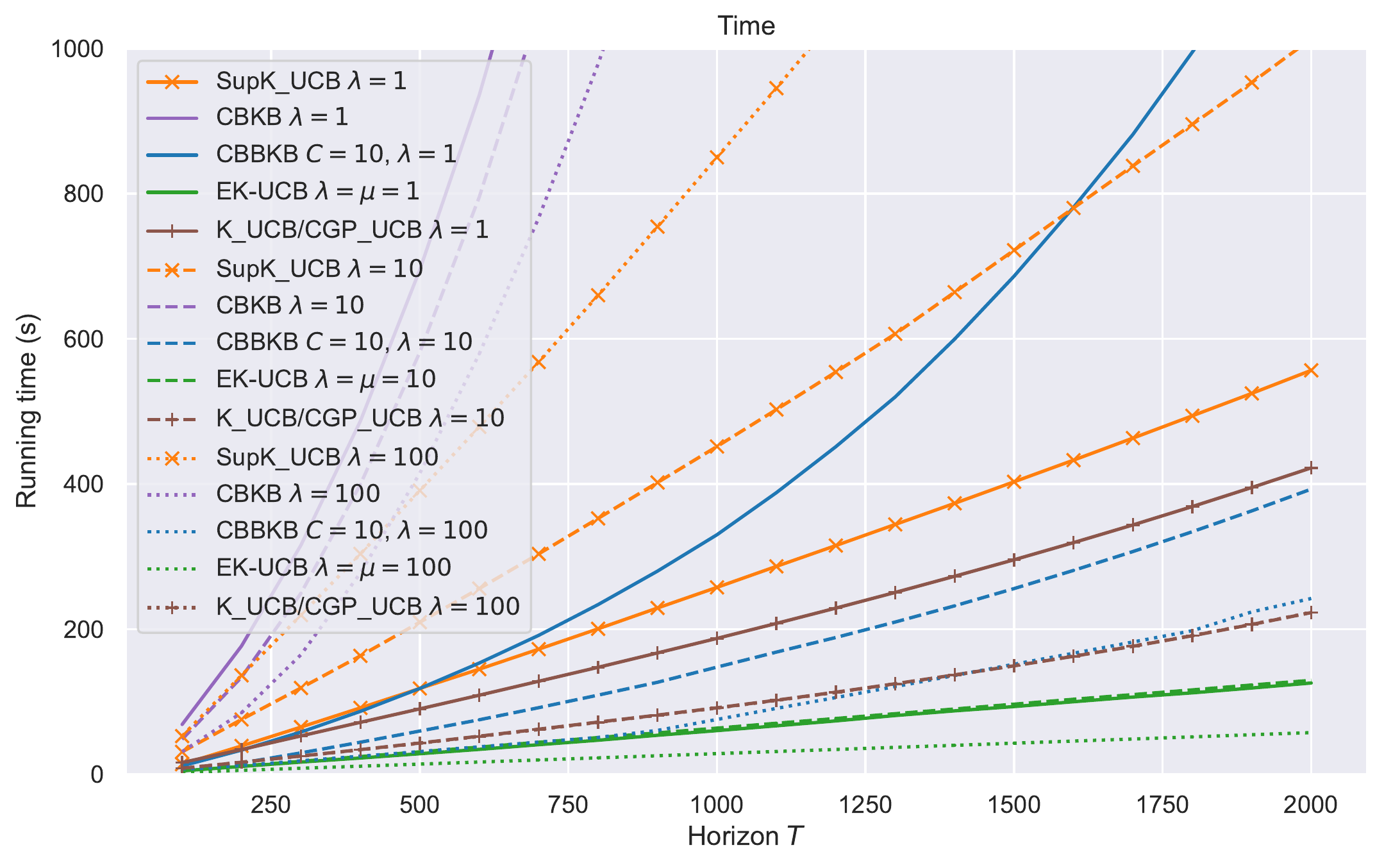}
  \label{fig:time}
\end{subfigure}

\caption{'Bump' setting: Regret and running times of EK-UCB, CBBKB and CBKB, with $T=2000$ and $\lambda=\mu$ with varying $\lambda$ and $C=10$ for CBBKB. EK-UCB matches the best regret-time compromise.}
\label{fig:experiment_bump_appx2}
\end{figure*}

\begin{figure*}[h!]
    \centering
\begin{subfigure}
  \centering
    \includegraphics[height=0.3\linewidth]{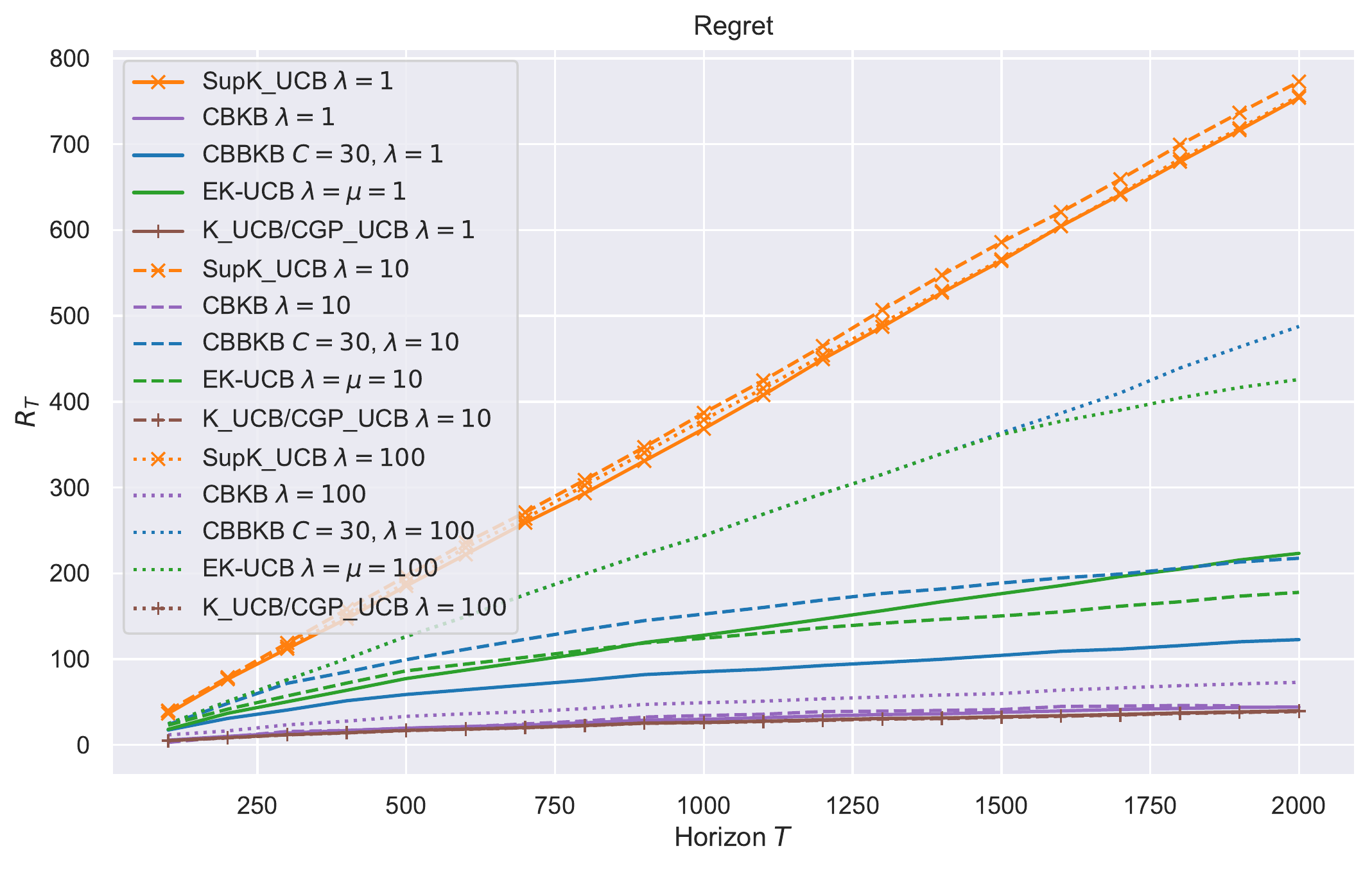}
  \label{fig:regret}
\end{subfigure}
\begin{subfigure}
  \centering
    \includegraphics[height=0.3\linewidth]{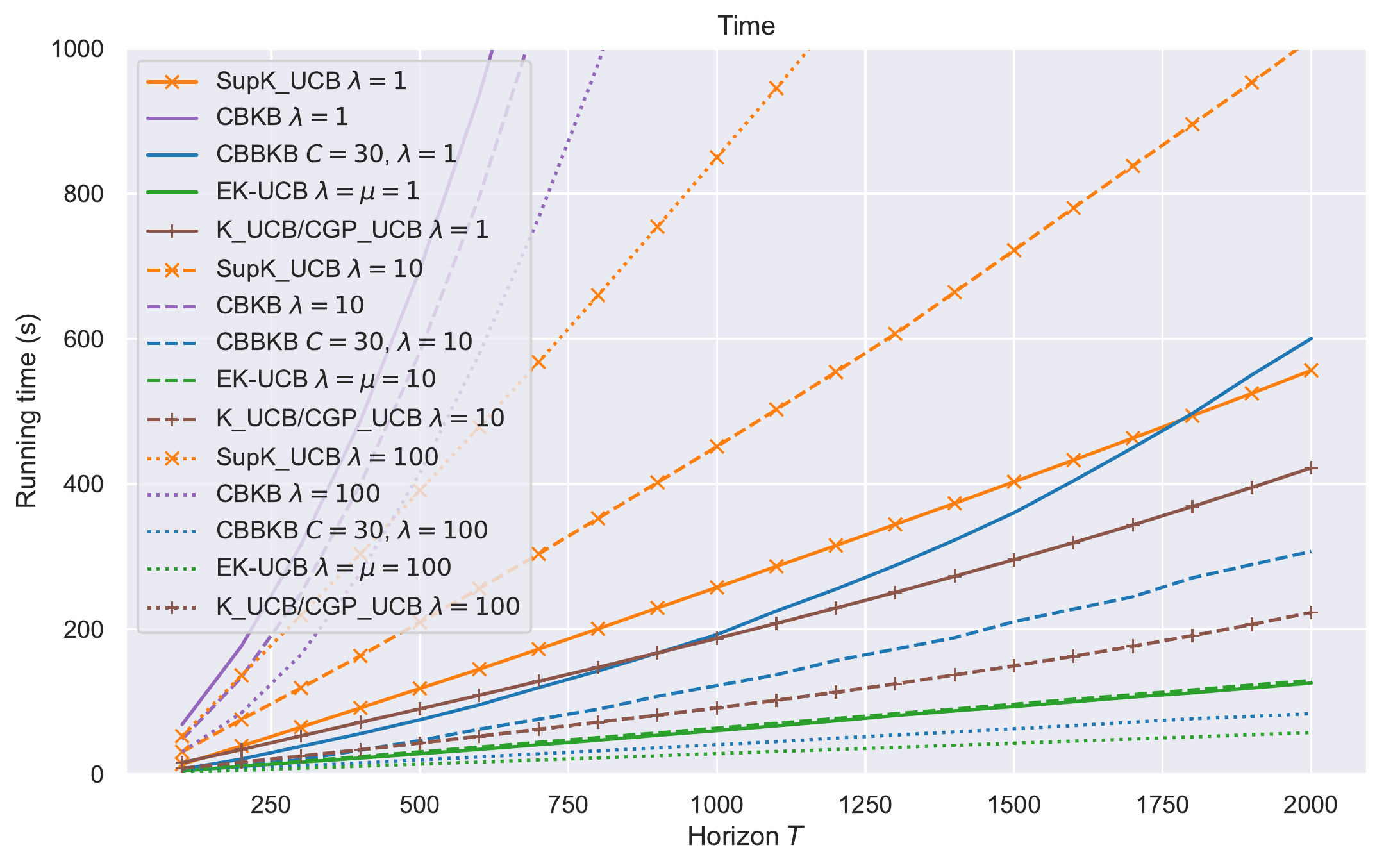}
  \label{fig:time}
\end{subfigure}

\caption{'Bump' setting: Regret and running times of EK-UCB, CBBKB and CBKB, with $T=2000$ and $\lambda=\mu$ with varying $\lambda$ and $C=30$ for CBBKB. EK-UCB matches the best regret-time compromise}
\label{fig:experiment_bump_appx3}
\end{figure*}

We present additional results on the synthetic setting presented in Section \ref{sec:exps} that we call 'Bump' in Figures \ref{fig:experiment_bump_appx1}, \ref{fig:experiment_bump_appx2}, \ref{fig:experiment_bump_appx3}. Here we fix $\lambda=\mu$ for EK-UCB and report the performances of the baselines with the same hyperparameters and make the accumulation threshold $C$ of CBBKB vary through the Figures \ref{fig:experiment_bump_appx1}, \ref{fig:experiment_bump_appx2}, \ref{fig:experiment_bump_appx3}. We provide more discussion on the methods we evaluated.

\paragraph{More dictionary updates lead to better regret but a higher computational complexity} We note that the CBKB baseline achieves satisfactory regret but with a drastically higher computational time. This is due to the fact that it resamples the dictionary at each step and therefore resamples a dictionary at the price of a higher time complexity. As for CBBKB, throughout the Figures \ref{fig:experiment_bump_appx1}, \ref{fig:experiment_bump_appx2}, \ref{fig:experiment_bump_appx3}, we can see that the accumulation threshold $C$ that controls the anchor point update frequency determines the regret-time compromise. The lower $C$, the better is the regret but the higher is the computational time. We can see through the figures that for all values of $C$, our EK-UCB method achieves similar or better (especially when $C=30$) regret than CBBKB while always being both faster than CBBKB but more importantly faster than K-UCB. Overall, EK-UCB proposes the most satisfactory regret-time compromise. Moreover, we see that the SupK-UCB method also performs poorly even with different parameters $\lambda$ and that the optimized K-UCB method also performs better than efficient strategies when the computational overheads of dictionary buildings overtake the efficient kernel approximations.

\paragraph{The regularization parameter controls the regret-time comprise in EK-UCB} In our method, we can see that the higher $\lambda$ (with $\lambda=\mu$) the faster the algorithm is but the worse is its regret. As discussed in Corollary~\ref{cor:bound_capacity_condition} and~\ref{cor:bound_capacity_condition_projected}, we use the heuristic to take $\lambda \approx \sqrt{T}$ and set $\mu=\lambda$ afterwards to enjoy the optimal guarantees of our algorithm.

\subsubsection{Additional synthetic settings}

\label{app:additional-settings}
In this section we introduce additional settings that we call the 'Chessboard' setting as well as the 'Step Diagonal' setting. The two settings lead to similar numerical conclusions as the previous one. We provide more discussions here.

\begin{figure*}[h!]
    \centering
\begin{subfigure}
  \centering
    \includegraphics[height=0.3\linewidth]{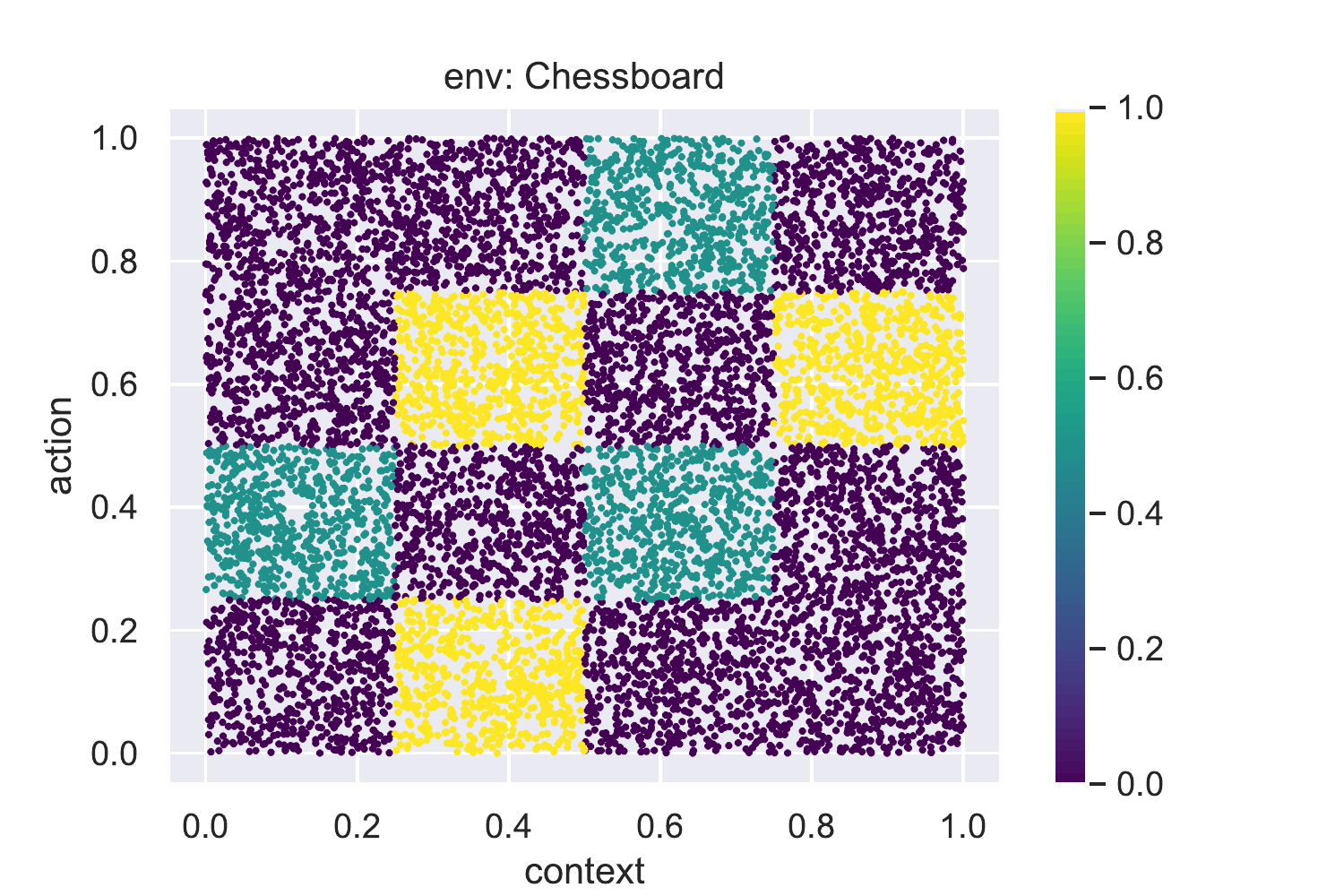}
%   \label{fig:regret}
\end{subfigure}
\begin{subfigure}
  \centering
    \includegraphics[height=0.3\linewidth]{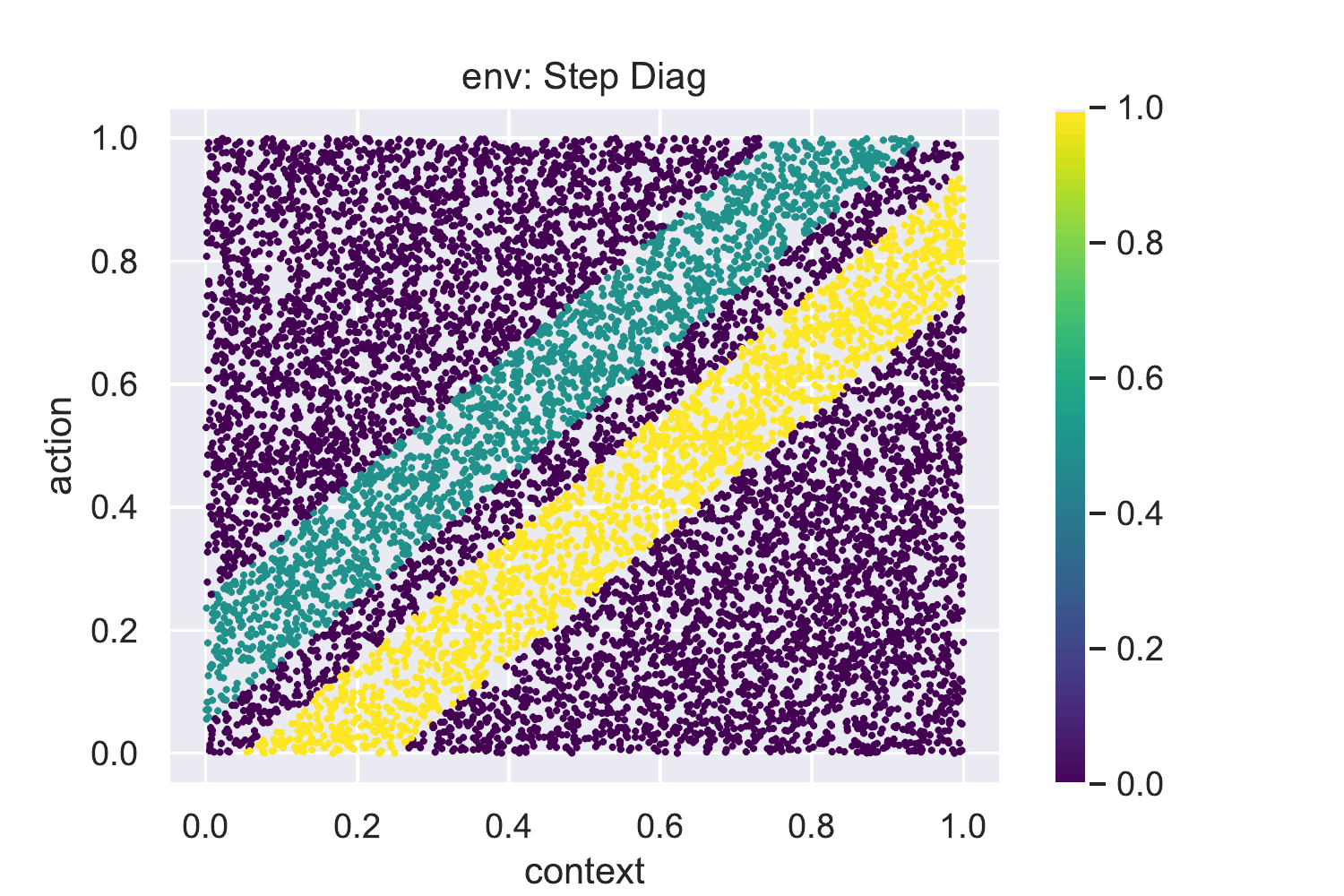}
%   \label{fig:time}
\end{subfigure}
\caption{Chessboard (left) and Step Diagonal (right) synthetic setups.}
\label{fig:2otherenv}
\end{figure*}

\begin{figure*}[h!]
    \centering
\begin{subfigure}
  \centering
    \includegraphics[height=0.3\linewidth]{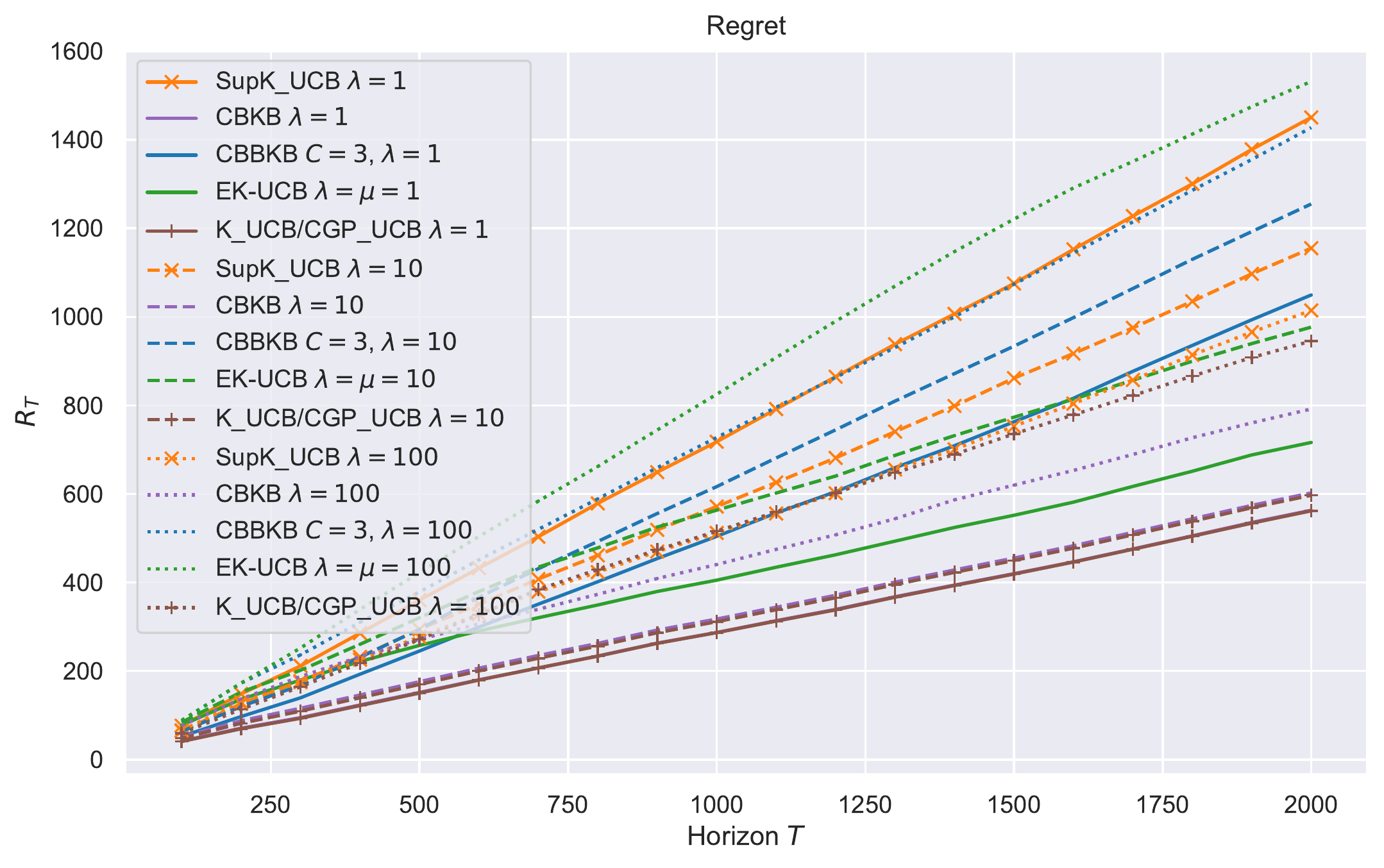}
  \label{fig:regret}
\end{subfigure}
\begin{subfigure}
  \centering
    \includegraphics[height=0.3\linewidth]{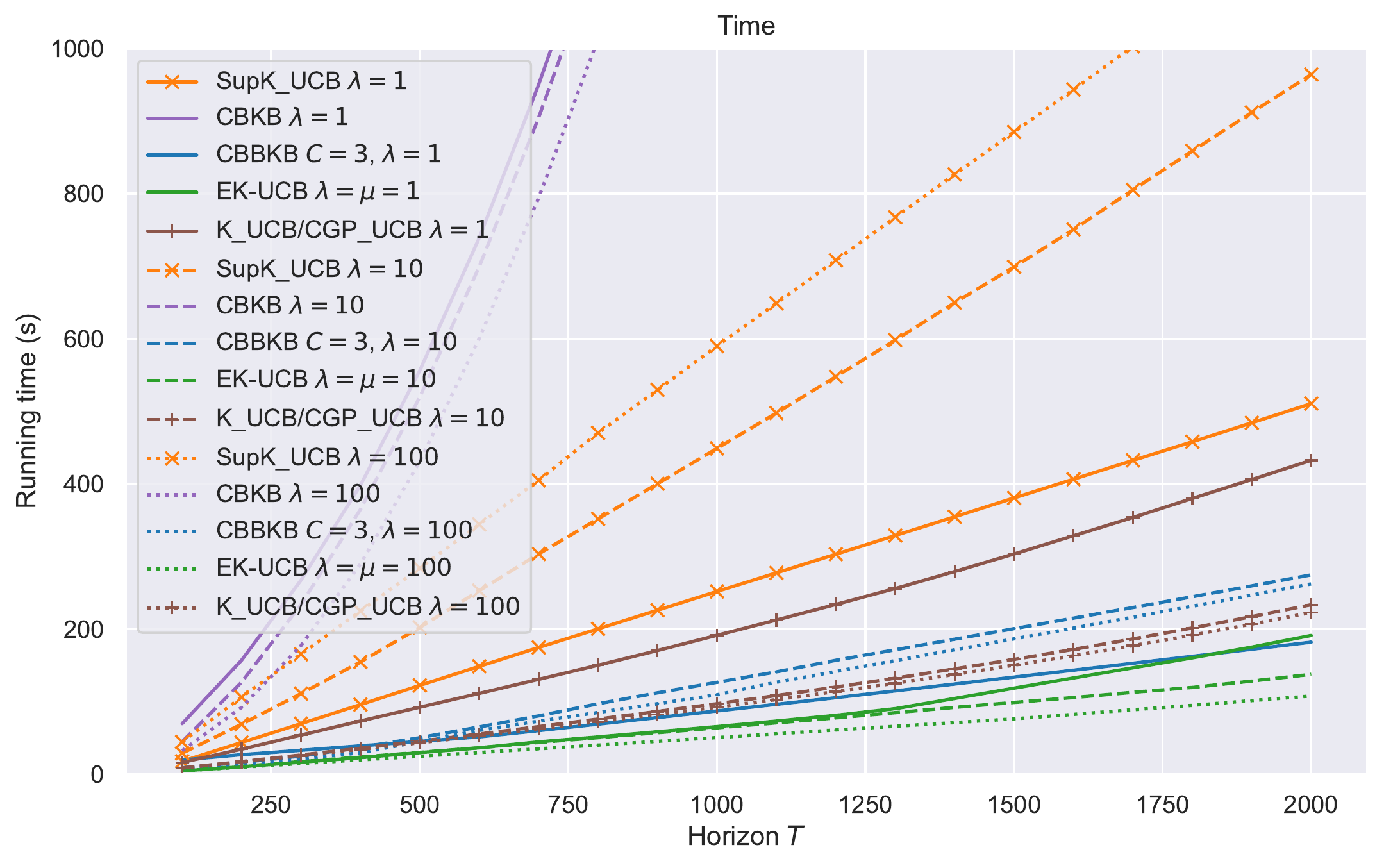}
  \label{fig:time}
\end{subfigure}

\caption{'Chessboard' setting: Regret and running times of EK-UCB, CBBKB and CBKB, with $T=2000$ and $\lambda=\mu$  with varying $\lambda$ and $C=3$ for CBBKB.}
\label{fig:experiment_squares}
\end{figure*}

\begin{figure*}[h!]
    \centering
\begin{subfigure}
  \centering
    \includegraphics[height=0.3\linewidth]{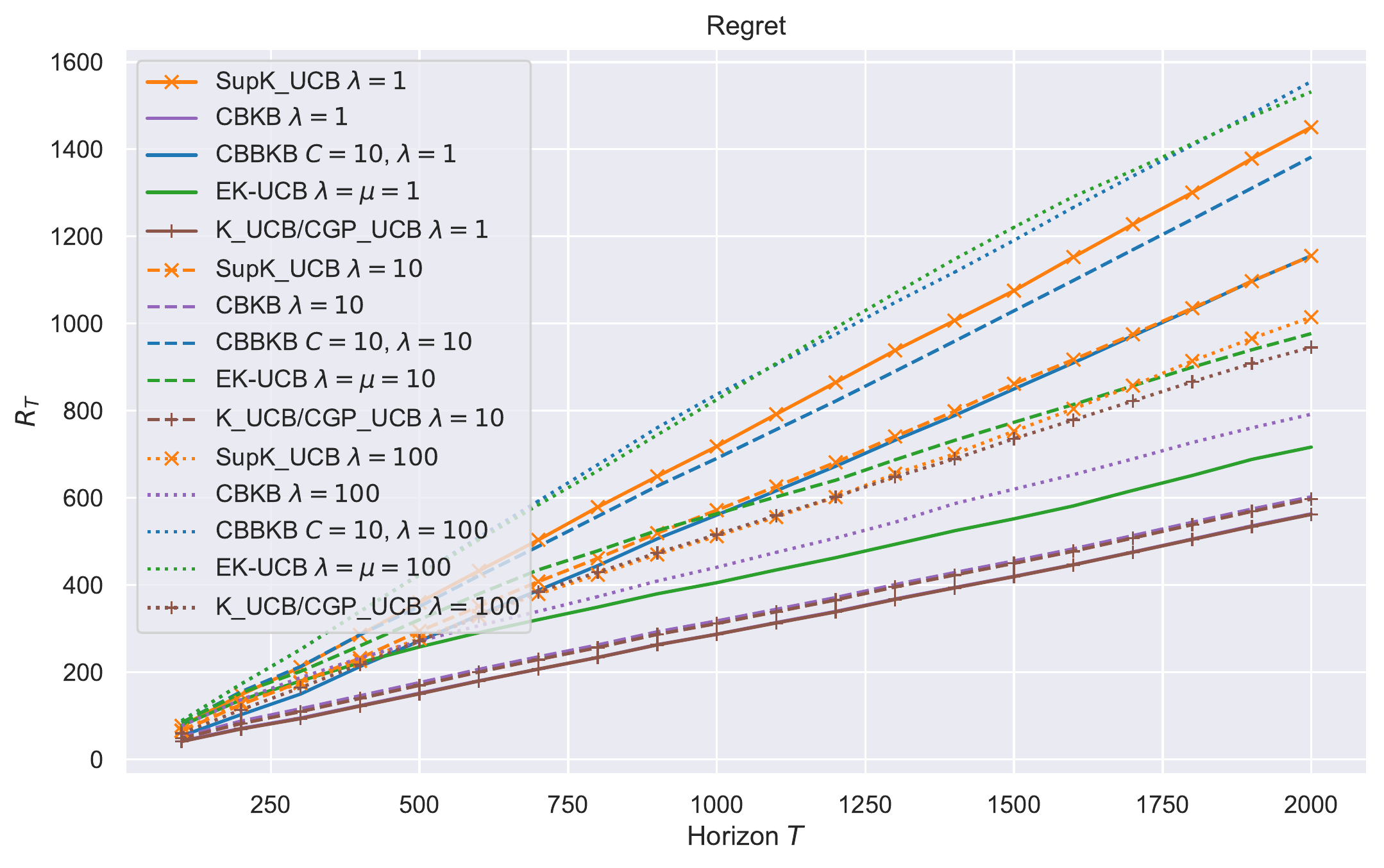}
  \label{fig:regret}
\end{subfigure}
\begin{subfigure}
  \centering
    \includegraphics[height=0.3\linewidth]{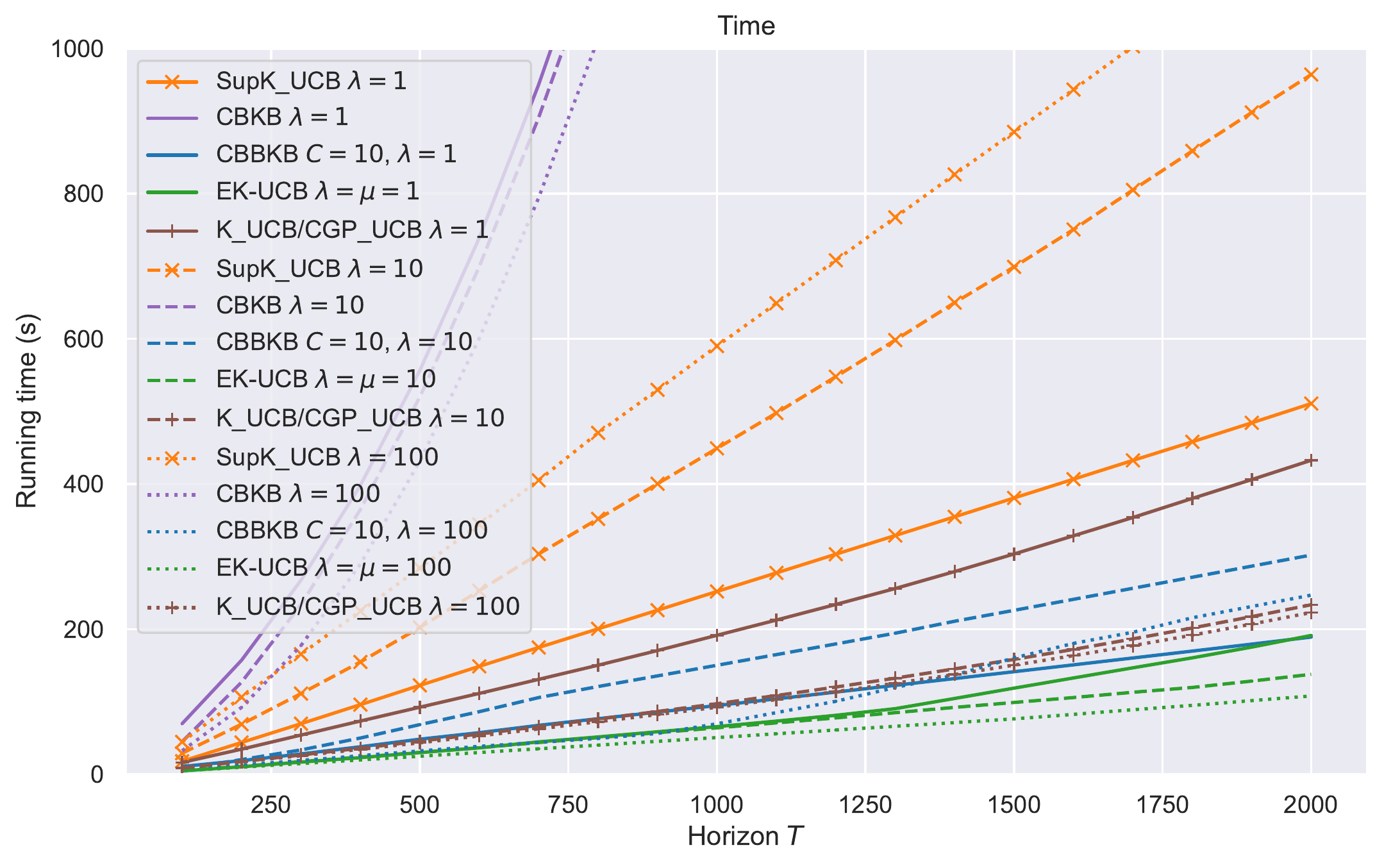}
  \label{fig:time}
\end{subfigure}

\caption{'Chessboard' setting: Regret and running times of EK-UCB, CBBKB and CBKB, with $T=2000$ and $\lambda=\mu$  with varying $\lambda$ and $C=10$ for CBBKB.}
\label{fig:experiment_squares2}
\end{figure*}

\begin{figure*}[h!]
    \centering
\begin{subfigure}
  \centering
    \includegraphics[height=0.3\linewidth]{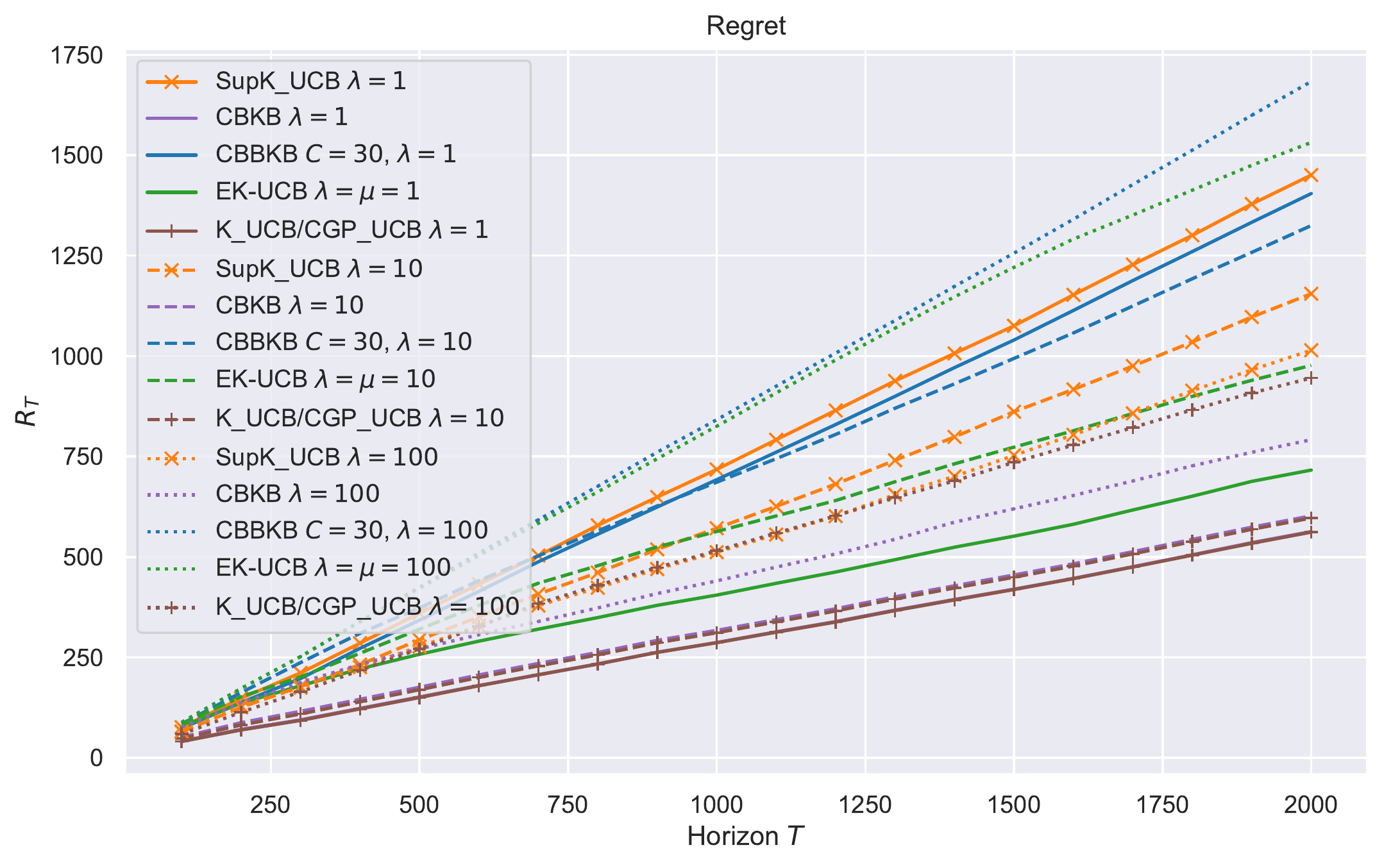}
  \label{fig:regret}
\end{subfigure}
\begin{subfigure}
  \centering
    \includegraphics[height=0.3\linewidth]{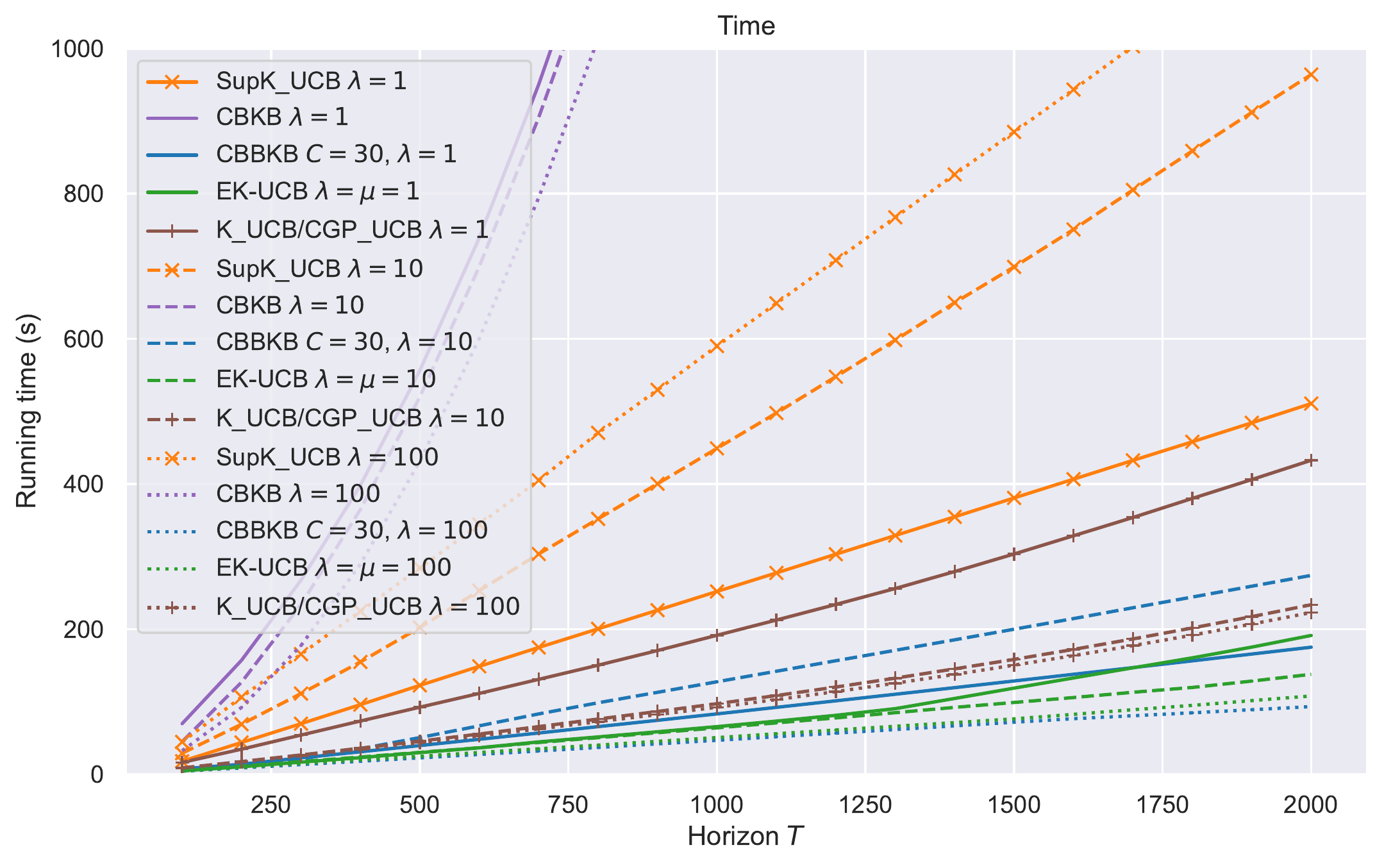}
  \label{fig:time}
\end{subfigure}

\caption{'Chessboard' setting: Regret and running times of EK-UCB, CBBKB and CBKB, with $T=2000$ and $\lambda=\mu$  with varying $\lambda$ and $C=30$ for CBBKB.}
\label{fig:experiment_squares3}
\end{figure*}

\begin{figure*}[h!]
    \centering
\begin{subfigure}
  \centering
    \includegraphics[height=0.3\linewidth]{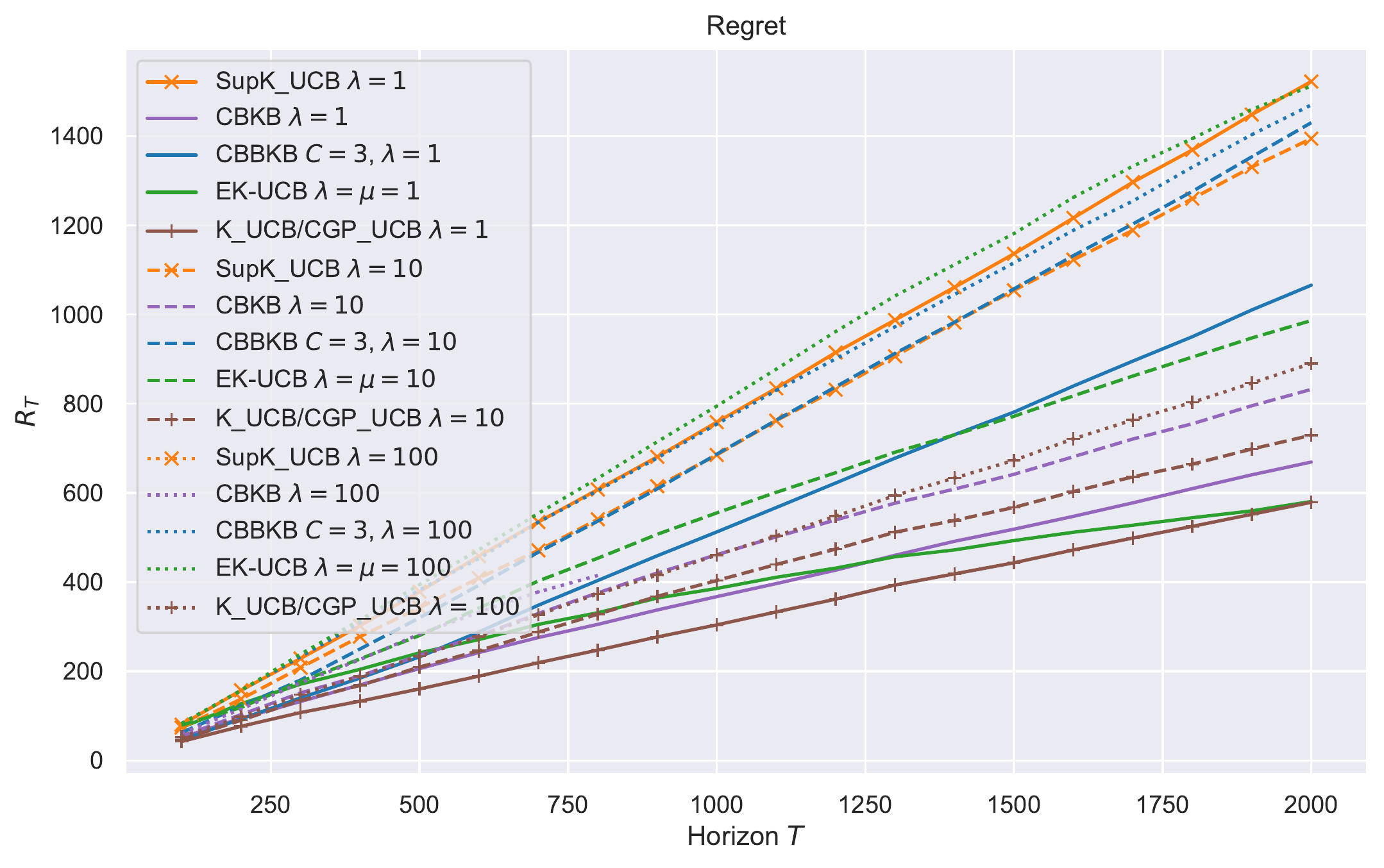}
  \label{fig:regret}
\end{subfigure}
\begin{subfigure}
  \centering
    \includegraphics[height=0.3\linewidth]{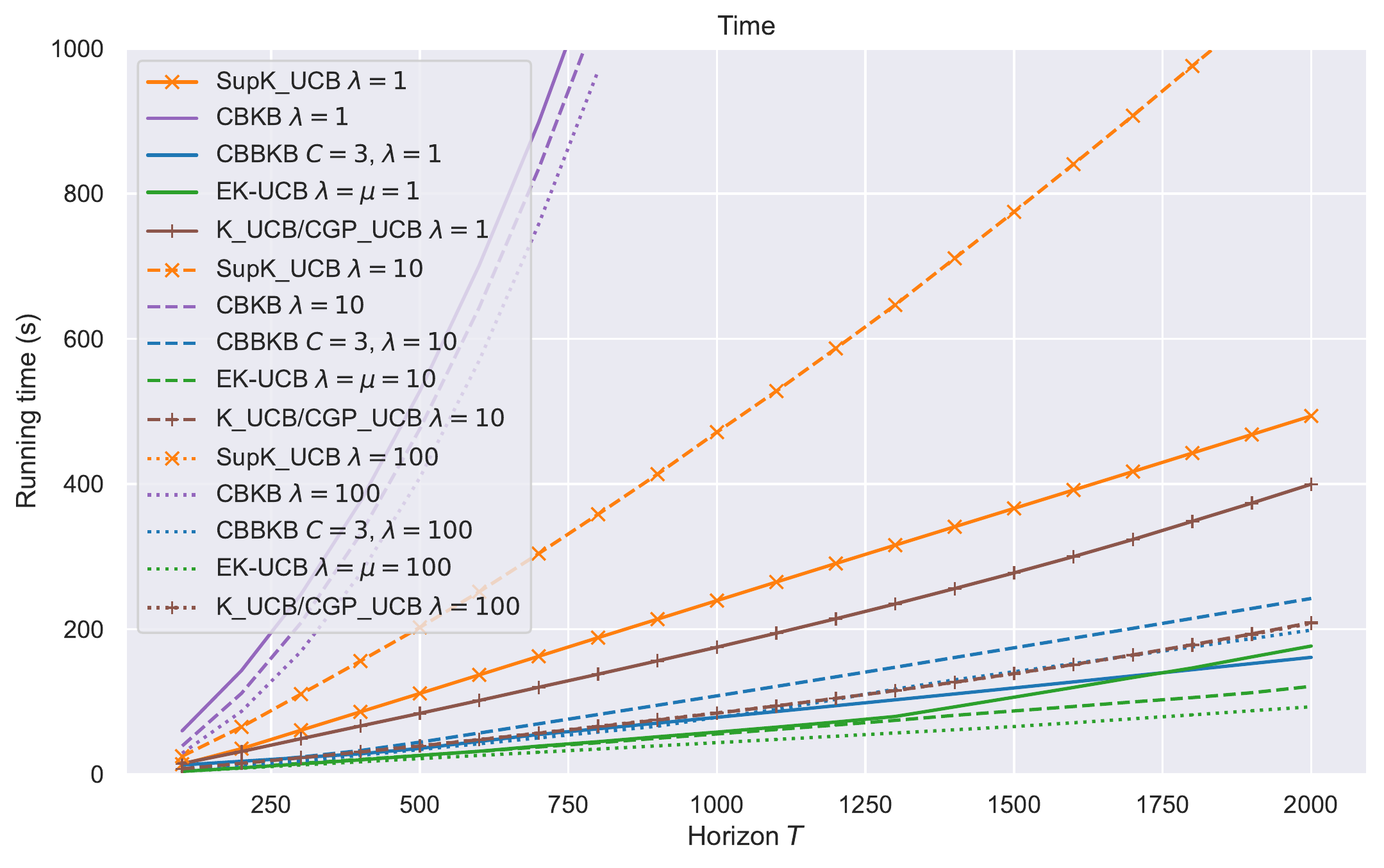}
  \label{fig:time}
\end{subfigure}

\caption{'Step Diagonal' setting: Regret and running times of EK-UCB, CBBKB and CBKB, with $T=2000$ and $\lambda=\mu$  with varying $\lambda$ and $C=3$ for CBBKB.}
\label{fig:experiment_stepdiag}
\end{figure*}

\begin{figure*}[h!]
    \centering
\begin{subfigure}
  \centering
    \includegraphics[height=0.3\linewidth]{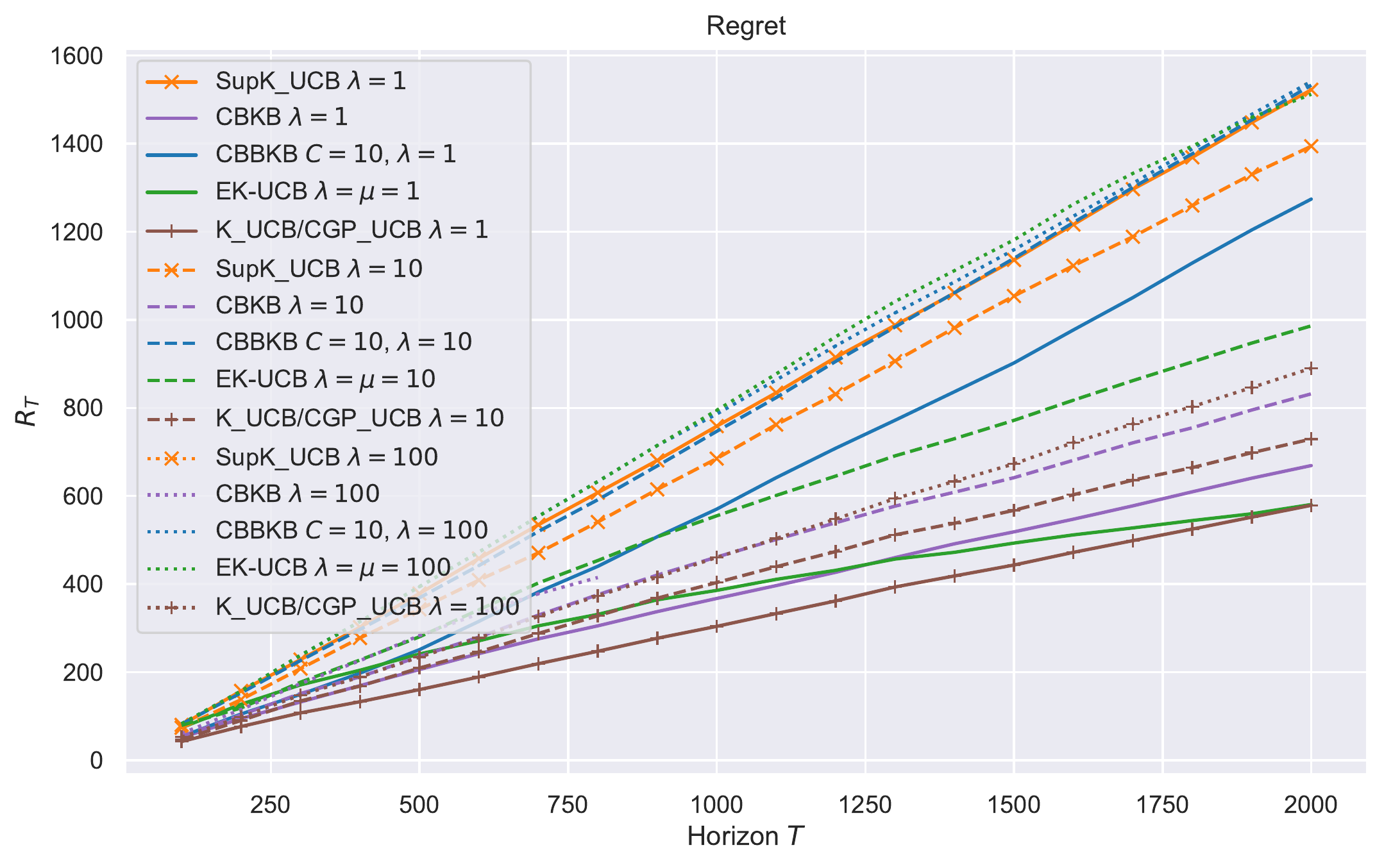}
  \label{fig:regret}
\end{subfigure}
\begin{subfigure}
  \centering
    \includegraphics[height=0.3\linewidth]{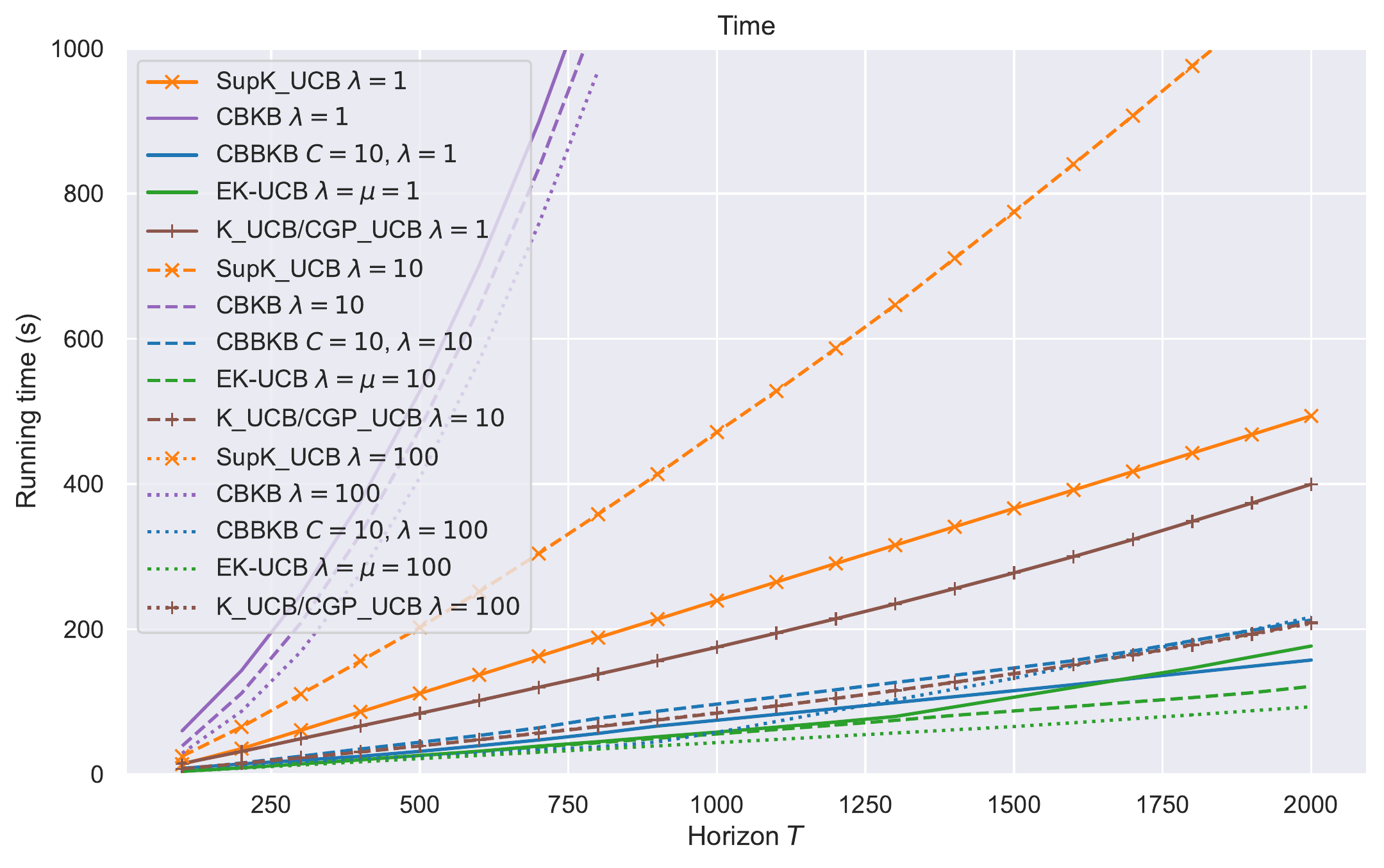}
  \label{fig:time}
\end{subfigure}

\caption{'Step Diagonal' setting: Regret and running times of EK-UCB, CBBKB and CBKB, with $T=2000$ and $\lambda=\mu$  with varying $\lambda$ and $C=10$ for CBBKB.}
\label{fig:experiment_stepdiag2}
\end{figure*}

\begin{figure*}[h!]
    \centering
\begin{subfigure}
  \centering
    \includegraphics[height=0.3\linewidth]{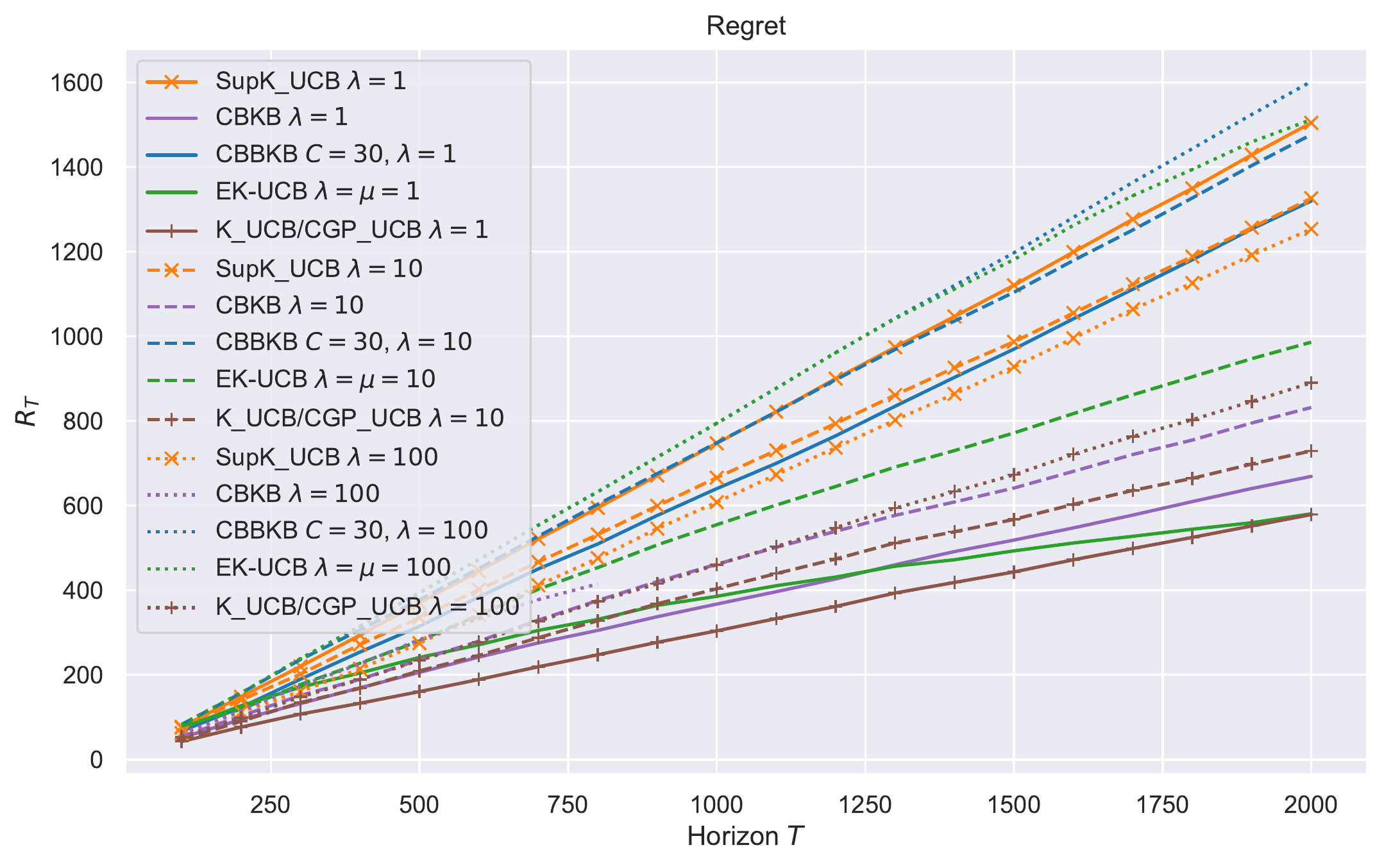}
  \label{fig:regret}
\end{subfigure}
\begin{subfigure}
  \centering
    \includegraphics[height=0.3\linewidth]{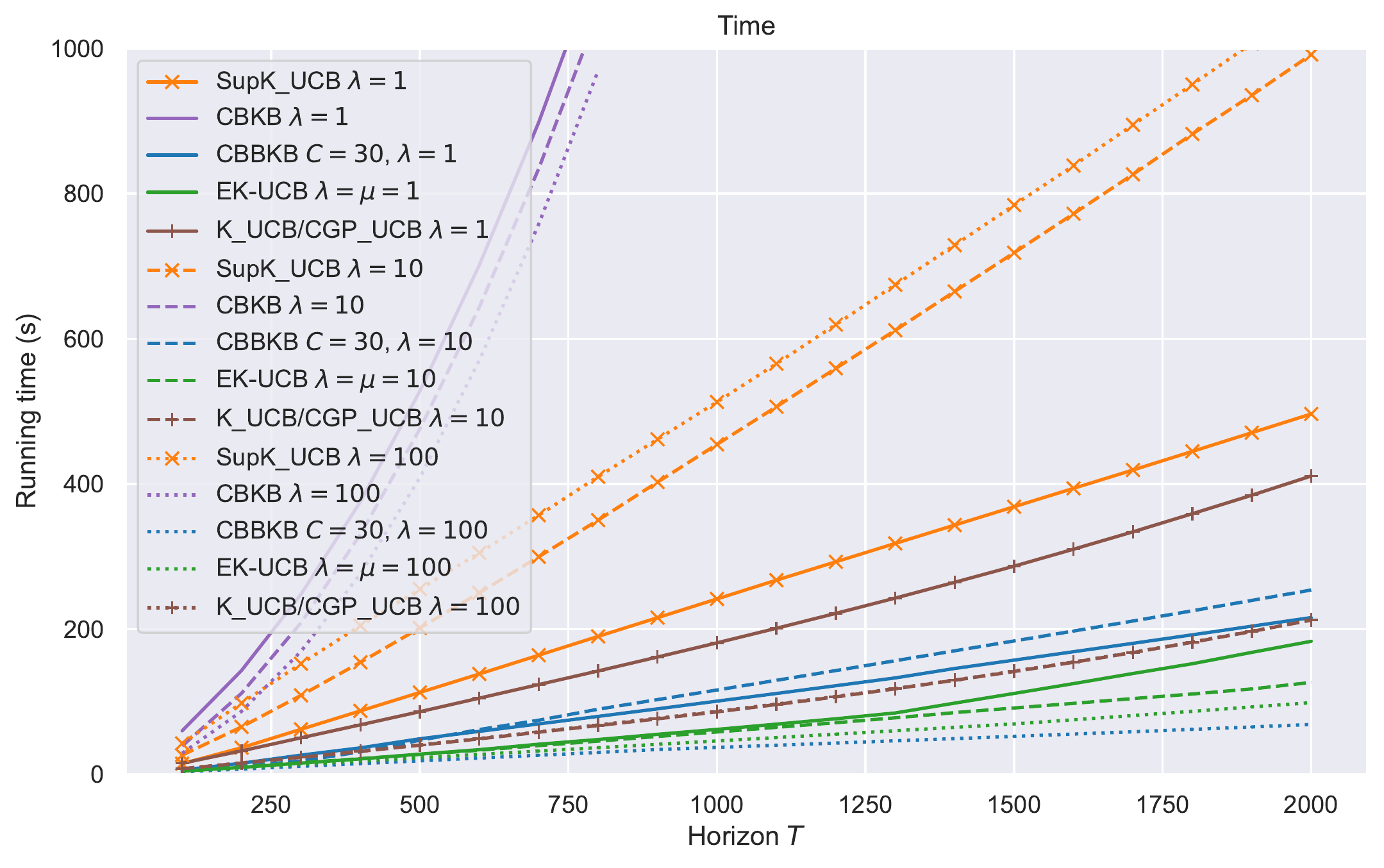}
  \label{fig:time}
\end{subfigure}

\caption{'Step Diagonal' setting: Regret and running times of EK-UCB, CBBKB and CBKB, with $T=2000$ and $\lambda=\mu$  with varying $\lambda$ and $C=30$ for CBBKB.}
\label{fig:experiment_stepdiag3}
\end{figure*}

\paragraph{Chessboard and Step Diagonal synthetic setups.} The 'Chessboard' synthetic setup is a contextual environment with a piecewise reward function over the joint context-action space $\cX \times \cA = [0,1] \times [0,1]$. More precisely, the joint 2D space is cut into a grid where the values are either 1, 0.5 or 0 according to the part of the grid.  Results are shown in Figures \ref{fig:experiment_squares}, \ref{fig:experiment_squares2}, \ref{fig:experiment_squares3}. The 'Step diagonal' synthetic setup is a contextual environment with a diagonal reward function over the joint context-action space $\cX \times \cA = [0,1] \times [0,1]$. More precisely, the joint 2D space has values of 0 everywhere except along two bands along the diagonal where the action and context values are identical with values 0.5 and 1 respectively on the sub diagonal and the above diagonal. Results are shown in Figures \ref{fig:experiment_stepdiag}, \ref{fig:experiment_stepdiag2}, \ref{fig:experiment_stepdiag3}. See the code for more details and an illustration of the settings in Fig \ref{fig:2otherenv}.

\paragraph{Regret-time compromise for CBBKB and EK-UCB.} The two settings show what both algorithms CBBKB and EK-UCB achieve as a regret-time compromise. In cases where $C$ is lower (note that CBKB corresponds to CBBKB with $C=1$) the regret often decreases at the price of higher computational time complexity. Similarly, we can notice that our method has better regrets when $\lambda$ is low, but with higher computational times, while still providing a benefit over to the K-UCB method, unlike CBBKB. We therefore note again that in practice, dictionary building computational overheads may influence the global computational complexity. Overall, our method with its incremental dictionary building strategy achieves the best satisfactory time-regret compromises in the Chessboard and Step Diagonal settings compared to both K-UCB and the efficient algorithms CBKB and CBBKB. 

\end{document}